\documentclass[journal]{IEEEtran}
\usepackage{tikz}
\usepackage{rotating}
\usepackage{algorithmic}
\usepackage{url}
\algsetup{linenosize=\tiny}
\usepackage{pgfplots}
\pgfplotsset{compat=newest}
\usepgfplotslibrary{units}
\usepackage{subfigure}
\usepackage{amsmath}
\usepackage{amsfonts}
\usepackage{multirow}
\usepackage{multicol}
\usepackage{booktabs}

\usepackage{mathrsfs}
\usepackage[inline]{enumitem}

\usepackage{wasysym}
\usepackage[linesnumbered,ruled,vlined]{algorithm2e} 
\ifCLASSINFOpdf
\else
 
\fi
\begin{document}
\title{Self Organizing Nebulous Growths for Robust and Incremental Data Visualization}
\author{Damith A. Senanayake,
Wei Wang, Shalin H. Naik,  
Saman Halgamuge 
\thanks{D. Senanayake, W. Wang and S. Halgamuge are all with the department of Mechanical Engineering of the University of Melbourne, Victoria, Australia}
\thanks{S. Naik is with the Walter and Eliza Hall Institute for Medical Research, Melbourne, Victoria, Australia}
}
\maketitle
\begin{abstract}

Non-parametric dimensionality reduction techniques, such as t-SNE and UMAP, are proficient in providing visualizations for datasets of fixed sizes. However, they cannot incrementally map and insert new data points into an already provided data visualization. We present Self-Organizing Nebulous Growths (SONG), a parametric nonlinear dimensionality reduction technique that supports incremental data visualization, i.e., incremental addition of new data while preserving the structure of the existing visualization. In addition, SONG is capable of handling new data increments, no matter whether they are similar or heterogeneous to the already observed data distribution. We test SONG on a variety of real and simulated datasets. The results show that SONG is superior to Parametric t-SNE, t-SNE and UMAP in incremental data visualization. Specifically, for heterogeneous increments, SONG improves over Parametric t-SNE by 14.98 \% on the Fashion MNIST dataset and 49.73\% on the MNIST dataset regarding the cluster quality measured by the Adjusted Mutual Information scores. On similar or homogeneous increments, the improvements are 8.36\% and 42.26\% respectively. Furthermore, even when the above datasets are presented all at once, SONG performs better or comparable to UMAP, and superior to t-SNE. We also demonstrate that the algorithmic foundations of SONG render it more tolerant to noise compared to UMAP and t-SNE, thus providing greater utility for data with high variance, high mixing of clusters, or noise.

\end{abstract}


\IEEEpeerreviewmaketitle

\section{Introduction}

\begin{figure}[t]
    \centering
    \includegraphics[width=0.75\linewidth]{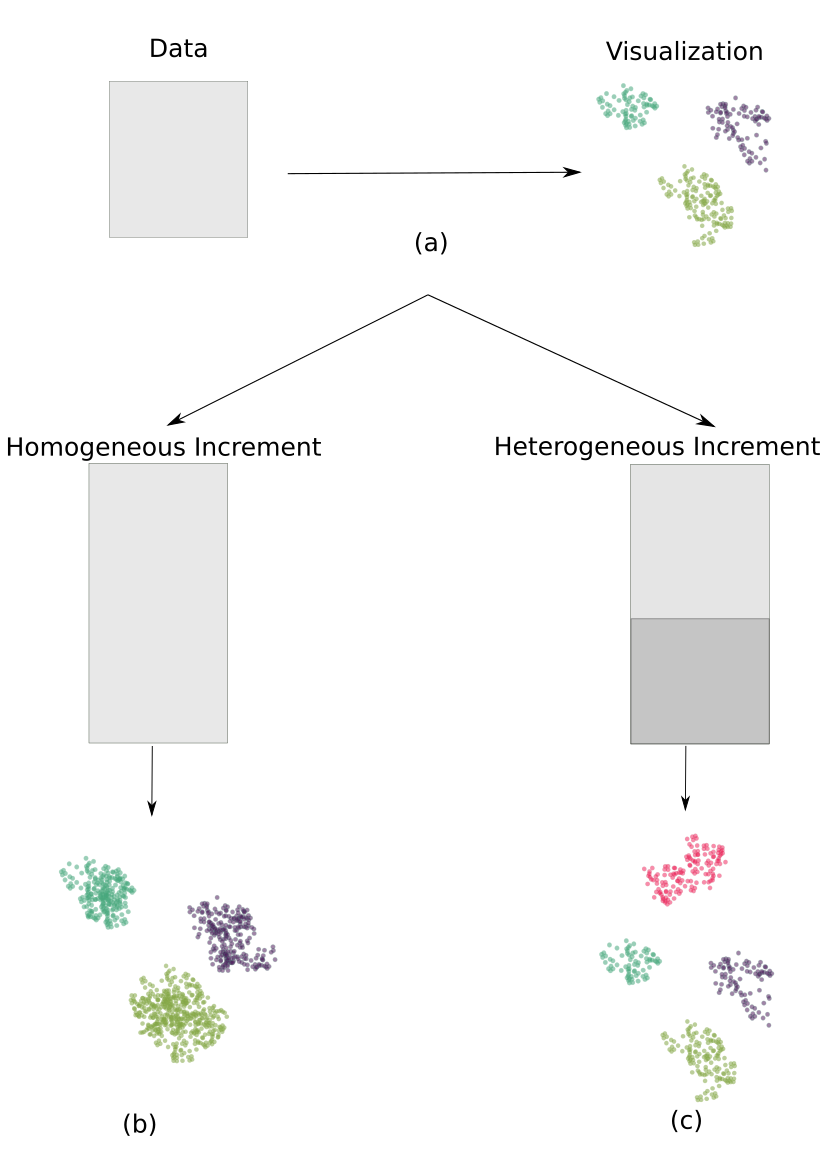}
    \caption{Illustration of incremental data visualization: a) the visualization of the initially available data contain three clusters; b) the initial data are augmented with homogeneous (similar) data, where clusters become denser in the visualization; c) the initial data are augmented with heterogeneous (dissimilar) data; new clusters are added to the visualization.}
    \label{inc_vis}
\end{figure}

\IEEEPARstart{I}{n} data analysis today, we often encounter high-dimensional datasets with each dimension representing a variable or feature. When analysing such datasets, reducing the data dimensionality is highly useful to gain insights into the structure of the data. Visualization of high-dimensional data is achieved by reducing the data down to two or three dimensions. 

In practice, we often assume static data visualization, i.e., the data are presented to the dimensionality reduction methods all at once. However, with the advent of big data, the data may be presented incrementally for the following two main reasons. First, the dataset may be so large that it has to be divided and processed sequentially \cite{fisher2012trust}. Second, there are scenarios where data is incrementally acquired through a series of experiments, such as the continuous acquisition of Geographical Data \cite{montoya2003geo} or data gathered by mining social media \cite{zhang2008visualization}. In Fig. \ref{inc_vis}, we show how data can be augmented with either homogeneous data (new data with a structure similar to the already observed one) or heterogeneous data (new data with a different structure to the already observed one). In real-world situations, both scenarios may be present indistinguishably, and these necessitate incremental data visualization where we either directly use or continually train an already trained model to visualize the incoming data increments. In addition, it is often required that the visualization on existing data does not change drastically after the new data is added for consistency in interpreting the visualizations. 

Recently, t-distributed Stochastic Neighbor Embedding (t-SNE) \cite{maaten2008visualizing} and Uniform Manifold Approximation and Projection (UMAP) \cite{mcinnes2018umap} have shown success in nonlinear dimensionality reduction of static datasets. However, t-SNE and UMAP  are non-parametric by design and therefore have to be \textit{reinitialized} and \textit{retrained} in the presence of new data, thus not ideally suited for incremental data visualization. The parametric variant of t-SNE, called parametric t-SNE \cite{van2009learning}, may suit better for such scenarios, as it can be further trained with either the new data or a combination of old and new data without re-initialization. However, as we will show later, the visualization quality of parametric t-SNE is not as good as t-SNE for the datasets considered in our experiments.

In this work, we propose a new parametric dimensionality reduction method called Self-Organizing Nebulous Growths (SONG) for incremental data visualization. SONG combines the ability of t-SNE and UMAP to provide noise-tolerant, interpretable visualizations, with the ability of variants of a parametric method called Self-Organizing Map (SOM) \cite{kohonen1982self} to learn a parametric model for visualizing high-dimensional data. We show that SONG provides visualization quality comparable to or better than that of UMAP and t-SNE in static data scenarios while being significantly more efficient than the parametric t-SNE for incremental data visualization.

\section{Related Work} 
In static data visualization scenarios, t-SNE and UMAP are two state-of-the-art methods frequently used for dimensionality reduction. Both t-SNE and UMAP are inspired by Stochastic Neighbor Embedding (SNE) \cite{hinton2003stochastic} which consists of two main stages. First, a graph is created in the input space, where the vertices are the input points and the edge-weights represent the pseudo-probability of two inputs being in the same local neighborhood. Second, these pseudo-probabilities are preserved on an output graph of low-dimensionality (typically two or three). Due to the probabilistic nature, the edges in SNE are stochastic, thus different from the binary edges used in conventional methods such as Isomap \cite{tenenbaum2000global} and LLE \cite{roweis2000nonlinear} to represent the similarity between two connected points. As a result, SNE is more tolerant to outliers than LLE or Isomap. Additionally, the base distributions for converting the pair-wise distances to pseudo-probabilities in SNE can vary, allowing us to control the cluster tightness and separation in the visualizations. Similarly, by varying the loss function of SNE which calculates the discrepancy between the two sets of pseudo-probability distributions in the input and output spaces, we can control the granularity of the preserved topology.

UMAP and t-SNE differ in the selection of base distributions and loss functions. t-SNE calculates the edge-weights by assuming Gaussian distributions for the pairwise distances in the input space but Student’s t-distributions in the output space for a clear cluster separation. In contrast, UMAP assumes that the local neighborhoods lie on a Riemannian manifold and normalizes the local pairwise distances to obtain a fuzzy simplicial set that represents a weighted graph similar to that of t-SNE. UMAP then uses a suitable rational quadratic kernel function in the low-dimensional output space to approximate the edge probabilities of the weighted graph. For the loss function, t-SNE uses KL-Divergence while UMAP uses cross-entropy. As a result, UMAP provides higher separation of clusters and more consistent global topology preservation in visualization than t-SNE.

Due to its success in static data visualization, t-SNE has inspired several general-purpose visualization methods, including Trimap \cite{amid2019trimap} with better global topology preservation, and LargeVis \cite{tang2016visualizing} which is more efficient in large datasets than t-SNE. Also, t-SNE has been made more efficient in Barnes-Hut-SNE \cite{van2014accelerating}, extended to suit specific applications \cite{linderman2019fast, song2019improved}, and inspired other application-specific visualization methods such as viSNE \cite{amir2013visne} used on single-cell transcriptomic data. 
 
However, being non-parametric, t-SNE and its successors \cite{amid2019trimap, tang2016visualizing, van2014accelerating} as well as UMAP need to be \textit{reinitialized} and \textit{retrained} at each increment of data. In UMAP, the heuristic initialization using Spectral Embedding (Laplacian Eigenmaps) \cite{belkin2003laplacian} provides some degree of stability in visualizing datasets from the same distribution as similar datasets have similar nearest neighbor graphs that provide similar graph Laplacians. However, it was not previously investigated how well such heuristic initializations perform when the new data have a heterogeneous structure to the existing data.

One previously explored strategy for retaining a parametric mapping in t-SNE is to train a parametric regression model using the training data and the visualizations obtained by t-SNE. For example, Kernel t-SNE \cite{gisbrecht2015parametric} builds a linear model between the currently embedded vectors and the normalized Gram Matrix of the new inputs. However, the Gram matrix and linear weights do not adjust to new data, thus Kernel t-SNE may not be suitable for heterogeneous increments. On the other hand, parametric t-SNE \cite{van2009learning} uses a feed-forward neural network as the regression model which may be continually trained to adapt to the heterogeneous increments. Another strategy, used in the single-cell RNA-Seq data visualization tool scvis \cite{ding2018interpretable}, is to train a Variational Autoencoder (VAE) on the input data with the coding space limited to 2 or 3-dimensions for visualization. The VAE is regularized by a t-SNE objective for better cluster separation in the visualization. However, scvis has been shown to perform poorly at separating distinct clusters compared to t-SNE \cite{becht2019dimensionality}. Additionally, parametric t-SNE and scvis suffer from issues commonly associated with deep neural networks such as requiring a large amount of training data \cite{zhang2016understanding}, high computational complexity \cite{srivastava2015training}, and lack of model interpretability \cite{sturm2016interpretable}.

Self-Organizing Map (SOM) \cite{kohonen1982self} and its variants are arguably the only dimensionality reduction methods that retain an adjustable parametric graph on the input space to approximate the input data distribution locally. SOMs obtain the graph by vector encoding or vector quantization, i.e., SOMs partition the input space into Voronoi regions by mapping each input to the closest element in a set of representative vectors called coding vectors. The coding vectors, now representing the centroids of the Voronoi regions, are then mapped onto a low-dimensional uniform output grid. In practice, SOM implementations often use a 2 or 3-dimensional uniform grid, that has either a square, a triangular or a hexagonal topology for the locally connected output vectors. The topology preservation of SOMs is achieved by moving the coding vectors in the input space such that the coding vectors corresponding to neighbors in the low-dimensional output grid are placed close together.
 
There are two main problems associated with the visualizations provided by SOMs. First, the SOM visualizations have poor cluster separation possibly due to the uniform and fixed output grid \cite{fritzke1994growing}. SNE-based methods are not affected by this problem as their outputs are adjusted to reflect the topology of the high-dimensional space. Second, the size of the map (the number of coding vectors) needs to be known \textit{a priori}. Growing Cell Structures (GCS) \cite{fritzke1994growing} and its successor Growing Neural Gas (GNG) \cite{fritzke1995growing} use a non-uniform triangulation (where the coding vectors represent the vertices of the triangles) of the input space to tackle these two problems. GCS uses a force-directed graph drawing \cite{kamada1989algorithm} which can only visualize 2 or 3-dimensional graphs. Force-directed drawing algorithms such as `Spring' \cite{kamada1989algorithm} for graphs of arbitrary dimensionality have been successfully used in specific applications such as single-cell transcriptomic trajectory visualization \cite{weinreb2018spring}, where the graph is sparse, i.e., number of edges is small compared to the number of possible pairs of vertices. However, these algorithms are not suited for visualizing the coding vectors of GNG with less-sparse graphs of arbitrary dimensionality.

By using cross-entropy minimization similar to UMAP for visualizing the coding vectors, Neural Gas Cross-Entropy (NG-CE) \cite{estevez2005cross} overcomes the inability to embed less-sparse graphs shown by spring-like algorithms. However, NG-CE has not been extended to support the dynamically growing nature of GNG. Growing SOM (GSOM) \cite{alahakoon2000dynamic} overcomes SOM's deficiency of unknown map size by using a uniform but progressively growing output grid. However, similar to SOM, GSOM uses a uniform output grid. 

We propose Self-Organizing Nebulous Growths (SONG) which draws inspiration from SNE and NG-CE in using the discrepancy between input and output probability distributions to obtain a topology-preserving visualization, and GNG and GSOM in robust parametrization of the input topology with a growing network of coding vectors. The proposed method is described in the following section. 

\section{Method}

In the proposed SONG, we use a set of coding vectors $ \mathbf{C}= \{\mathbf{c}\in \mathbb{R}^D\} $ to partition and represent the input dataset $\mathbf{X} = \{\mathbf{x} \in \mathbb{R}^D \} $. For an input $\mathbf{x}_i \in \mathbf{X}$, we define an index set $\mathbf{I}^{(k)} = \{i_l | l = 1, ..., k\}$ for a user-defined $k$, where $\mathbf{c}_{i_l}$ is the $l$-th closest coding vector to $\mathbf{x}_i$. Moreover, we define a set of directional edges between the coding vectors $\mathbf{C}$, and an adjacency matrix $\mathbf{E}$, such that if a coding vector $\mathbf{c}_m$ is one of the closest neighbors to another coding vector $\mathbf{c}_l$, they are connected by an edge with edge strength $\mathbf{E}(l,m) > 0$. We organize the graph $\{\mathbf{C}, \mathbf{E}\}$ to approximate the input topology.  

We also define a set of low-dimensional vectors $\mathbf{Y}= \{\mathbf{y}\in\mathbb{R}^d \}$, $d << D$ which has a bijective correspondence with the set of coding vectors $\mathbf{C}$. When $d=2$ or $3$, $\mathbf{Y}$ represents the visualization of the input space, i.e., the input $\mathbf{x}_i \in \mathbf{X} $ is visualized as $\mathbf{y}_{i_1} \in \mathbf{Y}$. We preserve the topology of $\mathbf{C}$ given by $\mathbf{E}$ in $\mathbf{Y}$ by positioning $\mathbf{Y}$ such that, if $\mathbf{E}(l,m) > 0$ or $\mathbf{E}(m,l) > 0$ then $\mathbf{y}_l$ and $\mathbf{y}_m$ will be close to each other in the visualization. Typically, the number of $\mathbf{c}\in \mathbf{C}$ and corresponding low dimensional vectors $\mathbf{y} \in \mathbf{Y}$ is far less than the number of input data points $\mathbf{X}$. By retaining the parameters $\mathbf{C}$, $\mathbf{E}$ and $\mathbf{Y}$, SONG obtains a parametric mapping from input data to visualization. 

We initialize a SONG model by randomly placing $d+1$ coding vectors $\mathbf{C}$ in the input space, since $d+1$ is the minimum number of coding vectors needed to obtain a topology preserving visualization in a $d$-dimensional visualization space (see Supplement Section 1.1 for proof). No edge connection is assumed at initialization (i.e., $\mathbf{E} = \mathbf{0}$). The corresponding $\mathbf{Y}$ are also randomly placed in the $d$-dimensional output space. Next, we approximate local topology of any given $\mathbf{x} \in \mathbf{X}$ using $\mathbf{C}$, and  project this approximated topology to $\mathbf{Y}$ in the visualization space. To be specific, SONG randomly samples an input point $\mathbf{x}_i \in \mathbf{X}$ and performs the following steps at each iteration until terminated: 
\begin{enumerate} 

    \item \textit{Updating the Directional Edges in $\mathbf{E}$ between Coding Vectors $\mathbf{C}$ based on }$\mathbf{x}_i$: This step modifies the adjacency matrix $\mathbf{E}$ to add or remove the edges between coding vectors based on local density information at $\mathbf{x}_i$. Eventually, if no edge is added or removed during the repeated sampling of every input $\mathbf{x}_i$ of the dataset, the graph is considered stable and the training of SONG is terminated. We describe this step in detail in Section \ref{Graph_Const}. 
    
    \item \textit{Self-Organization of Coding Vectors } $\mathbf{C}$:
    This step moves $\mathbf{x}_i$'s closest coding vector $\mathbf{c}_{i_1}$ closer to $\mathbf{x}_i$, along with any coding vectors $\mathbf{c}_j$ if $\mathbf{c}_{i_1}$ and $\mathbf{c}_j$ are connected by an edge as indicated by $\mathbf{E}(i_1,j)>0$. This movement enforces the closeness of coding vectors connected by edges. We describe this in detail in Section \ref{VQ}. 
   
    \item \textit{Topology Preservation of the Low-dimensional Points } $\mathbf{Y}$: 
    Given that $\mathbf{c}_{i_1}$ encodes $\mathbf{x}_i$ and corresponds to $\mathbf{y}_{i_1}$ in the output space, we organize low-dimensional points $\mathbf{y}_j \in \mathbf{Y}$ in the locality of $\mathbf{y}_{i_1}$ such that the coding vector topology at $\mathbf{c}_{i_1}$ is preserved in the output space. This step is described in Section \ref{Dimensionality}.
     
    \item \textit{Growing $\mathbf{C}$ and $\mathbf{Y}$ to Refine the Inferred Topology}: There may be cases where $\mathbf{c}_{i_1}$ and its neighboring coding vectors are insufficient to capture the local fine topology at $\mathbf{x}_i$, e.g., inputs from multiple small clusters may have the same $\mathbf{c}_{i_1}$. In such cases, we place new coding vectors close to $\mathbf{c}_{i_1}$, and new corresponding low-dimensional vectors close to $\mathbf{y}_{i_1}$, without reinitializing the parametric model $\{ \mathbf{C}, \mathbf{Y},  \mathbf{E}\}$. Details of this are in Section \ref{MapGrowth}.
    
\end{enumerate}
For a given epoch, we randomly sample (without replacement) a new input $\mathbf{x}_i \in \mathbf{X}$ and repeat the four steps, until all $\mathbf{x_i} \in \mathbf{X}$ are sampled. The algorithm is terminated if the graph becomes stable in Step 1) or we have executed the maximum number of epochs. When new data $\mathbf{X}'$ are presented, we simply allow $\mathbf{x}_i$ to be sampled from $\mathbf{X}'$ at the next iteration, and continue training without reinitializing the parameters $\mathbf{C}$, $\mathbf{Y}$ and $\mathbf{E}$. 

\subsection{Updating the Directional Edges in $\mathbf{E}$ between Coding Vectors based on $\mathbf{x}_i$}
\label{Graph_Const}

For each input $\mathbf{x}_i$ randomly sampled from $\mathbf{X}$, we conduct three operations to any edge-strength $e_{i_1 j} = \mathbf{E}(i_1, j) $ at current iteration $t$:
\begin{itemize}
    \item \textit{Renewal}: we reset $e_{i_1 j}^t$ to 1 if $ j \in \mathbf{I}^{(k)}$. 
    \item \textit{Decay}: if $e_{i_1 j}^{t-1}>0$, we decay it by a constant multiplier $\epsilon \in (0, 1)$, i.e., $e_{i_1 j}^t=e_{i_1 j}^{t-1}\cdot\epsilon$. Note that for edges with strengths $e_{i_1 j}^{t-1} >0$ for $j \notin \mathbf{I}^{(k)}$ at current iteration $t$, this decay operation would weaken such edges repeatedly. Therefore, edges created at an earlier stage of training that no longer connects two close coding vectors will be weakened.
    \item \textit{Pruning}: we set $e_{i_1 j}^t$ to 0 if $e_{i_1 j}^{t-1}<e_{\text{min}}$. This helps to obtain a sparse graph by removing edges that connect distant coding vectors as such edges would be weakened due to a lack of frequent renewal and frequent decay. Note that $e_{\text{min}}$ is predefined to obtain the desired degree of sparseness in the graph. 
\end{itemize}

The edge strength $e_{i_1 j} \in [0, 1]$ reflects the rate the edge is renewed at and is proportional to $p_{i_1 j}$, which is the probability of $i_1$ and $j$ being close neighbors to the input $\mathbf{x}_i$. Considering this proportionality, larger edge strengths can be interpreted as $\mathbf{c}_{i_1}$ and neighboring coding vectors representing finer topologies (shorter distances between $\mathbf{C}$). Conversely, the smaller the edge strengths, the coarser the topology represented by such edges. Note that here we define finer and coarser topologies in their conventional sense \cite{munkres2014topology}.

Note that edge strength $e_{i_1 j}$ obtained above is directional and thus the adjacency matrix $\mathbf{E}$ is asymmetric. We observe faster convergence in subsequent optimization with a symmetric adjacency matrix, which is simply calculated as:
\begin{equation}\label{eq:symmetric_adjacency}
    \mathbf{E}_s = \frac{\mathbf{E} + \mathbf{E}^T}{ 2 }
\end{equation}

Next, we use the coding vector graph to approximate the topology of the input through self-organization.

\subsection{Self-Organization of Coding Vectors $\mathbf{C}$}
\label{VQ}
To ensure the coding vectors $\mathbf{C}$ are located at the centers of input regions with high probability densities (such as cluster centers), we move the coding vectors $\mathbf{c}_{i_1}$ towards $\mathbf{x}_i$ by a small amount to minimize the following Quantization Error (QE): 
\begin{equation}
    QE(x) = \frac{1}{2} \Vert \mathbf{x}_i - \mathbf{c}_{i_1} \Vert ^ 2
\end{equation}
However, moving $\mathbf{c}_{i_1}$ independent of other coding vectors may cause the coding vectors sharing an edge with $\mathbf{c}_{i_1}$ to be no longer close to $\mathbf{c}_{i_1}$, which disorganizes the graph. To avoid this, we also move $\mathbf{c}_{i_1}$'s neighboring coding vectors $\mathbf{c}_{j}$ (as indicated by $\mathbf{E}_s(i_1, j) > 0$) towards $\mathbf{x}_i$ by a smaller amount than that of $\mathbf{c}_{i_1}$. Moreover, the more distant $\mathbf{c}_{j}$ is from $\mathbf{c}_{i_1}$, the smaller the movement of $\mathbf{c}_{j}$ should be. This ensures that the organization of distant neighbors is proportionately preserved by this movement. Therefore, we define a loss function that monotonically decreases when the distance from the coding vectors to $\mathbf{x}_i$ increases. In addition, to penalize large neighborhoods (and thereby large edge lengths), we scale the loss function by a constant specific to the neighborhood $\mathbf{I}^{(k)}$, and in this work, we select the square of the largest distance from $\mathbf{x}_i$, i.e., $\Vert \mathbf{x}_i - \mathbf{c}_{i_k} \Vert^2$ as this constant. The final loss function is:
\begin{equation}
    \mathscr{L} (\mathbf{x}_i) = - \frac{\Vert \mathbf{x}_i - \mathbf{c}_{i_k}\Vert^2 }{2} \sum_{\mathbf{c}_j \in \mathscr{N}_{i_1}} \exp({- \frac{\Vert \mathbf{x}_i - \mathbf{c}_j \Vert ^ 2}{\Vert \mathbf{x}_i - \mathbf{c}_{i_k}\Vert^2} })
\end{equation}
where $\mathscr{N}_{i_1} = \{\mathbf{c}_j \mid \mathbf{E}_s(j, i_1) > 0 \}$ is the set of $\mathbf{c}_{i_1}$'s neighboring coding vectors. Note that we treat the distance to the $k^{th}$ coding vector from $x$ as a constant for the considered neighborhood, and the gradient of $\Vert \mathbf{x}_i - \mathbf{c}_{i_k} \Vert$ w.r.t. $\mathbf{c}_{i_k}$ is not calculated. Using stochastic gradient descent to minimize this loss, we calculate the partial derivatives of the loss w.r.t. a given $\mathbf{c}_j$ for the sampled $\mathbf{x}_i$ as: 
\begin{equation}
    \frac{\partial \mathscr{L}(\mathbf{x}_i)}{\partial \mathbf{c}_j} = (\mathbf{x}_i-\mathbf{c}_j) \times \exp{(-\frac{\Vert \mathbf{x}_i - \mathbf{c}_j \Vert^2}{\Vert \mathbf{x}_i - \mathbf{c}_{i_k}\Vert^2})}
\end{equation}

Next, we describe how we optimize the output embedding (the placement of $\mathbf{Y}$) to reflect the topology inferred in the input space. 

\subsection{Topology Preservation of the Low-dimensional Points $\mathbf{Y}$}
\label{Dimensionality}

Similar to UMAP and inspired by NG-CE, SONG optimizes the embedding $\mathbf{Y}$ by minimizing the Cross Entropy ($CE$) between the probability distribution $p$ in the input space and a predefined low-dimensional probability distribution $q$ in the output space. We define the local cross entropy for a given $\mathbf{x}_i \in \mathbf{X}$ as: 
\begin{equation}
\label{eq: ce}
    CE(\mathbf{x}_i) =  \sum_{\forall j} - p_{i_1 j} \log(q_{i_1 j}) - (1 - p_{i_1 j}) \log(1 - q_{i_1 j})
\end{equation}
where $p_{i_1 j}$ is the probability that coding vectors $\mathbf{c}_{i_1}$ and $\mathbf{c}_j$ are located close to each other in the input space. Its estimation will be provided shortly. Similarly, $q_{i_1 j}$ is the probability that output points $\mathbf{y}_{i_1}$ and $\mathbf{y}_j$ are located close together, and is calculated using the following rational quadratic function:
\begin{equation}
\label{eq:qij}
    q_{i_1 j} = \frac{1}{1 + a\Vert \mathbf{y}_{i_1} - \mathbf{y}_j \Vert ^ {2b}}
\end{equation}
See Supplement Section 2 for how to calculate the hyper-parameters $a$ and $b$. 

The cross entropy (see Eq.~\ref{eq: ce}) can be interpreted as two sub-components: an attraction component $CE_{\text{attr}} = \sum -p_{i_1 j} \log (q_{i_1 j})$ that attracts $\mathbf{y}_{i_1}$ towards $\mathbf{y}_{j}$, and a repulsion component $CE_{\text{rep}}= \sum -(1-p_{i_1 j} )\log (1-q_{i_1 j})$ that repulses $\mathbf{y}_{i_1}$ from $\mathbf{y}_{j}$. Since distant $\mathbf{y}_{j}$ results in $p_{i_1 j}=0$, $CE_{\text{attr}}$ heavily influences the local arrangement at $\mathbf{y}_{i_1}$ and conversely, $CE_{\text{rep}}$ influences the global arrangement of neighborhoods.  Due to the difference of influences, we derive the gradients of these two components separately. The gradients for the attraction and repulsion components are given by Equations \ref{eqn:attr_grad} and \ref{eqn:rep_grad} respectively: 
\begin{equation}
\label{eqn:attr_grad}
    \frac{\partial CE_{\text{attr}}}{\partial \mathbf{y}} = (\mathbf{y}_{i_1} - \mathbf{y}_j) \cdot \frac{2ab\cdot p_{ij}\cdot \Vert \mathbf{y}_j - \mathbf{y}_{i_1} \Vert ^{2b-2}}{1+\Vert \mathbf{y}_j - \mathbf{y}_{i_1} \Vert ^{2b}}
\end{equation}

\begin{equation}
\label{eqn:rep_grad}
    \frac{\partial CE_{\text{rep}}}{\partial \mathbf{y}} = (\mathbf{y}_{i_1} - \mathbf{y}_j) \cdot \frac{2b \cdot (1- p_{ij})}{\Vert \mathbf{y}_j - \mathbf{y}_{i_1} \Vert^2  (1+\Vert \mathbf{y}_j - \mathbf{y}_{i_1} \Vert ^{2b})} 
\end{equation}

We use stochastic gradient descent to minimize $CE_{\text{attr}}$ and $CE_{\text{rep}}$ w.r.t $\mathbf{y}_{i_1}, \mathbf{y}_j \in \mathbf{Y}$. We select stochastic gradient descent over batch gradient descent to avoid convergence on sub-optimal organizations \cite{fort2002advantages}. Moreover, since the edge renewal rate is proportional to the $p_{i_1 j}$, we propose to use the symmetric edge strengths $\hat{e}_{i_1 j}\in E_s$ as an approximation of $p_{i_1 j}$ in $CE_{\text{attr}}$. This avoids the explicit assumptions on $p_{i_1 j}$ as made by t-SNE and UMAP. In a similar fashion, we use the negative sampling of edges (i.e., sampling of $j$  such that $\mathbf{E}_s(i_1,j) = 0$) to approximate $(1-p_{i_1 j})$ in $CE_{\text{rep}}$. Specifically, for a very large dataset, we randomly sample a set of non-edges, such that for each $\mathbf{x}_i$, the number of sampled non-edges $n_{ns}$ equals the number of edges connected to $\mathbf{c}_{i_1}$ ($n(e_i)$) multiplied by a constant rate. 
Similar ideas have been used in Word2Vec \cite{mikolov2013distributed} and UMAP \cite{mcinnes2018umap}. The algorithmic summary of this step is shown in Algorithm \ref{alg:dr}.

\subsection{Growing $\mathbf{C}$ and $\mathbf{Y}$ to Refine the Inferred Topology}
\label{MapGrowth}

By iterating the above three steps from Sections \ref{Graph_Const} - \ref{Dimensionality}, the topology of $\mathbf{x}$ can be approximated using $\{\mathbf{C}, \mathbf{E} \}$ and preserved onto $\mathbf{Y}$ in the visualization space. However, since the optimal number of coding vectors $\mathbf{C}$ is unknown \textit{a priori}, SONG starts with a small number of $\mathbf{C}$ which may be insufficient to capture all the structures in $\mathbf{X}$ such as clusters and sub-clusters. Therefore, we grow the sizes of $\mathbf{C}$ and $\mathbf{Y}$ as needed during training.  Additionally, such growth can accommodate structural changes, e.g., addition of new clusters, when new data are presented in the incremental data visualization scenarios. 

Inspired by the GNG, we define a Growth Error associated with  $\mathbf{c}_{i_1}$ as: 
\begin{equation}
    G_{i_1}(t) \leftarrow G_{i_1}(t-1) + \Vert \mathbf{x}_i - \mathbf{c}_{i_1} \Vert
\end{equation}
where $t$ is the index of current iteration. When any $G_{i_1}(t)$ exceeds a predefined threshold $\theta_g$, we place a new coding vector $\mathbf{c}$ at the centroid between $\mathbf{x}_i$ and its $k$ nearest coding vectors, so that the regions that have high Growth Error get more populated with coding vectors. In Supplement Section 1.2, we describe how we have calculated the $\theta_g$ from a hyperparameter called the Spread Factor ($SF$) as defined in \cite{alahakoon2000dynamic, chan2009investigation}. Due to the stochastic sampling, the current $\mathbf{x}_i$ and its neighboring data may not be sampled in the next iterations, thus the newly created coding vector may not be duly connected in subsequent repetitions of Step 1 and it may eventually drift away. To avoid this, at the current iteration, we add new edges from the newly added coding vectors to all neighbors of the $\mathbf{c}_{i_1}$. Similarly, for faster convergence of the output visualization, rather than placing new $\mathbf{y}$ at random, we place it in the close neighborhood of $\mathbf{y}_{i_1}$. We summarize this step in Algorithm \ref{alg:growth}. 

These four steps form a complete iteration of the SONG algorithm, which we summarize in Algorithm \ref{alg:song}. In the next section, we evaluate the performance of the SONG algorithm. 

\begin{algorithm}[ht]

\algsetup{linenosize=\tiny}
\SetAlgoLined

 $t \leftarrow \text{Iteration index, initialized as 0}$\;
 $t_{max} \leftarrow \text{Maximum number of iterations}$ \;
 $d \leftarrow \text{Output dimensionality; usually 2 or 3}$\;
 $k \leftarrow $ Number of neighbors to consider at a given locality, $k \geq d + 1$\;
 $\alpha \leftarrow $ Learning rate starting at $\alpha_0$\;
 $\mathbf{C} \leftarrow \text{Random matrix of size } (d+1) \times D$\;
 $E \leftarrow $ Edges on C, from each c to other cs, all initialized as non-edges (0)\;
 $\mathbf{Y} \leftarrow \text{Random matrix of size} (d+1) \times d$\;
 $r \gets $ Number of negative edges to select, per positive edge for negative sampling\;
 $ a, b \leftarrow $ Appropriate parameters to get desired  spread and tightness as per Eq. \ref{eq:qij}\;
 
 $ \theta_g \leftarrow $ User defined growth threshold\;
 
 \While{$t < t_{max}$}{

    \For{ $ x_i \in X$} {
        Update $E_s$ as per Section \ref{Graph_Const}\;
        Update $\mathbf{I}^{(k)}$\;
        $ns = r\cdot n(\hat{e}_{i_1})$, here $n(\hat{e}_{i_1})$ is the number of edges from or to $\mathbf{c}_{i_1}$\;
        Record the neighbors of $\mathbf{c}_{i_1}$ as $\mathscr{N}_{i_1}^{t-1}  \gets \{ j \mid E_s(i_1, j) > 0 \} $ \;
        Perform Edge Curation as per Algorithm \ref{edge_cur} \;
        
        Record the new set of neighbors $\mathscr{N}^t_{i_1}$\;
        \If{ $\mathscr{N}^{t-1}_{i_1} == \mathscr{N}^t_{i_1}$ } {End the execution of the algorithm and \Return \;}

        Perform Self-Organization of Coding Vectors $\mathbf{C}$:
        
        \For{${j} \in \mathscr{N}^{t}_{i_1}$}{
        $\mathbf{c}_{j}\leftarrow \mathbf{c}_j + \alpha\cdot \frac{\partial \mathscr{L}(\mathbf{x}_i)}{\partial \mathbf{c}_j} $ 
        }
        Update $\mathbf{Y}$ as per Algorithm \ref{alg:dr}\;
        
        $G_{i_1} \gets G_{i_1} + \Vert \mathbf{x}_i - \mathbf{c}_{i_1}\Vert$\;
        
        \If{$G_{i_1} > \theta_g$}{
            Grow $\mathbf{C}$ and $\mathbf{Y}$ as per Algorithm \ref{alg:growth}\;
        }
    }
  
  $t \leftarrow t + 1 $\;
  $ \alpha \leftarrow \alpha_0 \times (1-\frac{t}{t_{max}})$\;
  
  }

\caption{\label{alg:song}SONG Algorithm with Decaying Learning Rate}
\end{algorithm}

\begin{algorithm}[ht]
\SetAlgoLined
\caption{Updating the Directional Edges in $\mathbf{E}$ between Coding Vectors based on $\mathbf{x}_i$ \label{edge_cur}}
\For{ $j \in \mathbf{I}^{(k)} $}{
            \uIf{ $ \Vert \mathbf{x}_i - \mathbf{c}_j \Vert  \leq \Vert \mathbf{x}_i - \mathbf{c}_k \Vert $}{
               Renew edges as $E(i_1,j) = 1$\;
            }\Else{
           Decay edges as $ E(i_1,j) \gets \epsilon \cdot E(i_1,j)$\;
           }
            \If{$E(i_1,j) < e_{\text{min}}$}{
            Prune edges as $E(i_1,j) \gets 0$\;
            }
        }
\end{algorithm}

\begin{algorithm}[ht]
\SetAlgoLined

/* Organization of Local Neighborhood */\;
\For{${j} \in \mathscr{N}^t_{i_1}$}{
    $\mathbf{y}_{j} \leftarrow \mathbf{y}_j + \alpha\cdot (\mathbf{y}_{i_1} - \mathbf{y}_j) \cdot \frac{2ab\cdot \hat{e}_{i_1 j} \cdot \Vert \mathbf{y}_j - \mathbf{y}_{i_1} \Vert ^{2b-2}}{1+\Vert \mathbf{y}_j - \mathbf{y}_{i_1} \Vert ^{2b}} $ 
}
/* Negative Sampling for Repulsion */\;
Select $ns$ random samples $J = \{j_1$, ... ,$j_{ns}\}$ with $E_s(i_1, j) = 0$ \;
\For{$j \in J$}{
    $\mathbf{y}_{j} \leftarrow \mathbf{y}_j - \alpha\cdot (\mathbf{y}_{i_1} - \mathbf{y}_j) \cdot \frac{2b}{\Vert \mathbf{y}_j - \mathbf{y}_{i_1} \Vert^2  (1+\Vert \mathbf{y}_j - \mathbf{y}_{i_1} \Vert ^{2b})} $ 
}
\caption{Topology Preservation of the Low-dimensional Points $\mathbf{Y}$ \label{alg:dr}}
\end{algorithm}

\begin{algorithm}[ht]
\SetAlgoLined
create new coding vector such that \quad
$\mathbf{w}_{{n}} \leftarrow \frac{1}{k} \underset{l = \{1 ... k \}}{\sum}\mathbf{w}_{i_l}$ \;

create new low-dimensional vector such that 
$\mathbf{y}_{{n}} \leftarrow \frac{1}{k} \underset{l = \{1 ... k \}}{\sum}\mathbf{y}_{i_l}$ \;

\For{$j \in \{ i_1, ... , i_k \}$}{
    $E(j, n) = 1 $\; 
}
\caption{Growing $\mathbf{C}$ and $\mathbf{Y}$ to Refine the Inferred Topology \label{alg:growth}}
\end{algorithm}

\section{Experiments and Results}

In this section, we compare SONG against Parametric t-SNE and non-parametric methods t-SNE \cite{maaten2008visualizing} and UMAP \cite{mcinnes2018umap} on a series of data visualization tasks. First, we consider incremental data visualization with heterogeneous increments of data in Section \ref{biased_exp} and homogeneous increments of data in Section \ref{unbiased_exp}. It is noteworthy that the former is more appropriate to be assumed as the case in real problems with incremental data streams lacking \textit{a priori} ground-truth. Furthermore, we evaluate the visualization quality of SONG in static data visualization scenarios: we assess the SONG's robustness to noisy and highly mixed clusters in Section \ref{noise_exp}, and SONG's capability in preservation of topologies in Section \ref{tp_exp}.

In our analysis, we define model-retaining methods as methods that reuse a pretrained model and refine it when presented with new data, while model-reinitializing methods reinitialize a model and retrain it from scratch. Therefore, we consider SONG and Parametric t-SNE as model-retaining methods, and t-SNE and UMAP as model-reinitializing methods. However, for a fair comparison, we introduce a model-reinitializing version of SONG called SONG + Reinit. The ``incremental visualizations" can only be fairly assessed with model-retaining methods, but for the sake of completeness, we extend this comparison to the model-reinitializing methods as well.

We conducted a study on SONG's sensitivity to hyper-parameters in our Supplement Section 3.3, where we identified four key hyper-parameters that affect the graph inference (number of coding vectors in a neighborhood: $k$, edge-strength decay rate: $\epsilon$, initial learning rate: $\alpha_0$ and Spread Factor: $SF$ as defined in \cite{alahakoon2000dynamic} and \cite{chan2009investigation}), and two parameters that affect the projection ($a$ and $b$). We found that higher spread-factors and lower $k$ values preserve a finer topology. Therefore for a more faithful representation of high dimensional data, we recommend a high Spread Factor ($SF>0.9$) and a small number of coding vectors in a neighborhood ($k<5)$. The hyperparameters for SONG used in the experiments in this work are obtained through these observations. 

We use the hyper-parameters in Table~\ref{table:hyperparams} for each method. For t-SNE \cite{maaten2008visualizing} and UMAP \cite{mcinnes2018umap}, the recommended hyper-parameters in the original papers were used as we did not observe any improvement in results by tuning these parameters. Similarly, for parametric t-SNE, we used the set of parameters provided by the GitHub implementation \footnote{https://github.com/jsilter/parametric\_tsne}.

\begin{table}[t]
\centering
\caption{\label{table:hyperparams}
Hyperparameters used for SONG, Parametric t-SNE, t-SNE and UMAP throughout our experiments}

\begin{tabular}{p{3cm} p{4.5cm}}
\toprule
Algorithm & Hyperparameters \\ 
\midrule
SONG/SONG-Reinit & $k=2$,$\epsilon=0.99$, $t_{\text{max}} = 100$, $\alpha_0 = 1.0$, $a = 1.577$ ,  $b =0.895$\\ 
Parametric t-SNE & Perplexity = 30, epochs = 400,
batch\_size = 128 \\
t-SNE & Perplexity = 30 \\
UMAP & n\_neighbors $= 15$, , $\alpha_0 = 1.0$,  $a = 1.577$ ,  $b =0.895$ \\ 
\bottomrule
\end{tabular}

\end{table}

\subsection{Visualization of Data with Heterogenous Increments}
\label{biased_exp}

We first evaluate SONG presented with heterogeneous increments, where new clusters or classes may be added to the existing datasets. 

\textbf{Setup}: Three datasets are used: Wong \cite{wong2016high}, MNIST \cite{lecun2010mnist} and Fashion MNIST \cite{xiao2017fashion}. The Wong dataset has over 327k single human T-cells measured for expression levels for 39 different surface markers (i.e., 39 dimensions) such as the CCR7 surface marker. There are many types of cells present in this dataset, such as lymphoid cells, naive T-cells, B-Cell Follides, NK T cells etc, which we expect to be clustered separately. However, there may be some cell types that cannot be clearly separated as clusters in visualizations \cite{becht2019dimensionality}.
Since we have no ground-truth labels and UMAP provides superior qualitative cluster separation on this dataset \cite{becht2019dimensionality}, we assume that the well separated clusters visible in the UMAP visualization of the dataset represent different cell types. This assumption allows us to initially sample 20k cells, and increment this sample to 50k, 100k and 327k cells such that at each increment we add one or several cell types to the data. However, due to the lack of ground truth, we can only conduct qualitative analysis on the cluster quality for each method. In addition, we conduct ``logicle transformation" \cite{parks2006new} to normalize the Wong dataset as a preprocessing step. On the other hand, MNIST dataset is a collection of 60k images of hand written digits, each image having $28\times28$ pixels, therefore 784 pixel intensity levels (dimensions), and an associated label from 0 to 9. Similar to MNIST dataset, Fashion MNIST dataset is a collection of 60k images of fashion items belonging to 10 classes, each with 784 pixels and a known ground-truth label. Since both MNIST and Fashion MNIST datasets have known ground-truth labels, we start with two randomly selected classes and present two more classes to the algorithm at each increment. We ran a K-Means clustering on the visualizations provided for MNIST and Fashion MNIST by each visualization method compared, and calculated the Adjusted Mutual Information (AMI) \cite{vinh2009information} scores against the ground-truth labels. The AMI scores were averaged over five iterations with random initializations.Additionally, we assume that the first 20 principle components capture most of the variance in the datasets \cite{cao2003comparison}. Therefore, both MNIST and Fashion MNIST datasets are reduced to 20 dimensions each using Principal Component Analysis (PCA) as a preprocessing step in order to reduce the running time of our experiments. Each of the intermediate and incrementally growing datasets of Wong, MNIST and Fashion MNIST datasets is visualized using SONG, SONG + Reinit, Parametric t-SNE, t-SNE, and UMAP.

\begin{figure*}
    \centering
    \begin{tabular}{l c c c c}
     & 20000 & 50000 & 100000 & 327000 \\
\begin{turn}{90}SONG\end{turn} & \includegraphics[width=0.2\linewidth, trim= {1.1cm, 0.8cm, 0cm, 0cm}, clip]{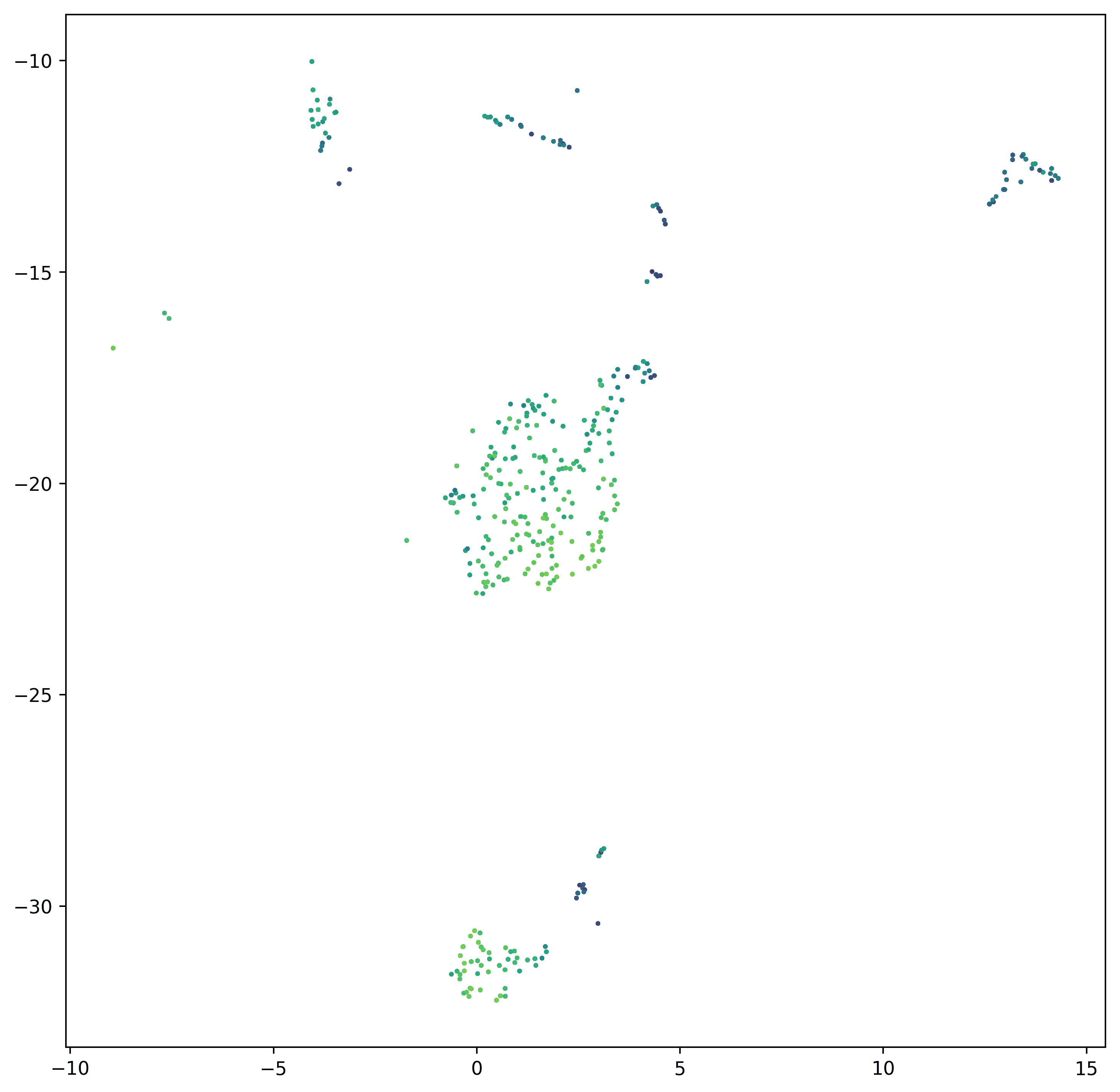} & \includegraphics[width=0.2\linewidth, trim= {1.1cm, 0.8cm, 0cm, 0cm}, clip]{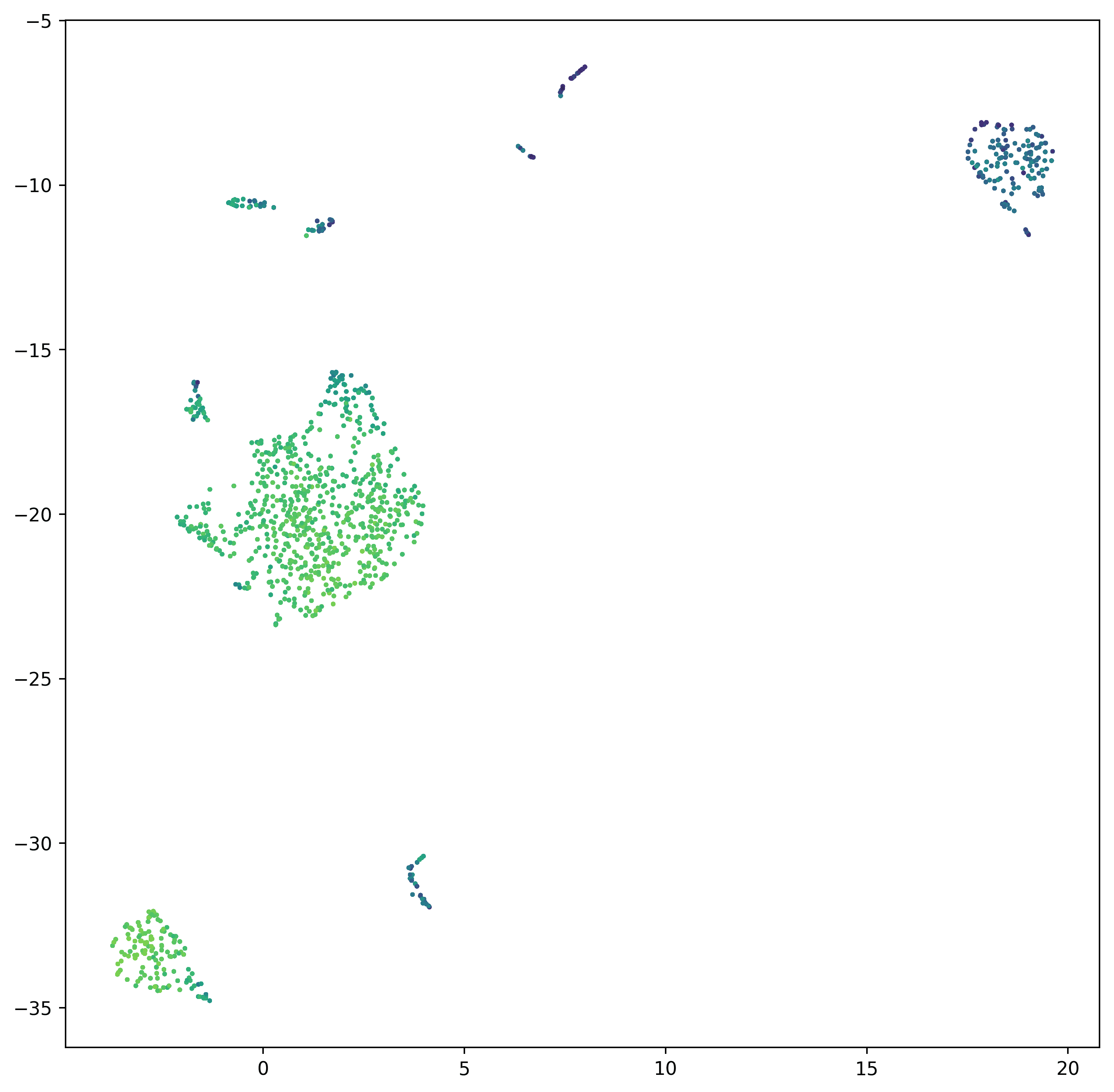} & \includegraphics[width=0.2\linewidth, trim= {1.1cm, 0.8cm, 0cm, 0cm}, clip]{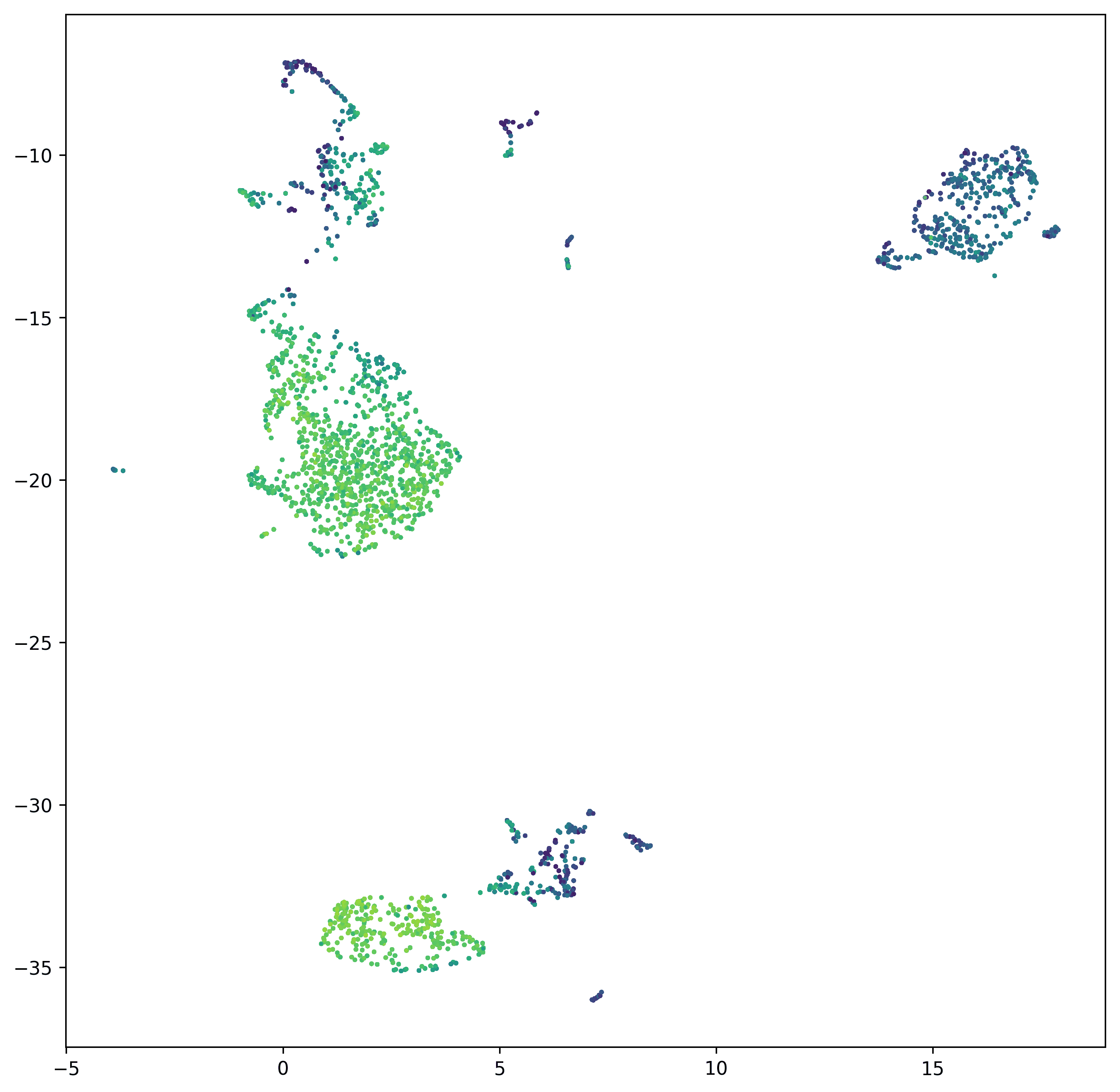}&\includegraphics[width=0.2\linewidth, trim= {1.1cm, 0.8cm, 0cm, 0cm}, clip]{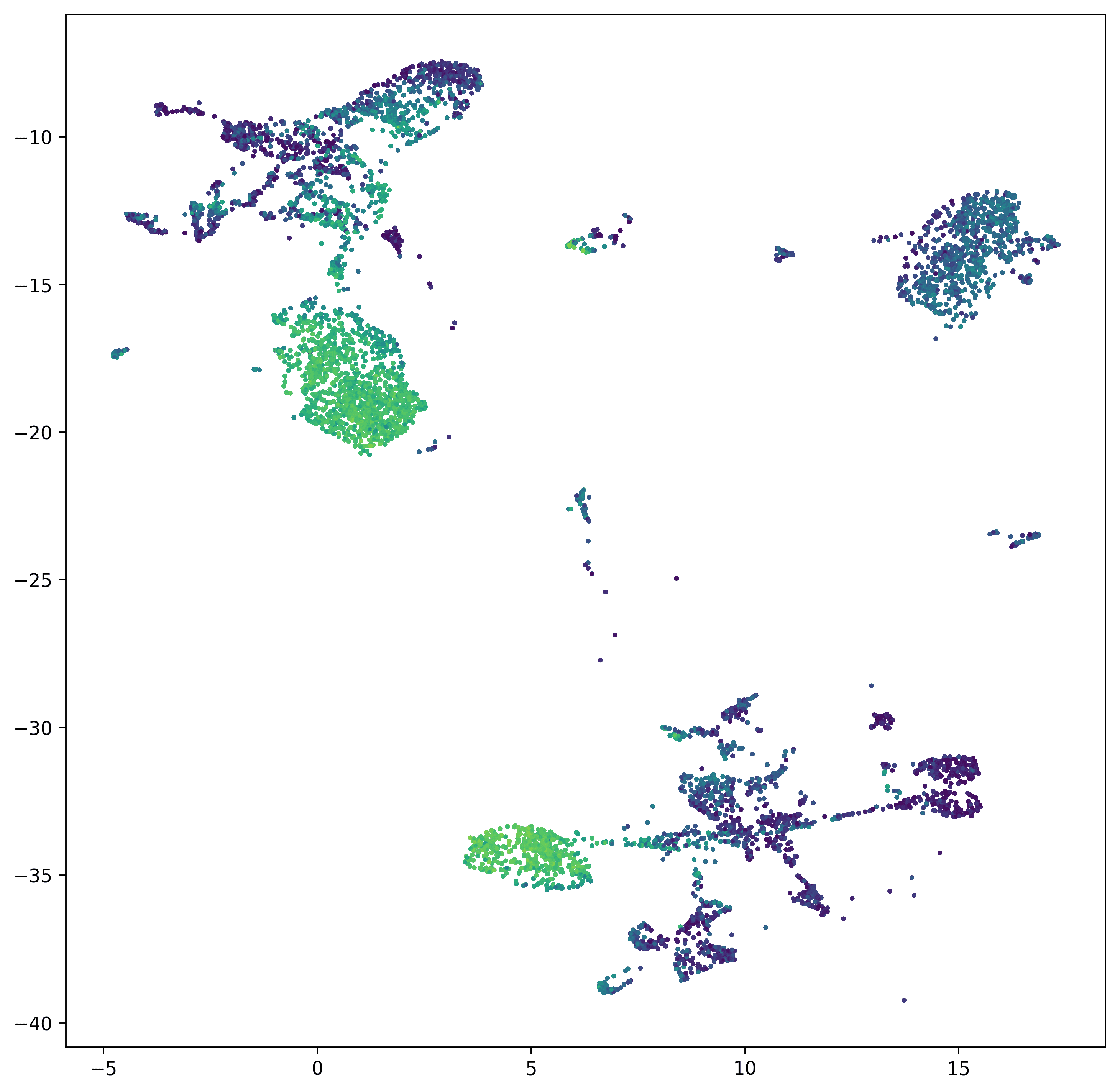}  \\

\begin{turn}{90} SONG + Reinit \end{turn} & \includegraphics[width=0.2\linewidth, trim= {1.1cm, 0.8cm, 0cm, 0cm}, clip]{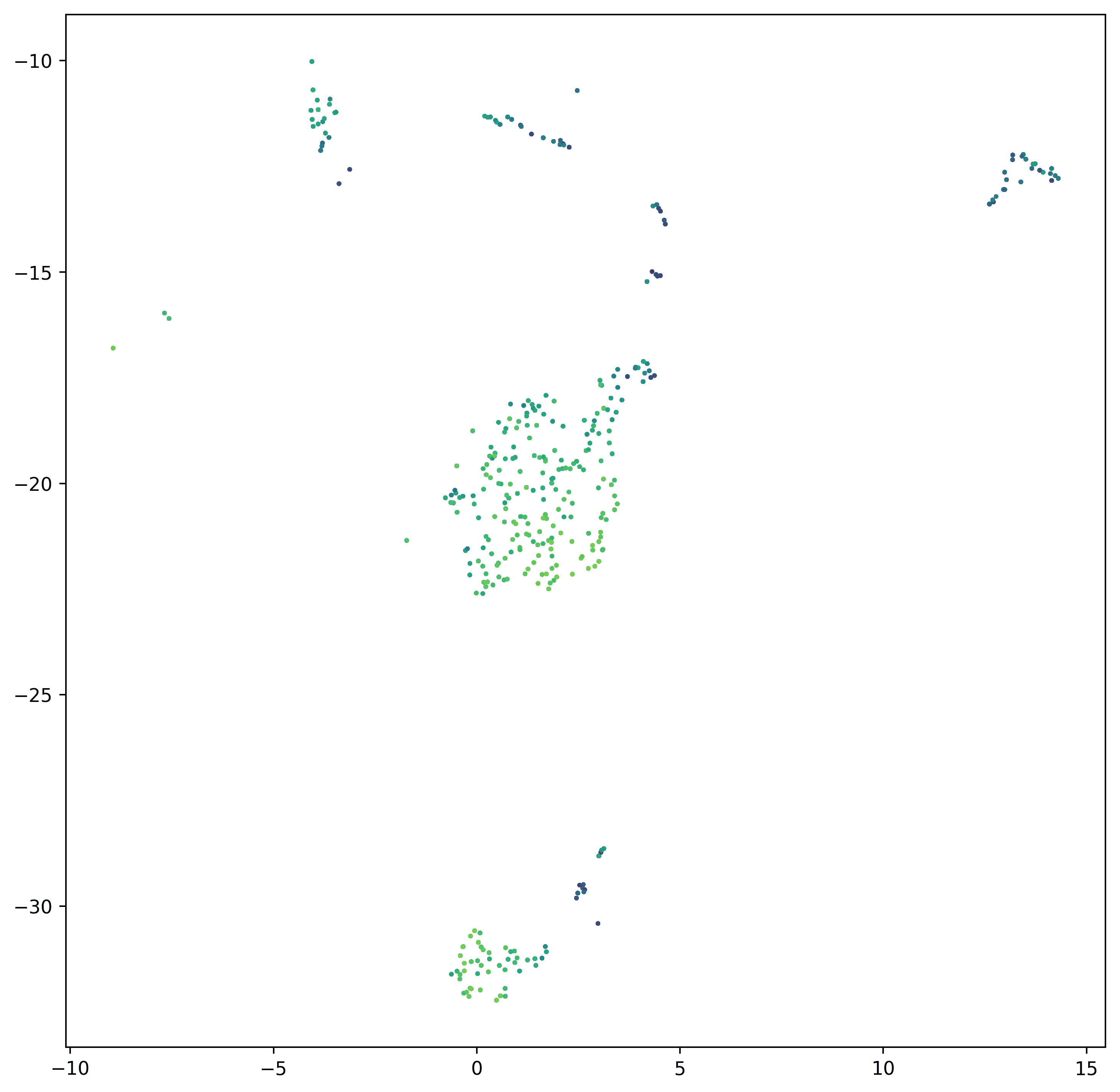} & \includegraphics[width=0.2\linewidth, trim= {1.1cm, 0.8cm, 0cm, 0cm}, clip]{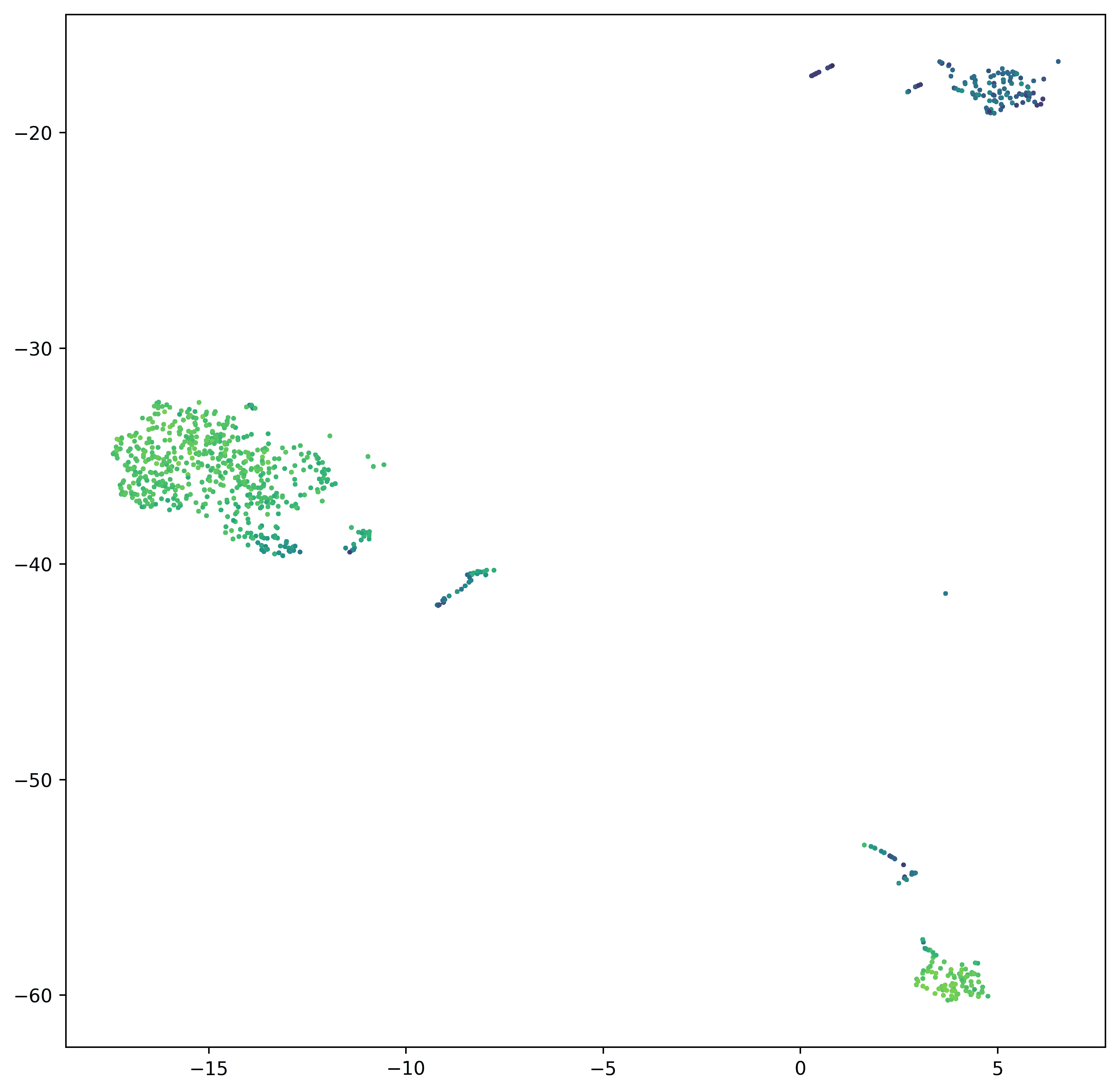} & \includegraphics[width=0.2\linewidth, trim= {1.1cm, 0.8cm, 0cm, 0cm}, clip]{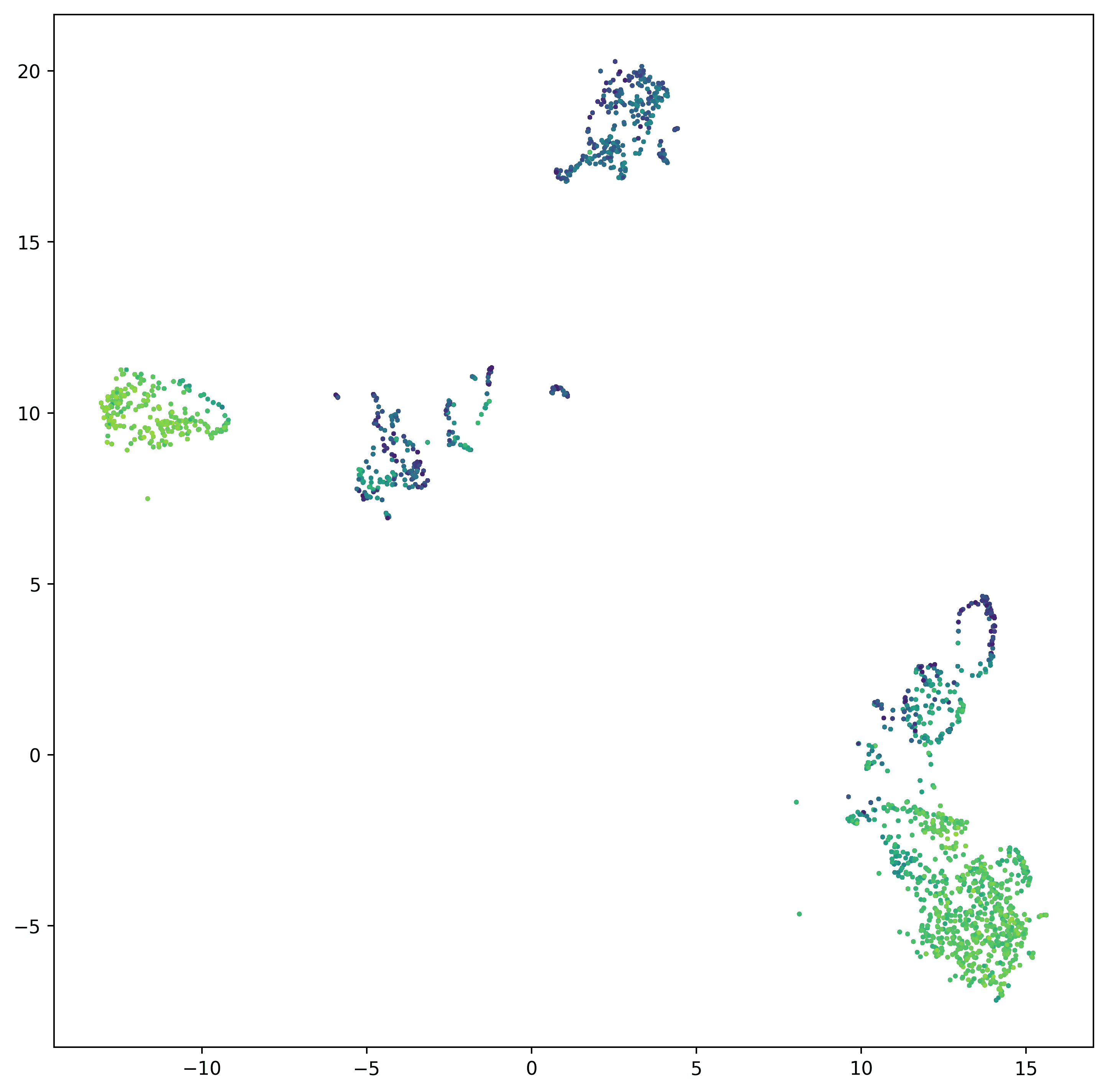} & \includegraphics[width=0.2\linewidth, trim= {1.1cm, 0.8cm, 0cm, 0cm}, clip]{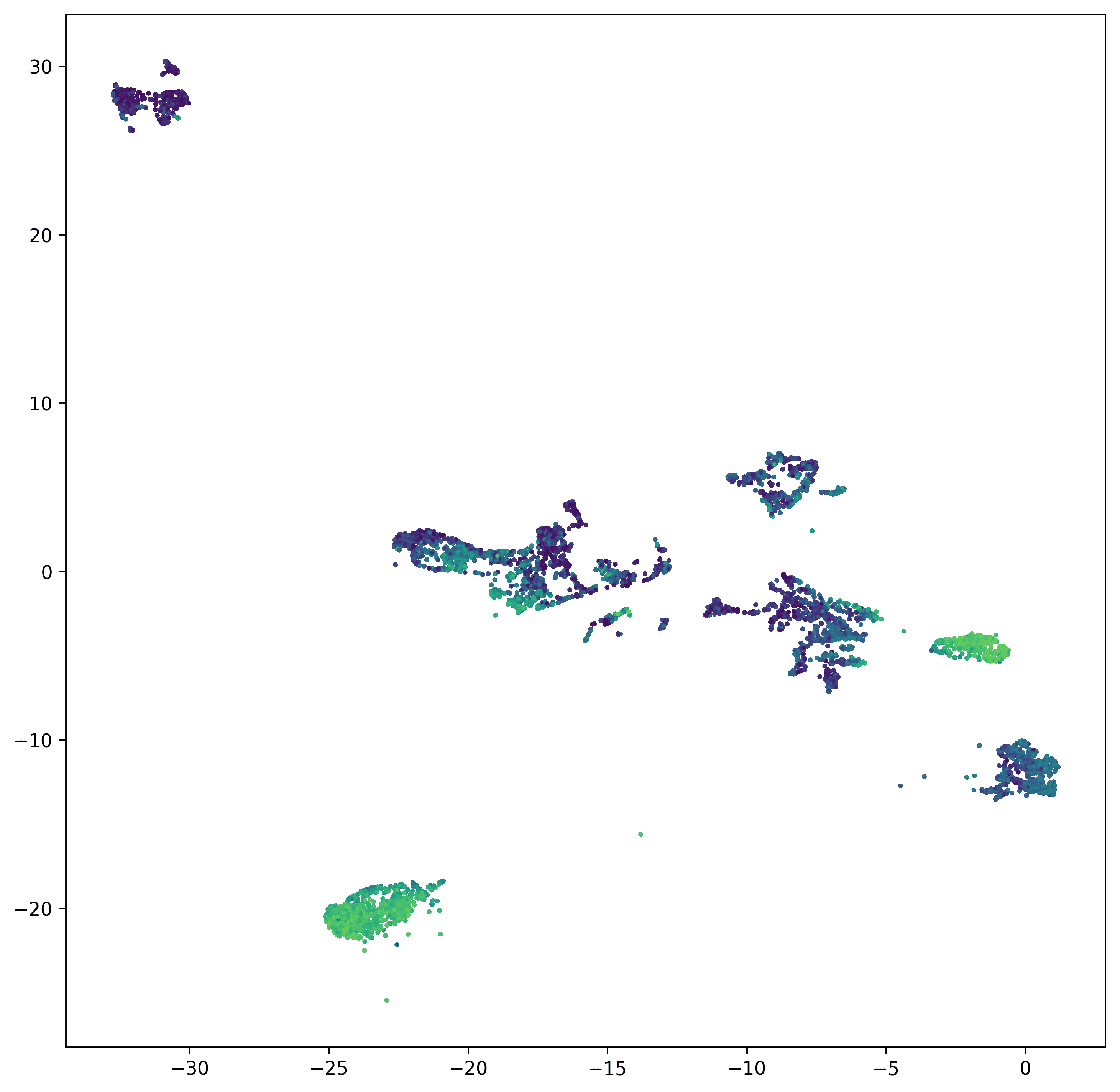}  \\

\begin{turn}{90}Parametric t-SNE \end{turn} & \includegraphics[width=0.2\linewidth, trim= {1.1cm, 0.8cm, 0cm, 0cm}, clip]{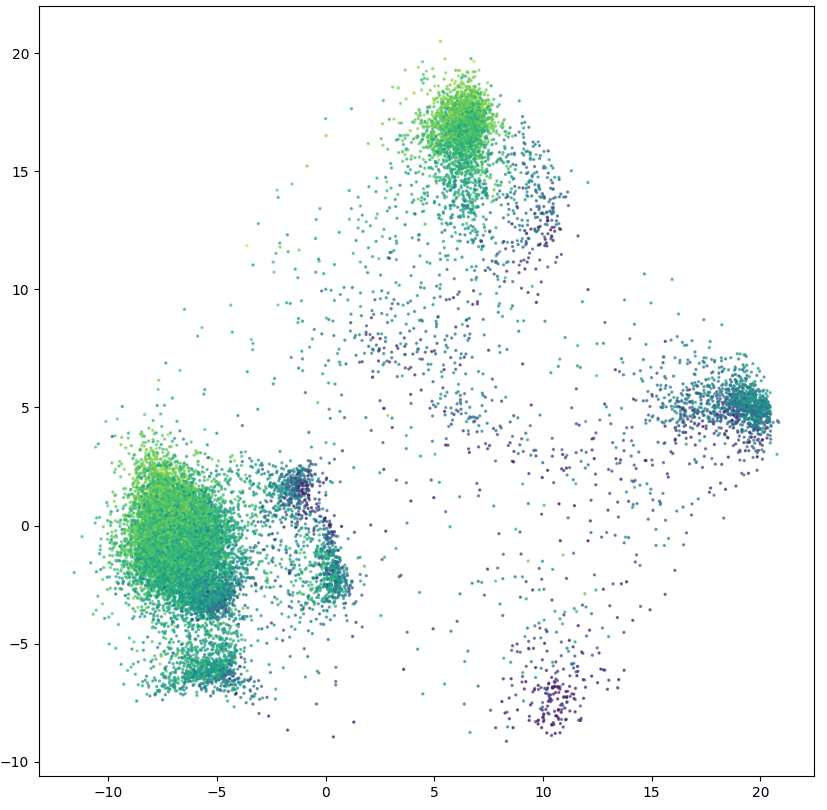} & \includegraphics[width=0.2\linewidth, trim= {1.1cm, 0.8cm, 0cm, 0cm}, clip]{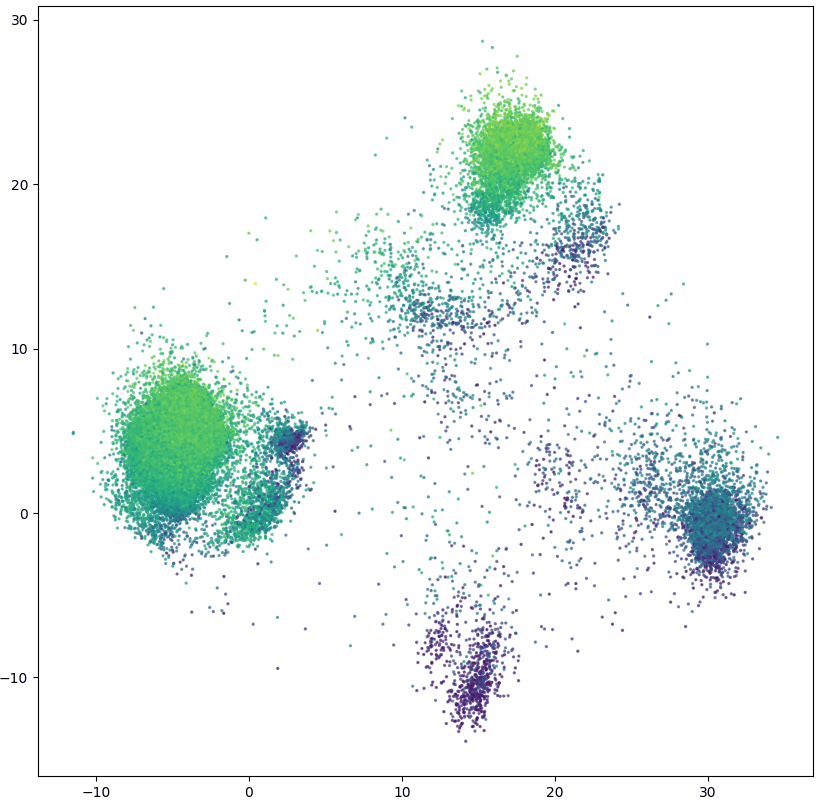} & \includegraphics[width=0.2\linewidth, trim= {1.1cm, 0.8cm, 0cm, 0cm}, clip]{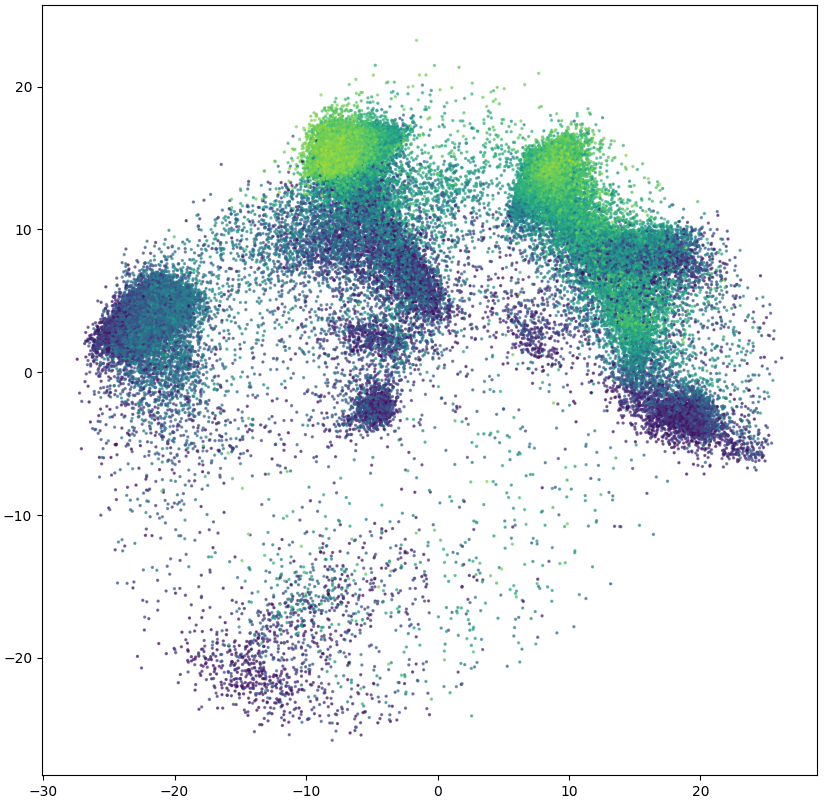} & \includegraphics[width=0.2\linewidth, trim= {1.1cm, 0.8cm, 0cm, 0cm}, clip]{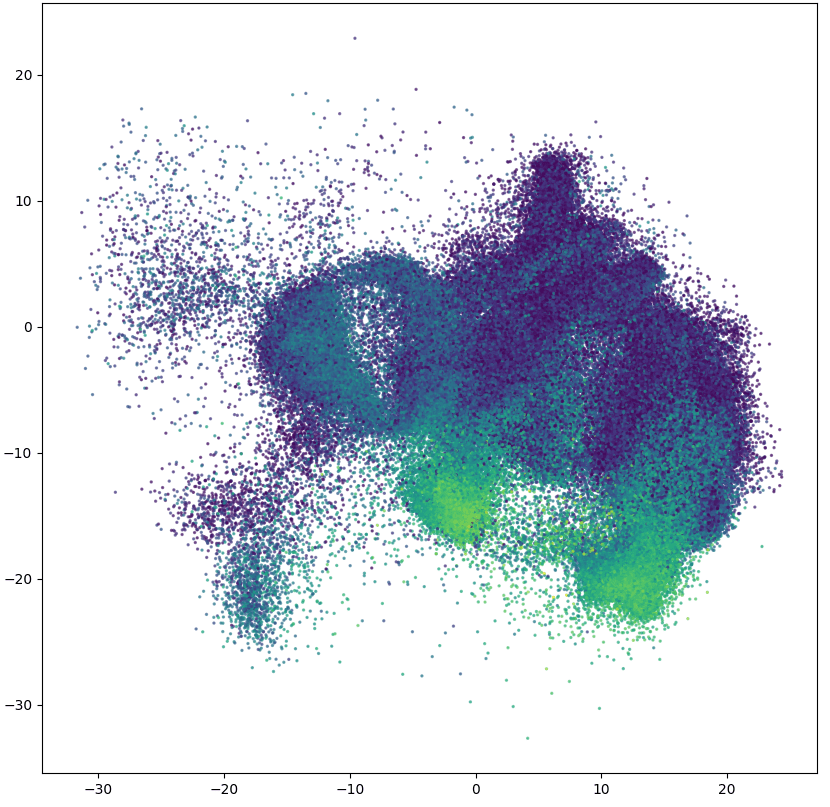}  \\

 \begin{turn}{90}t-SNE \end{turn} & \includegraphics[width=0.2\linewidth, trim= {1.1cm, 0.8cm, 0cm, 0cm}, clip]{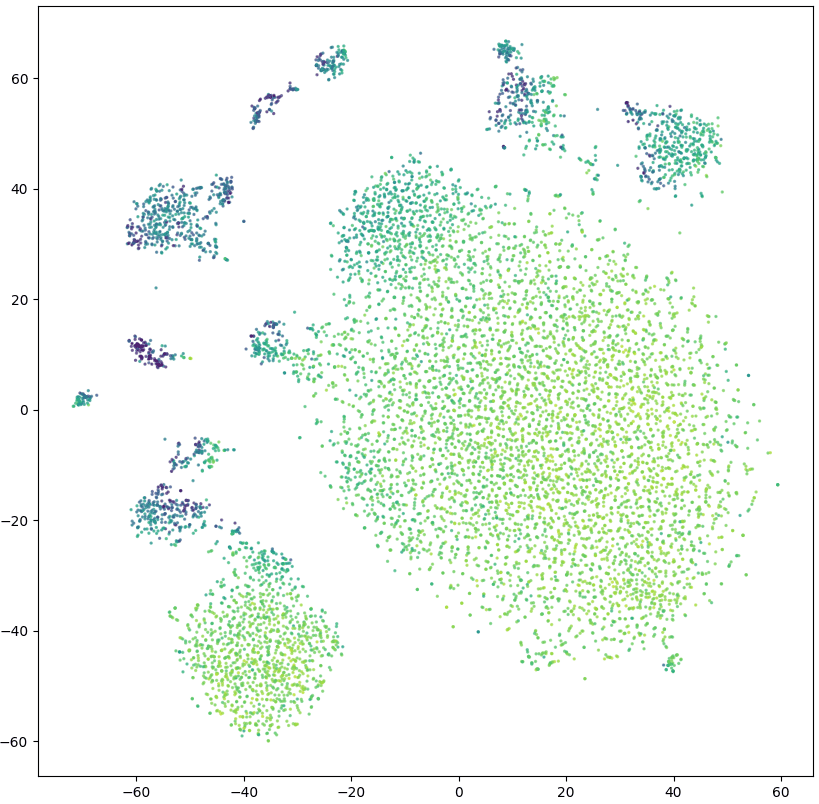} & \includegraphics[width=0.2\linewidth, trim= {1.15cm, 0.8cm, 0cm, 0cm}, clip]{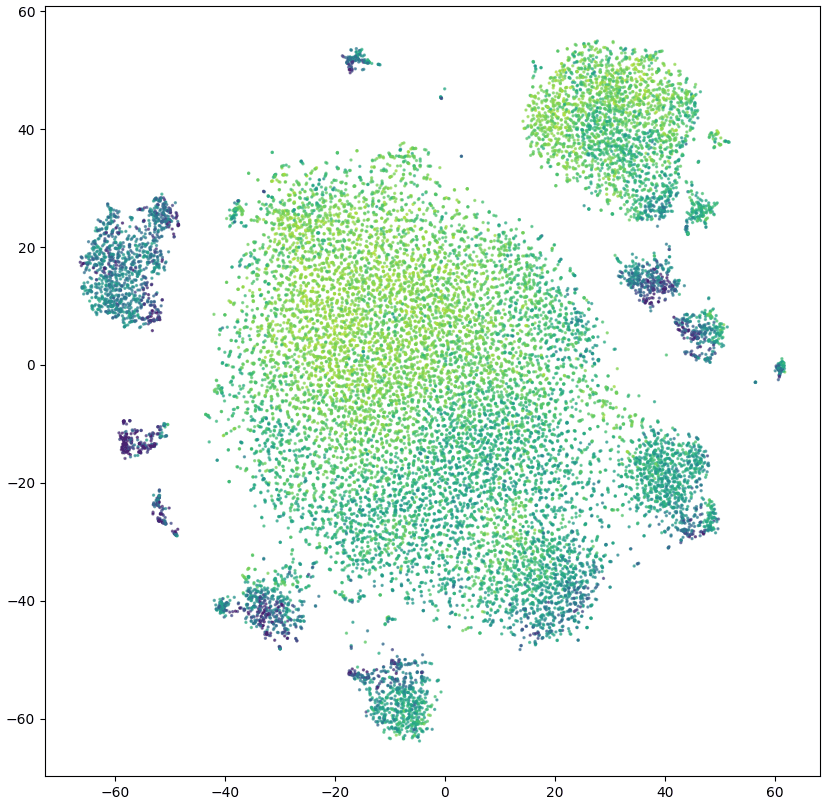} & \includegraphics[width=0.2\linewidth, trim= {1.1cm, 0.8cm, 0cm, 0cm}, clip]{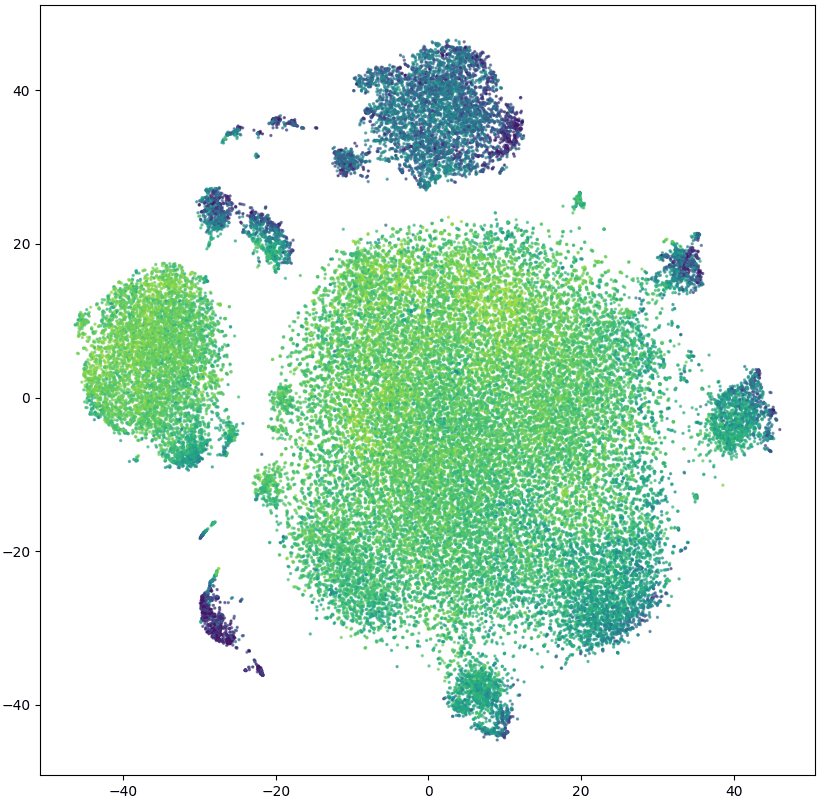} & \includegraphics[width=0.2\linewidth, trim= {1.1cm, 0.8cm, 0cm, 0cm}, clip]{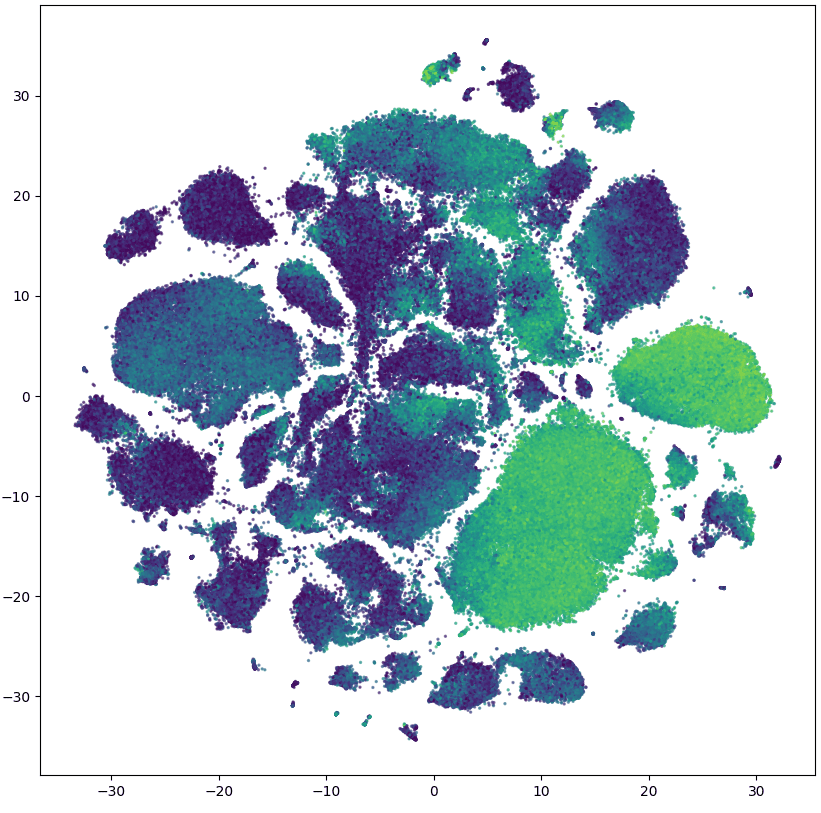}  \\

 \begin{turn}{90}    UMAP \end{turn}& \includegraphics[width=0.2\linewidth, trim= {1.15cm, 0.8cm, 0cm, 0cm}, clip]{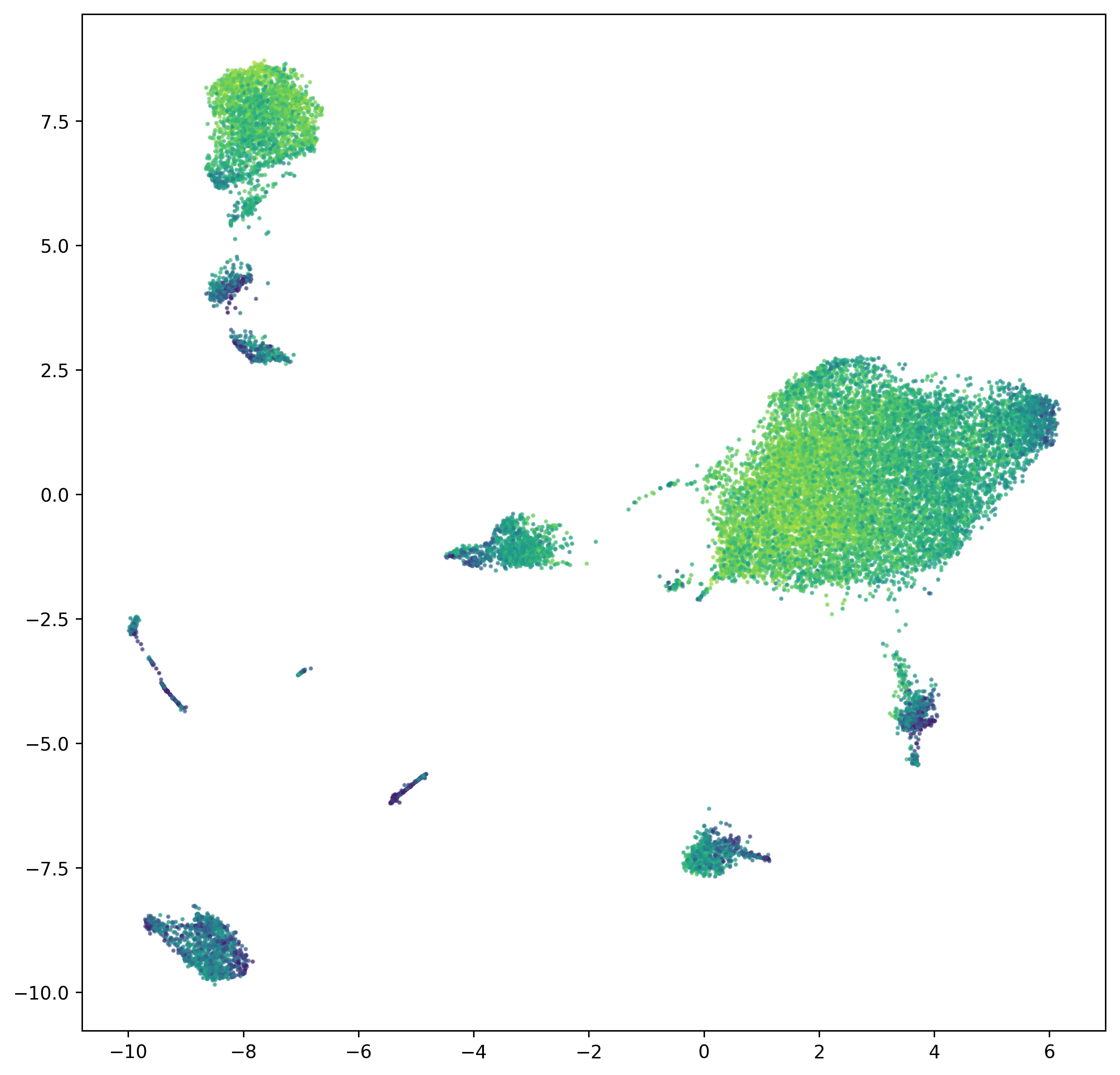} & \includegraphics[width=0.2\linewidth, trim= {1.1cm, 0.8cm, 0cm, 0cm}, clip]{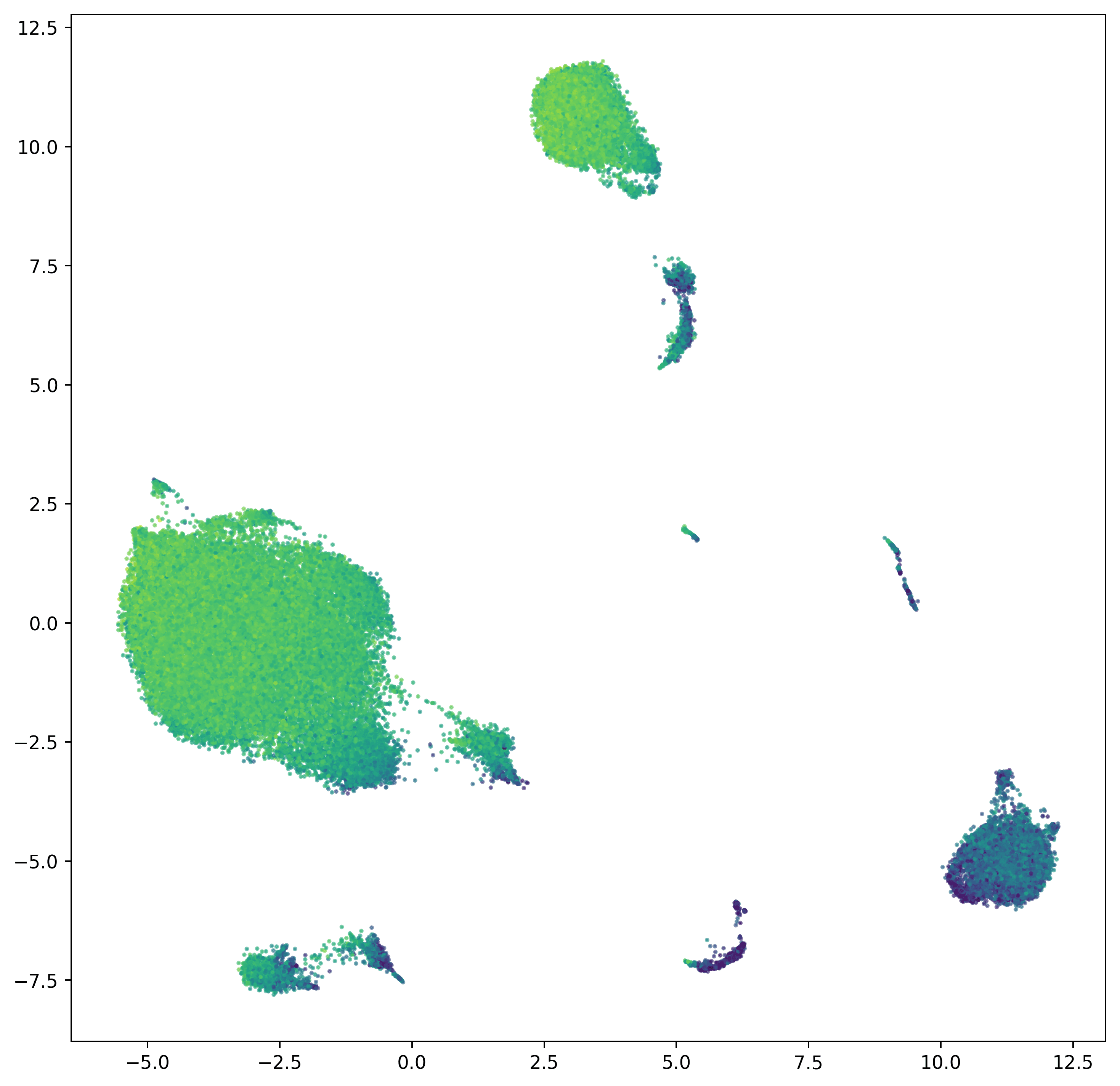} & \includegraphics[width=0.2\linewidth, trim= {1.1cm, 0.8cm, 0cm, 0cm}, clip]{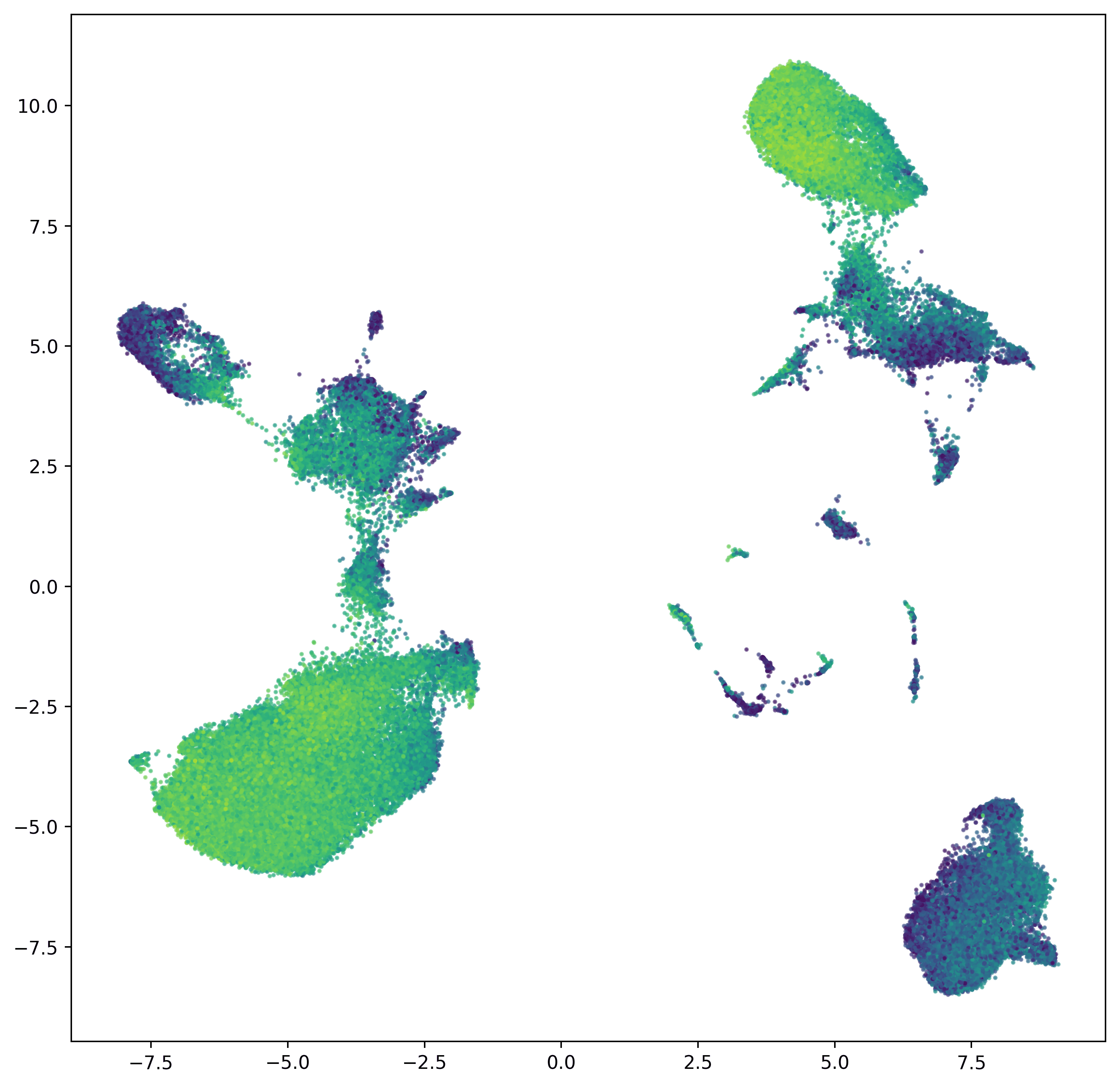} & \includegraphics[width=0.2\linewidth, trim= {1.1cm, 0.8cm, 0cm, 0cm}, clip]{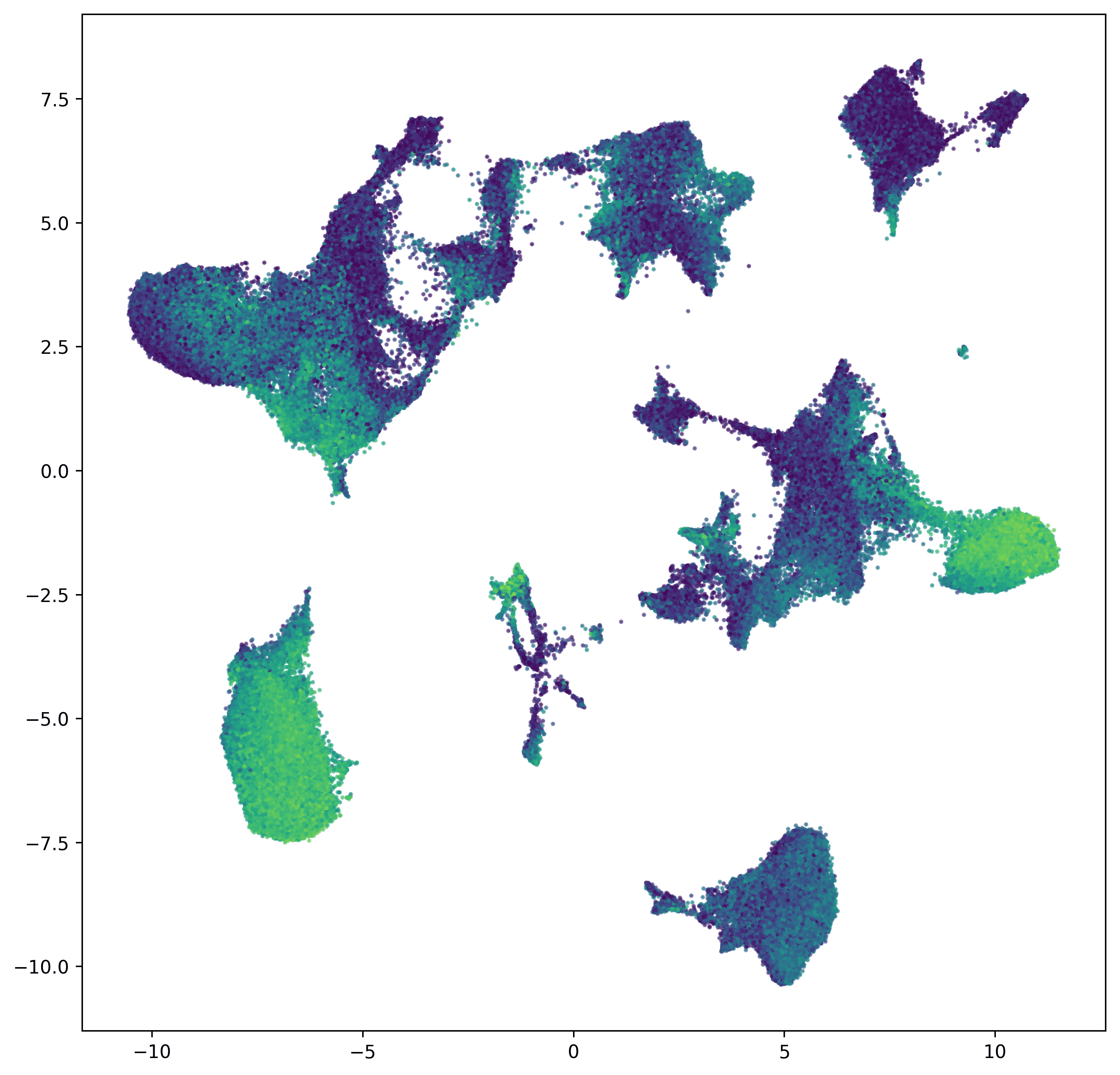}  \\

    \end{tabular}
    \caption{The Wong dataset visualized by SONG, SONG + Reinit, Parametric t-SNE and UMAP. The colors represent the CCR7 expression levels  following the visualizations provided in \cite{becht2019dimensionality}, where Light Green represents high CCR7 expression and Dark Purple represents low CCR7 expressions.}
    \label{wong_biased}
\end{figure*}

\begin{figure*}[ht]
\centering
\begin{tabular}{l c c c c c}
        & 2-classes & 4-classes & 6-classes & 8-classes & 10-classes \\
     \begin{turn}{90}SONG\end{turn} & \includegraphics[width=0.18\linewidth]{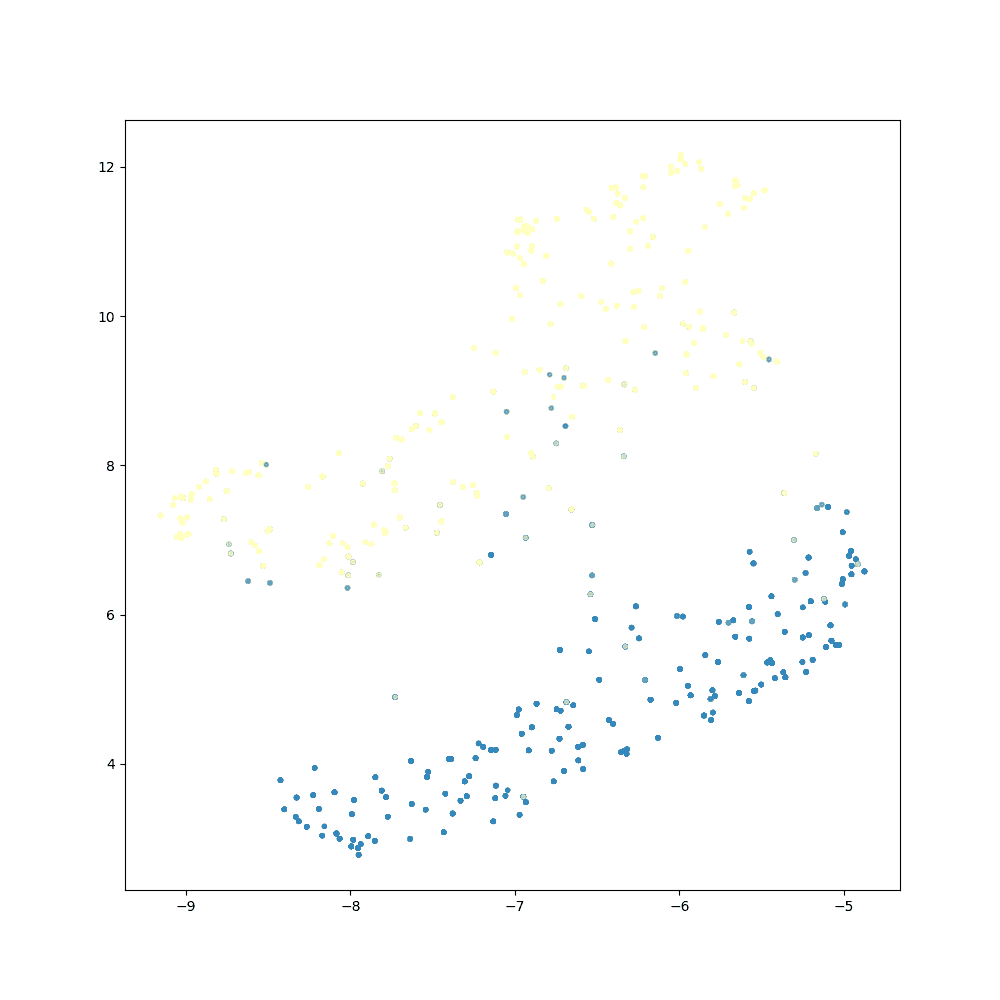} & \includegraphics[width=0.18\linewidth]{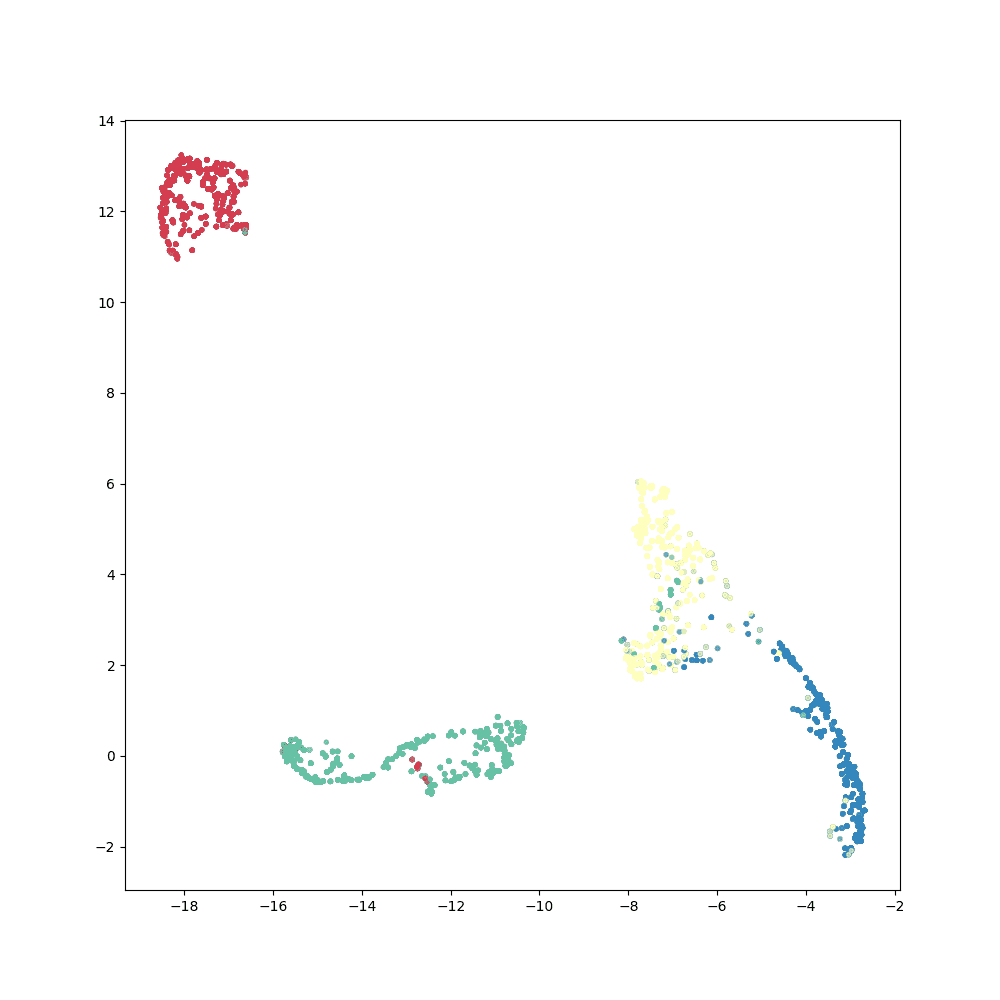} & \includegraphics[width=0.18\linewidth]{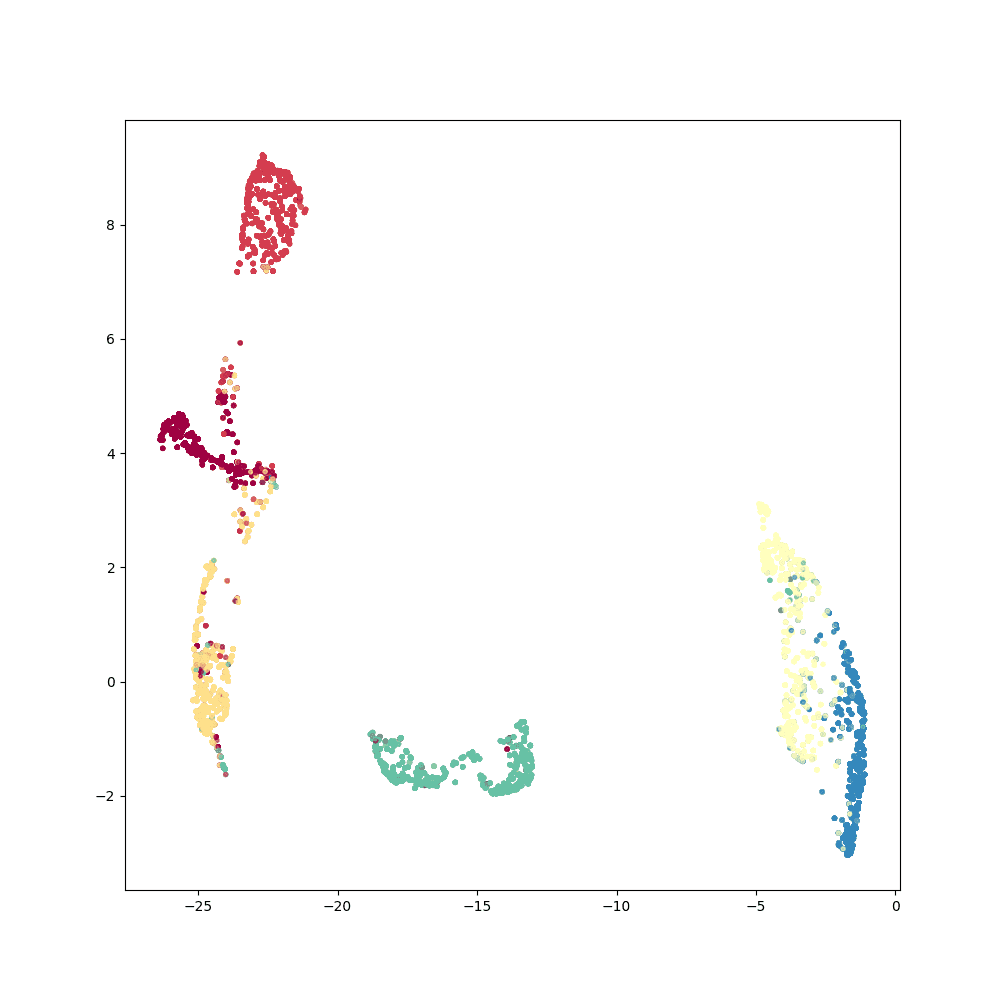} &  \includegraphics[width=0.18\linewidth]{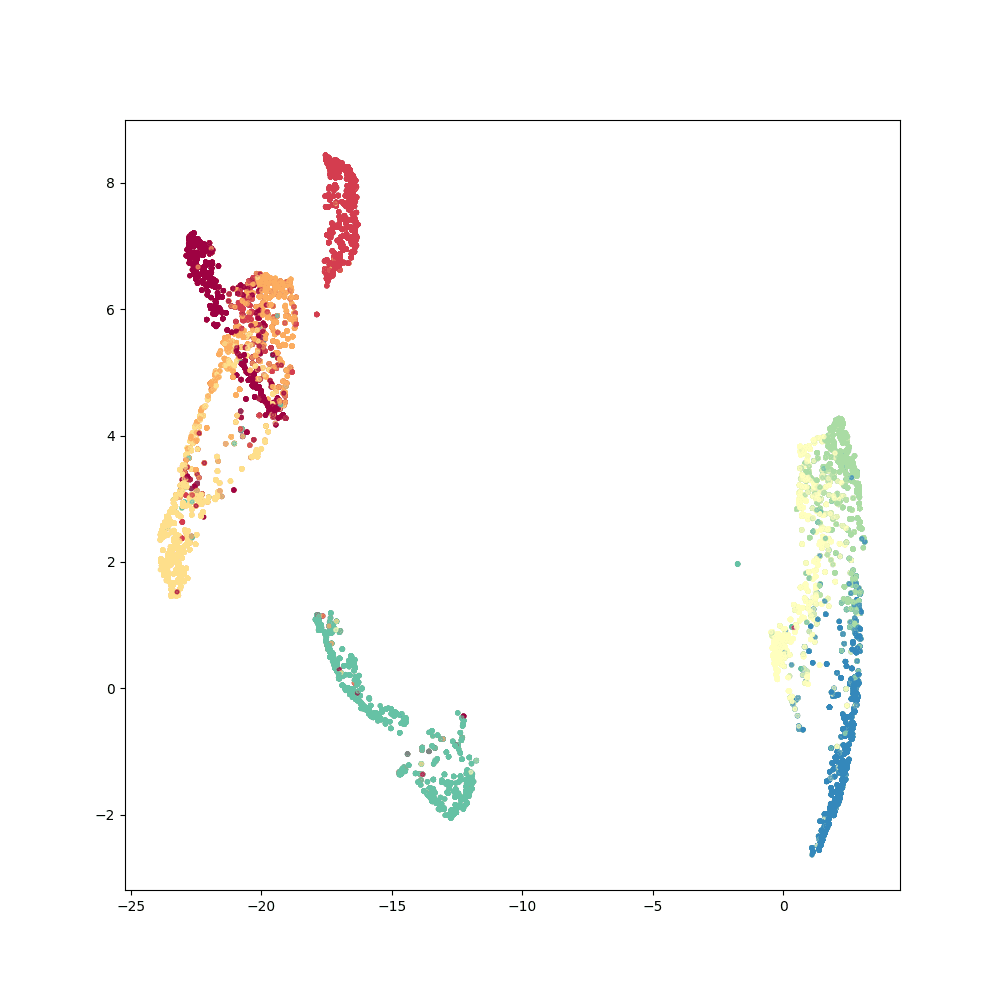} & \includegraphics[width=0.18\linewidth]{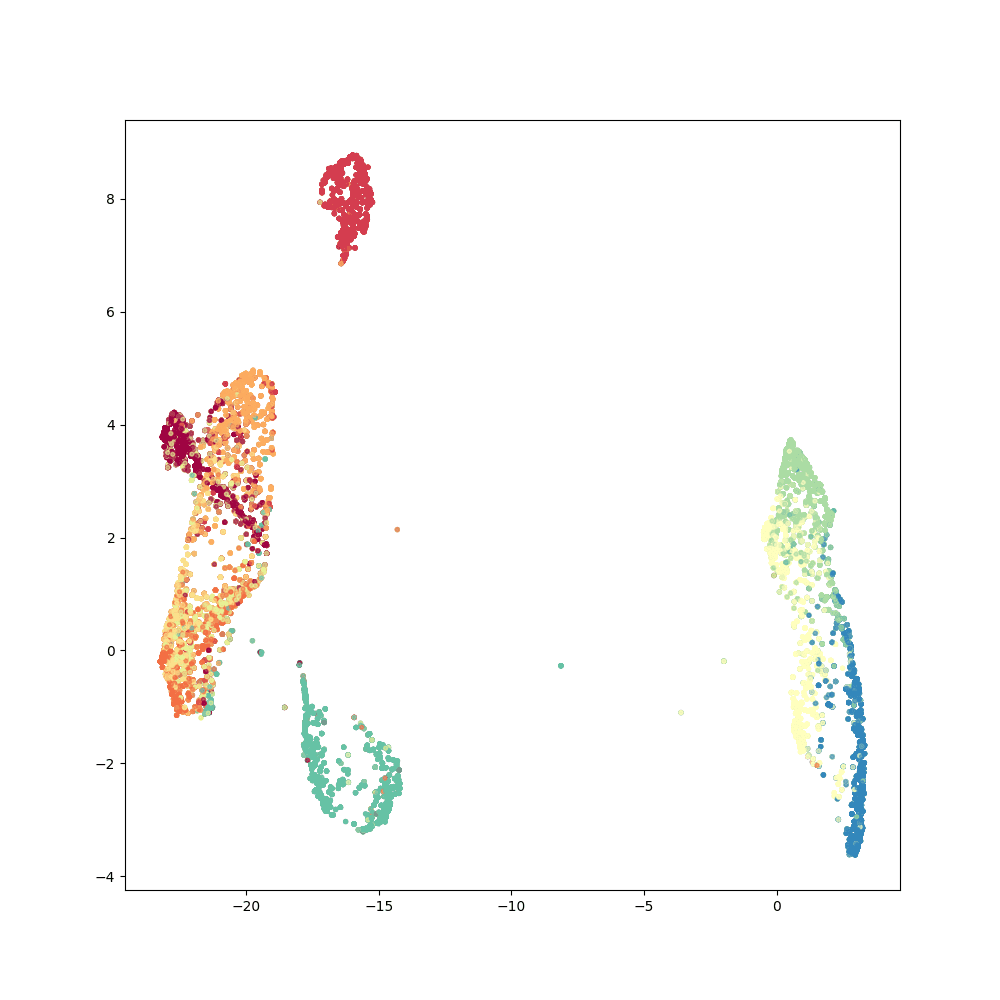} \\
     
     \begin{turn}{90}SONG + Reinit\end{turn} &\includegraphics[width=0.18\linewidth]{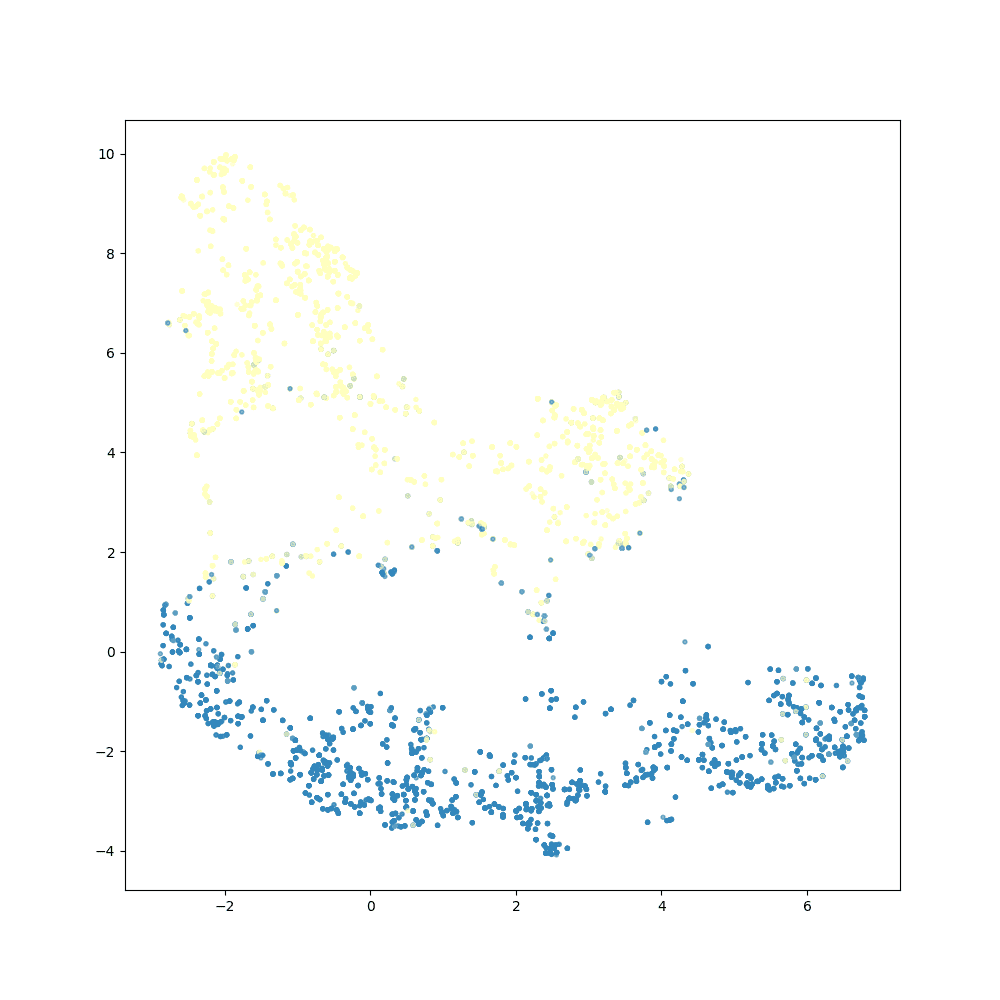} & \includegraphics[width=0.18\linewidth]{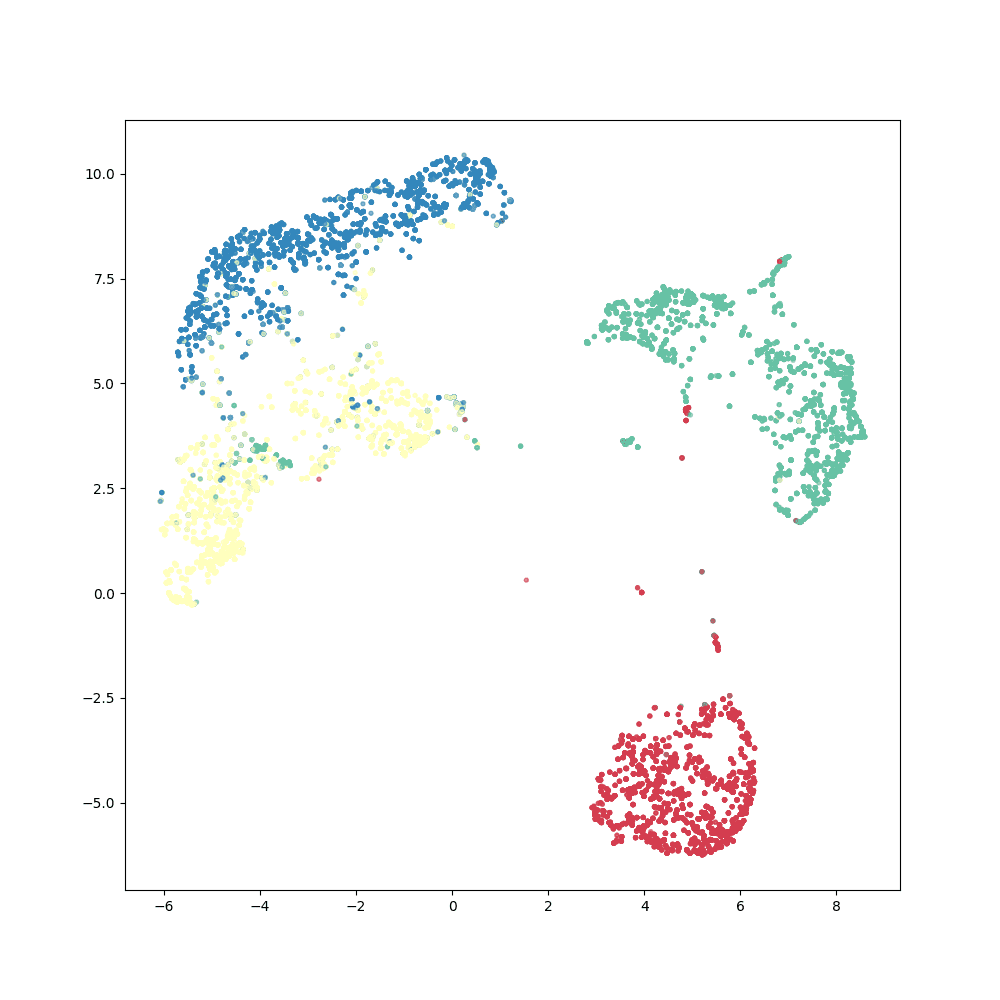} & \includegraphics[width=0.18\linewidth]{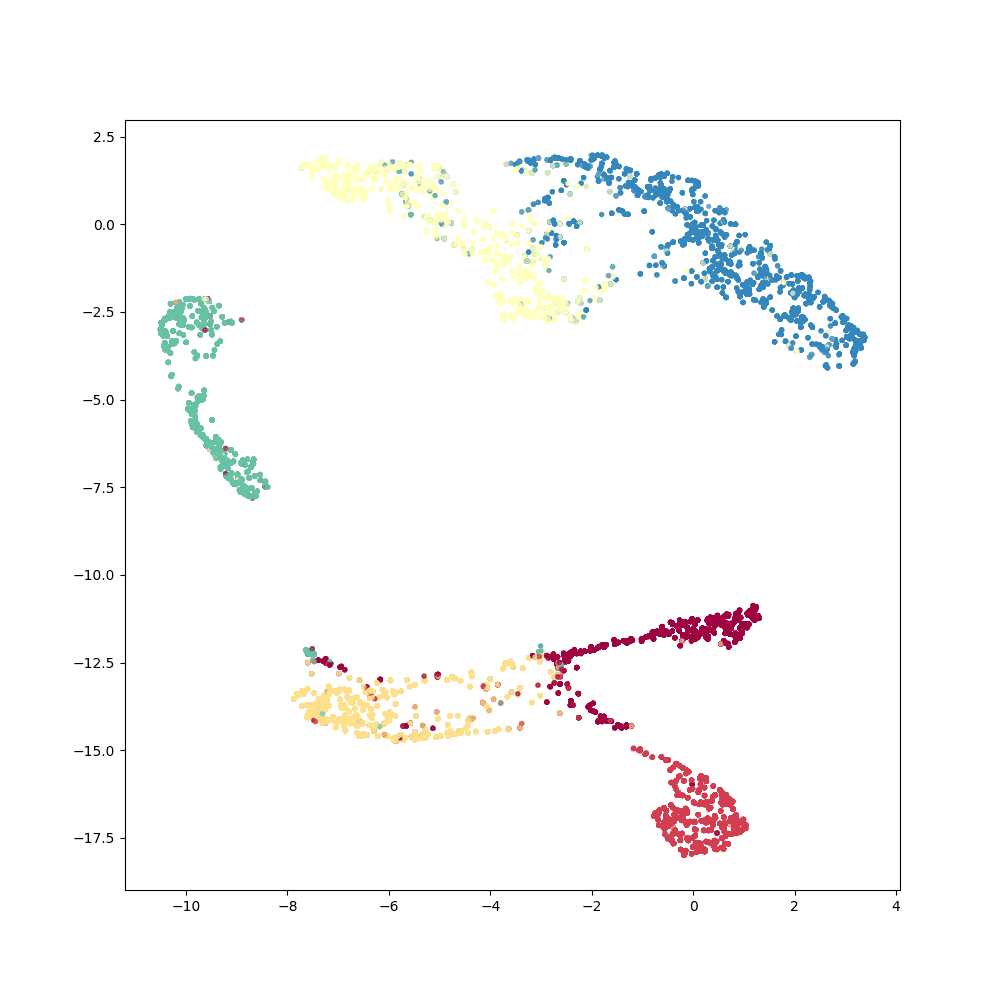} & \includegraphics[width=0.18\linewidth]{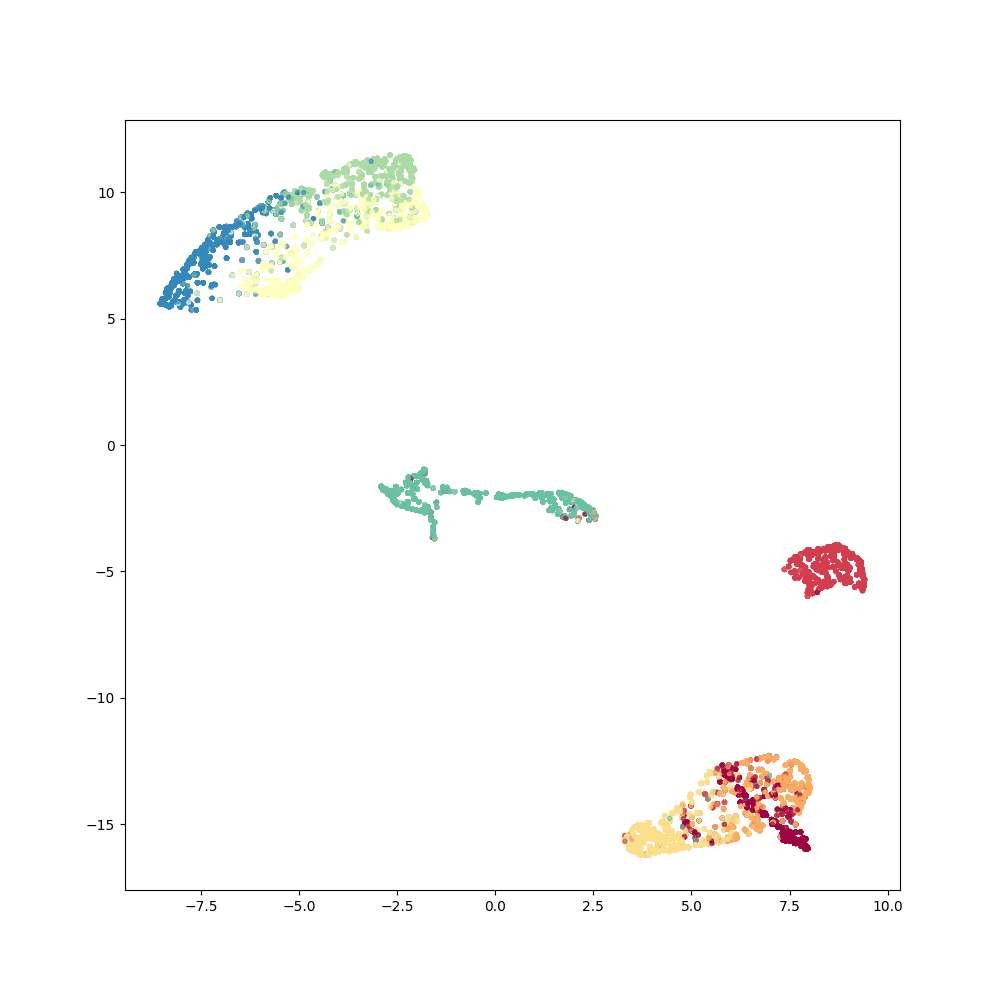} & \includegraphics[width=0.18\linewidth]{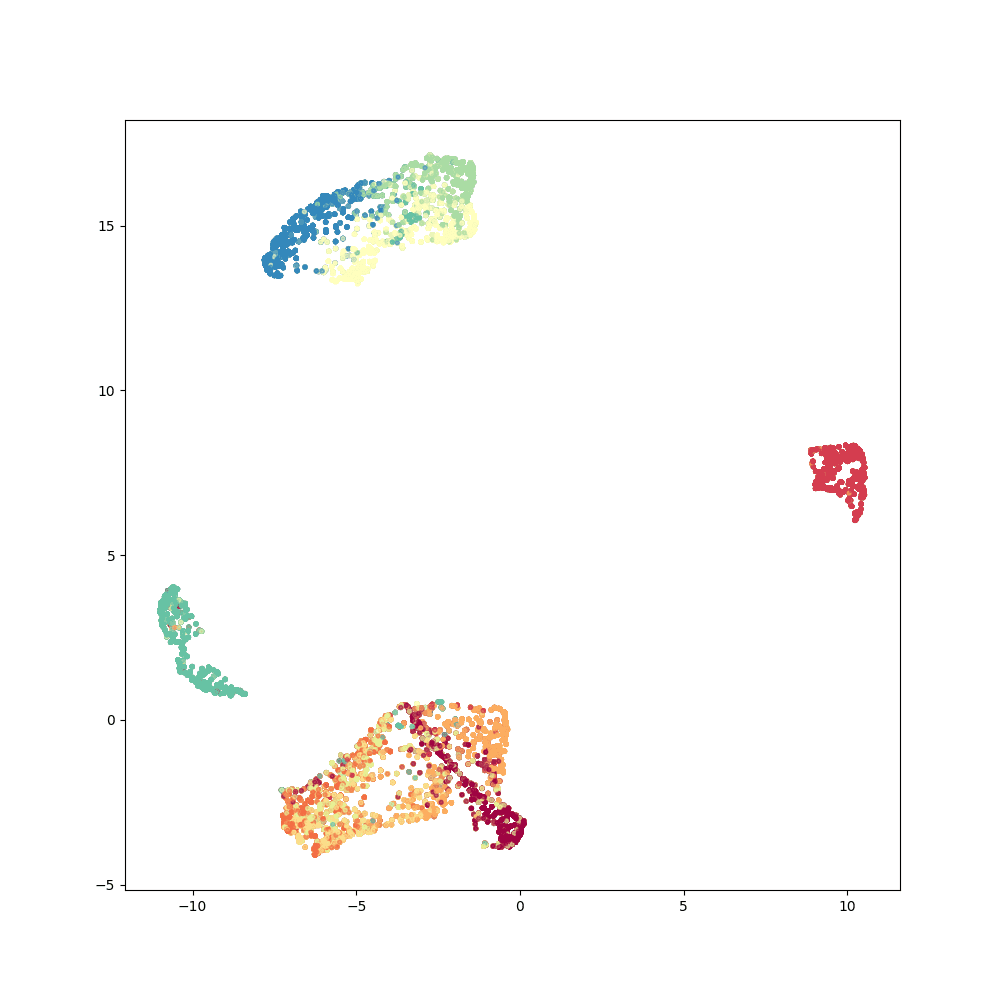} \\
     
     \begin{turn}{90}Parametric t-SNE\end{turn} &\includegraphics[width=0.18\linewidth]{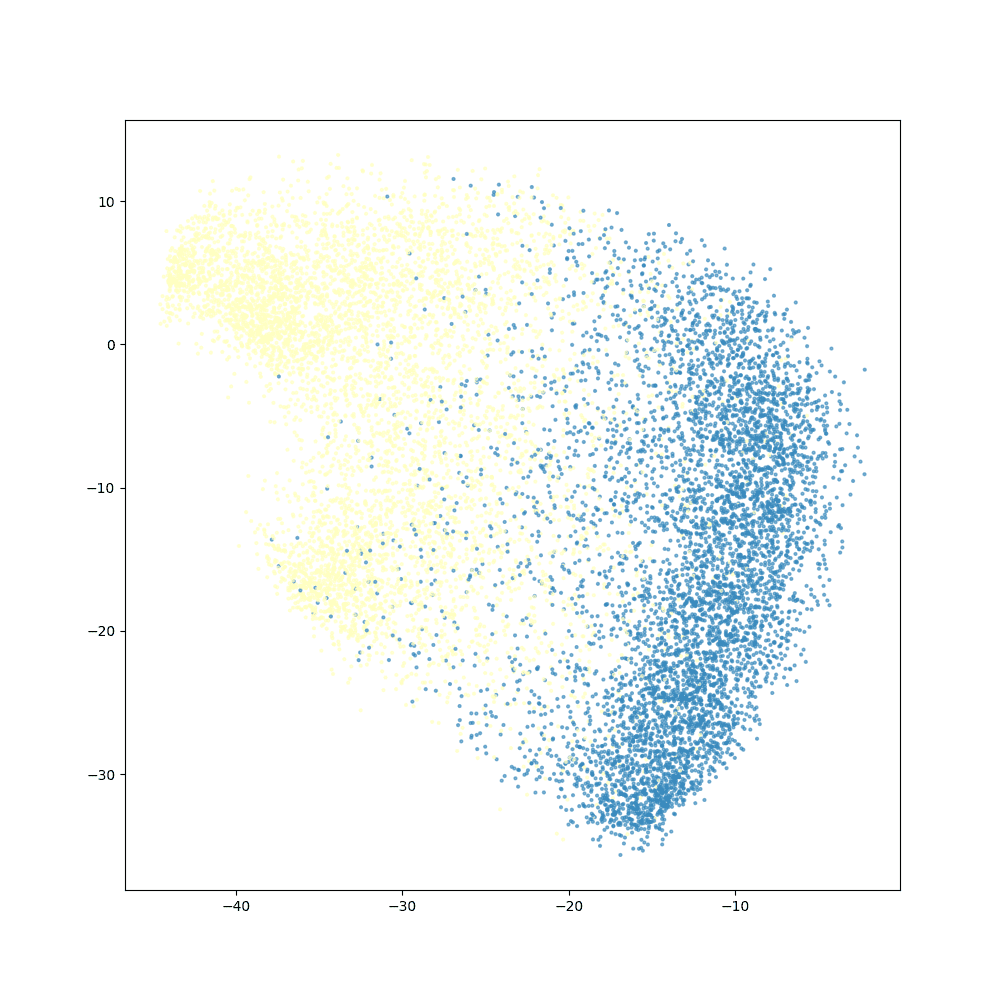} & \includegraphics[width=0.18\linewidth]{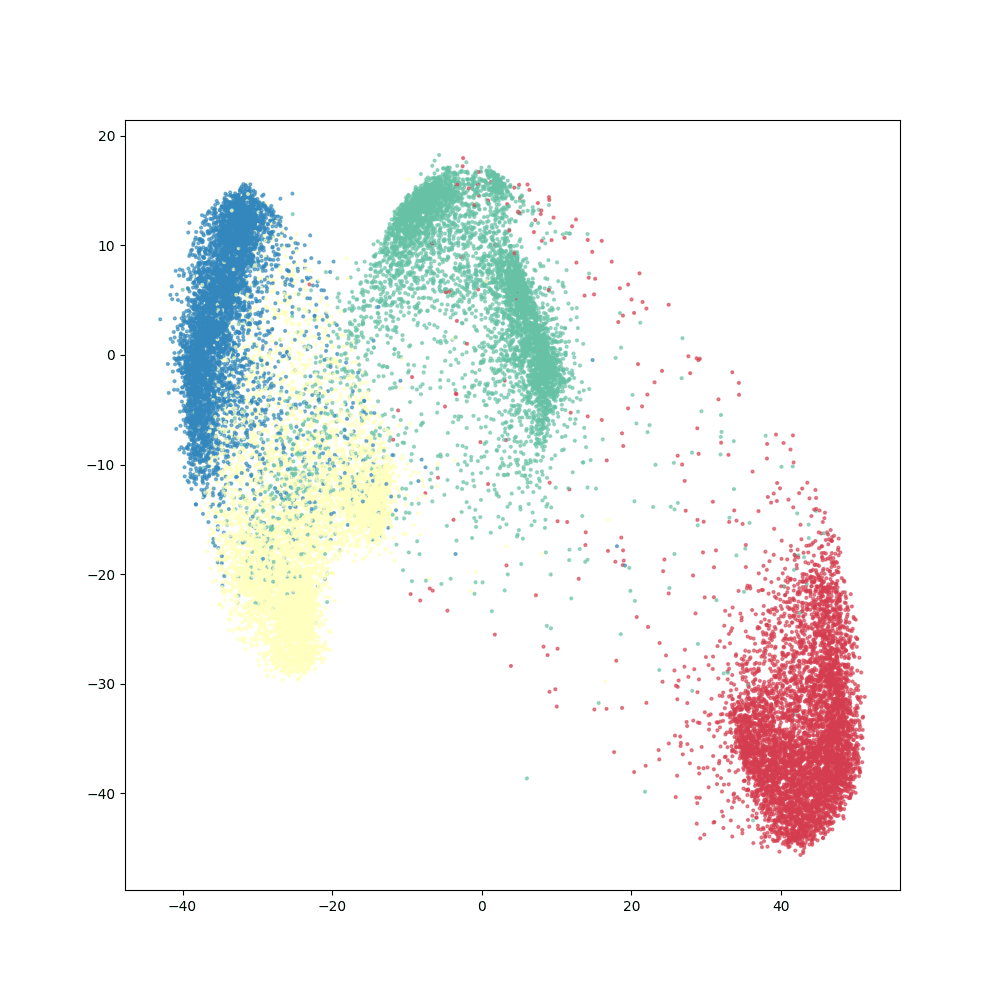} & \includegraphics[width=0.18\linewidth]{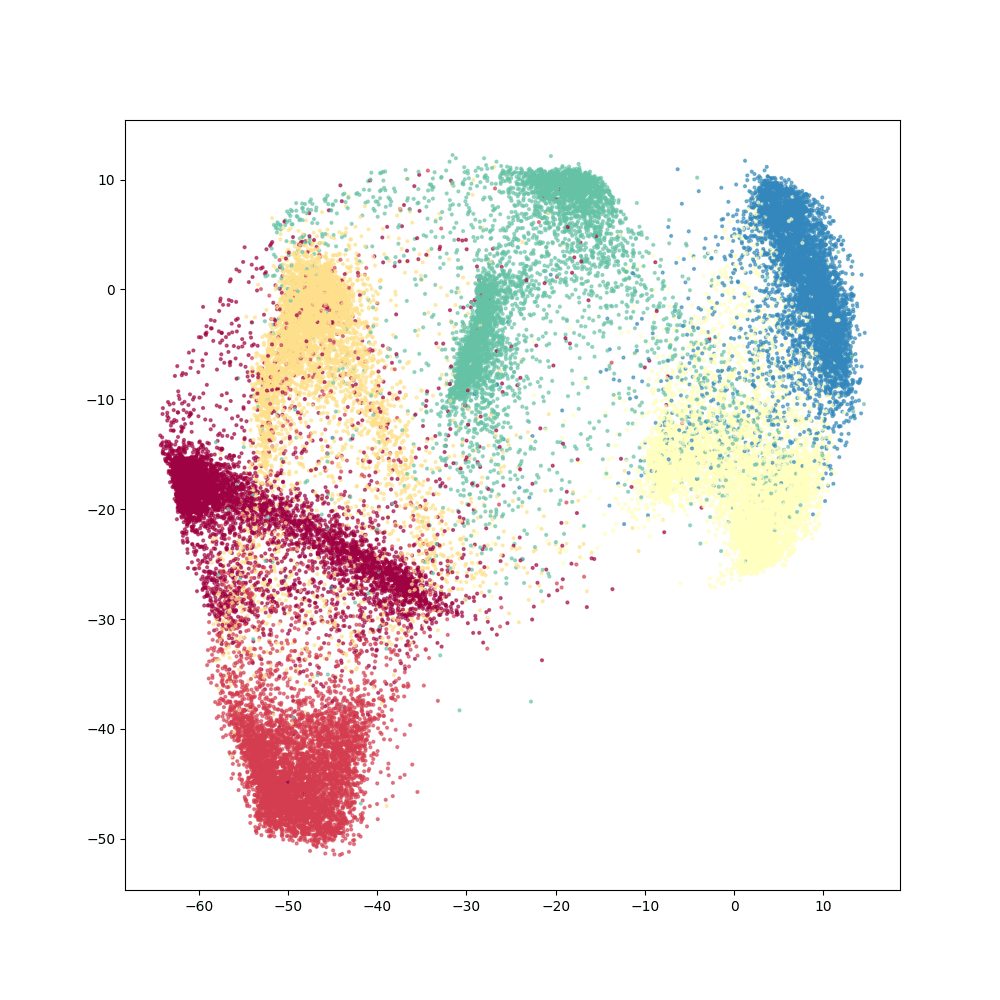} & \includegraphics[width=0.18\linewidth]{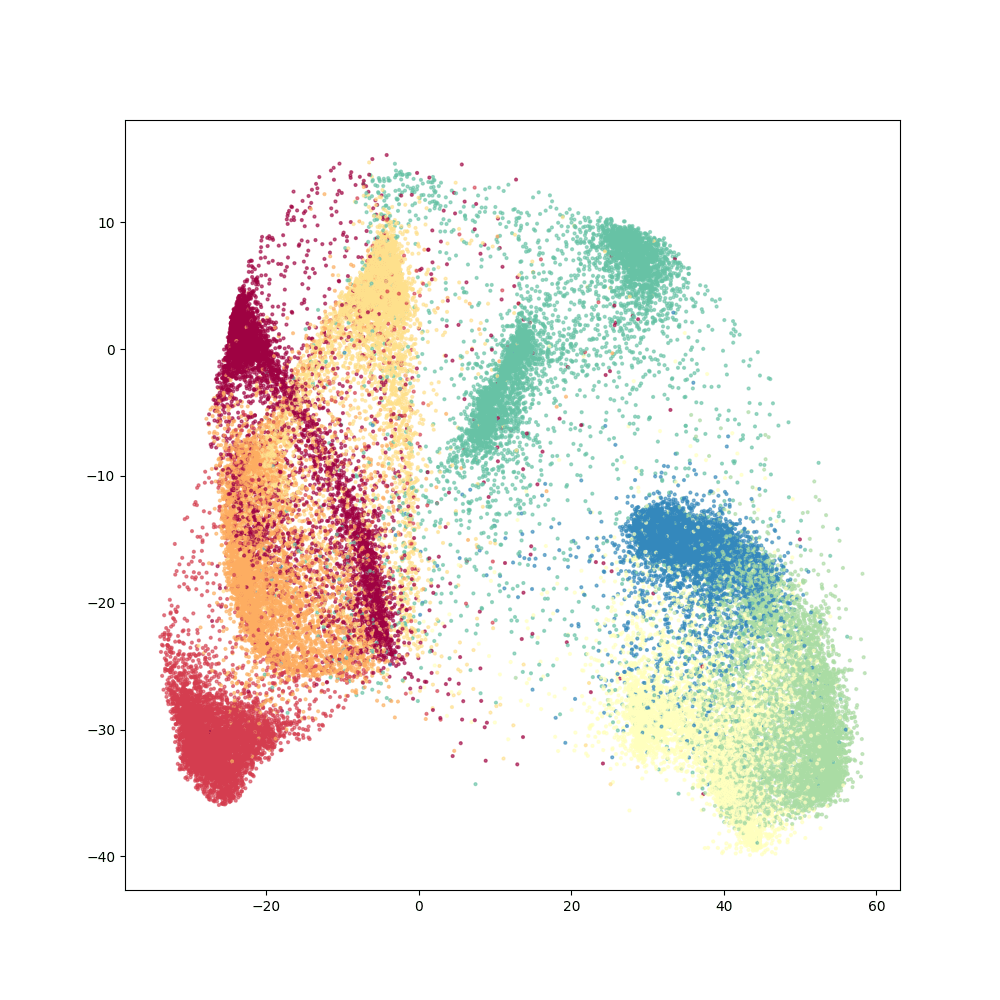} & \includegraphics[width=0.18\linewidth]{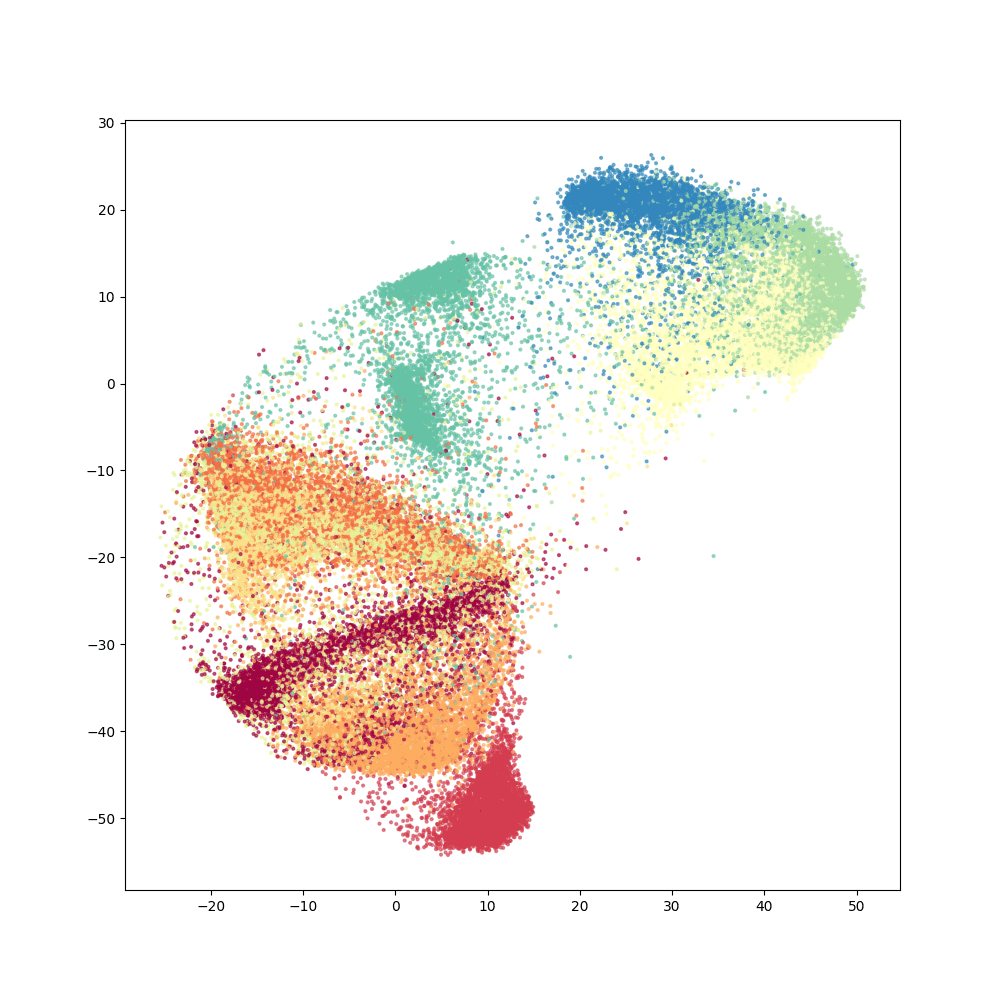} \\
     
      \begin{turn}{90}t-SNE\end{turn} &\includegraphics[width=0.18\linewidth]{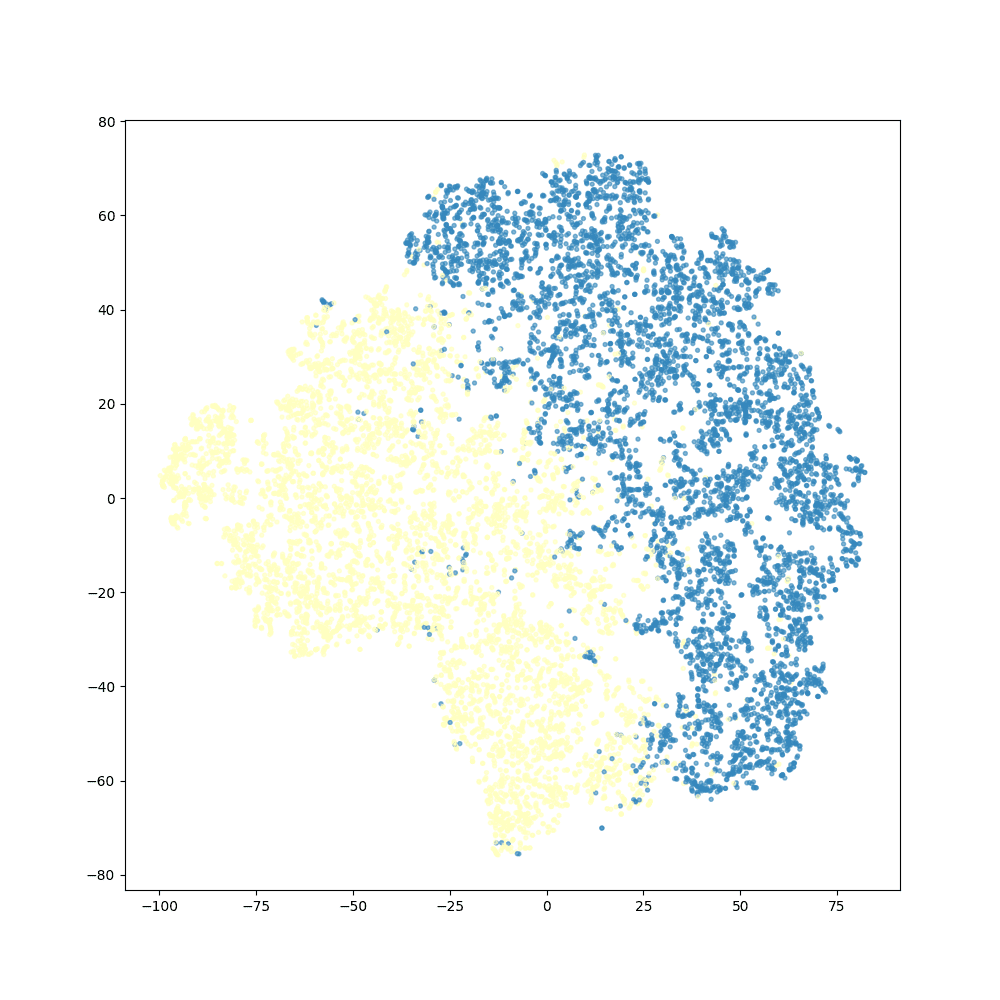} & \includegraphics[width=0.18\linewidth]{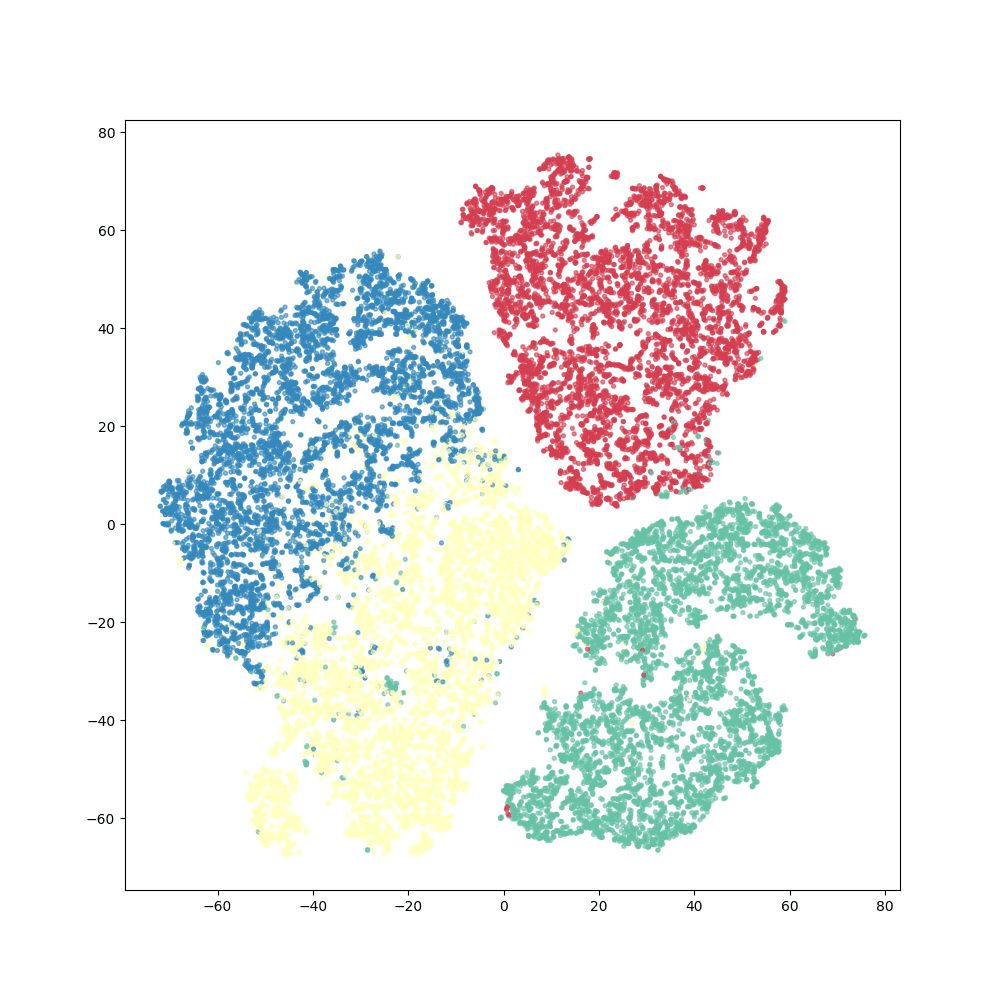} & \includegraphics[width=0.18\linewidth]{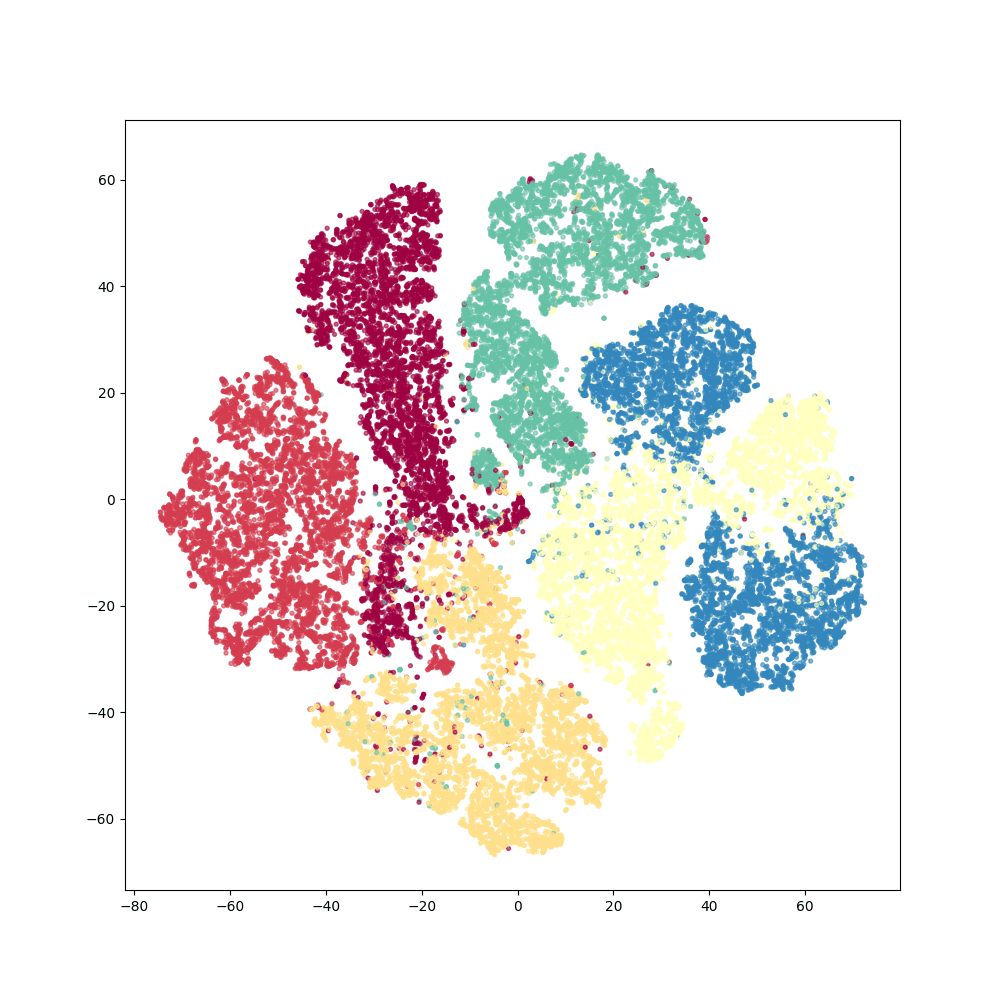} & \includegraphics[width=0.18\linewidth]{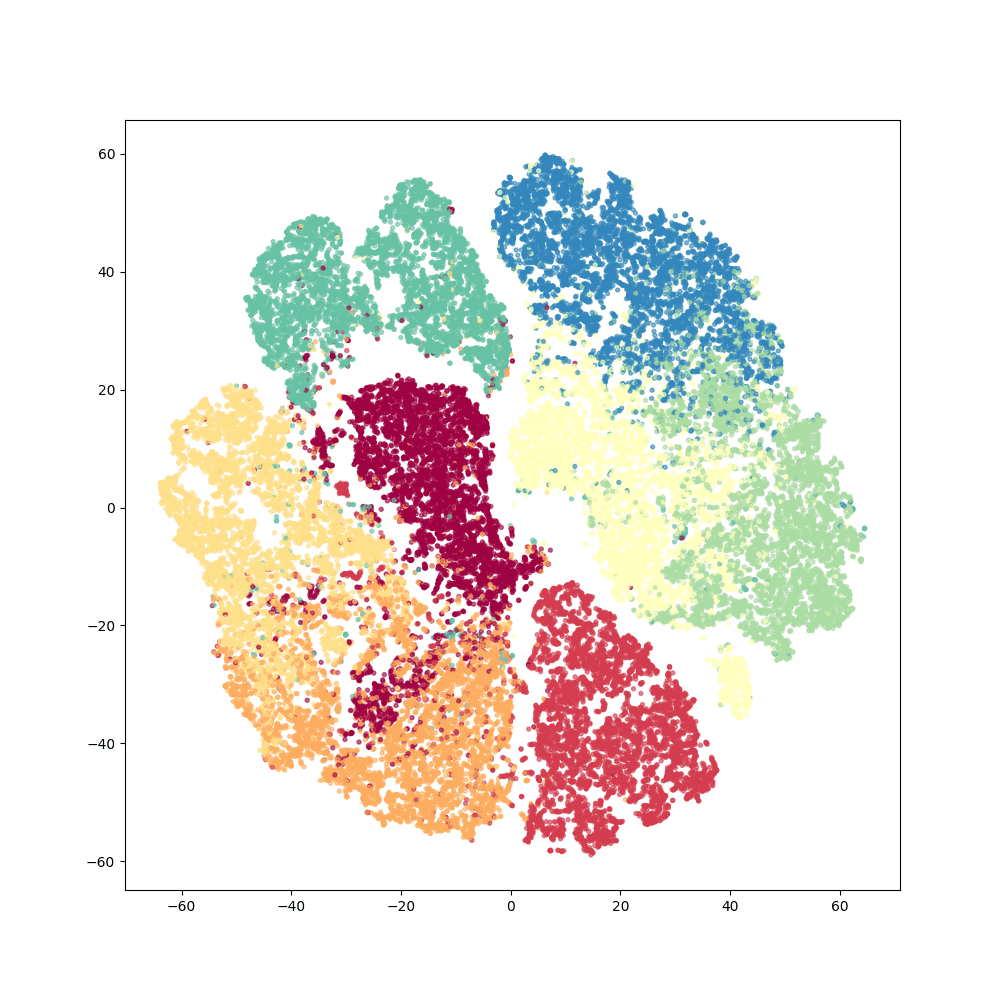} & \includegraphics[width=0.18\linewidth]{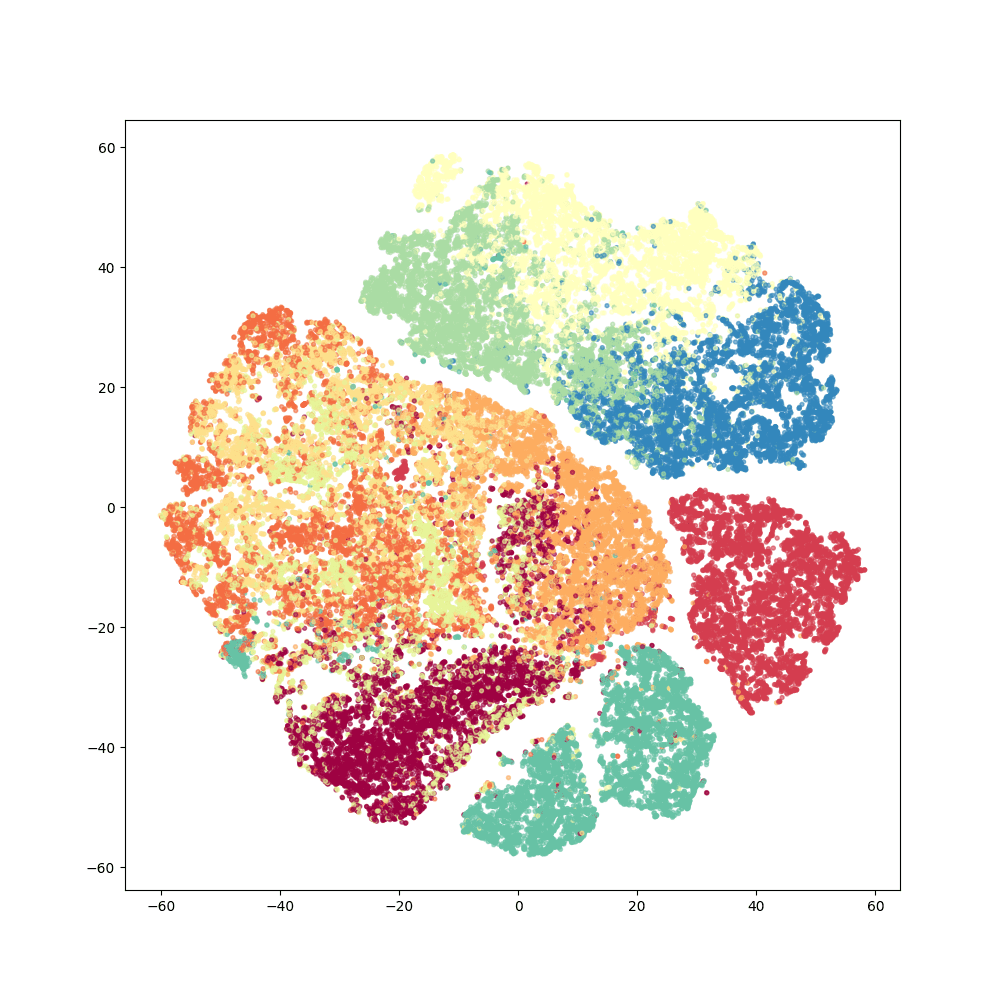} \\
      
     \begin{turn}{90}UMAP\end{turn} &\includegraphics[width=0.18\linewidth]{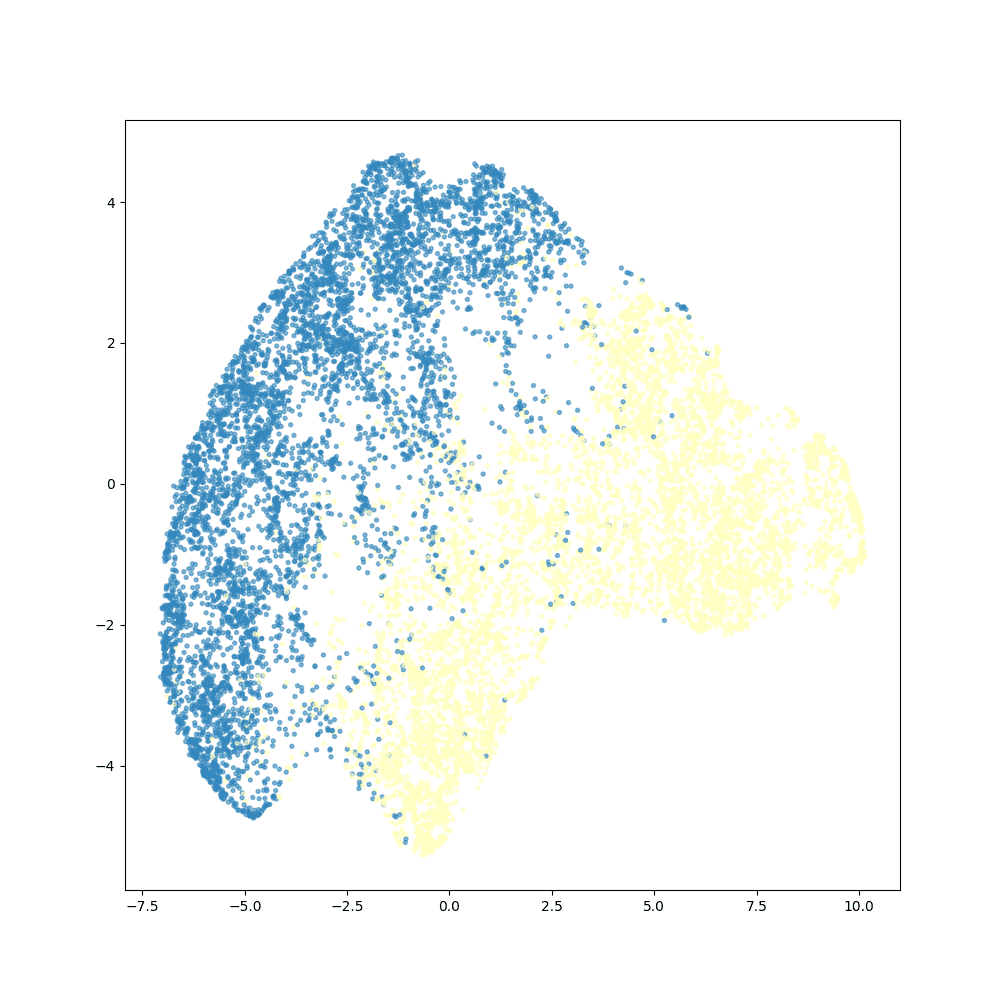} & \includegraphics[width=0.18\linewidth]{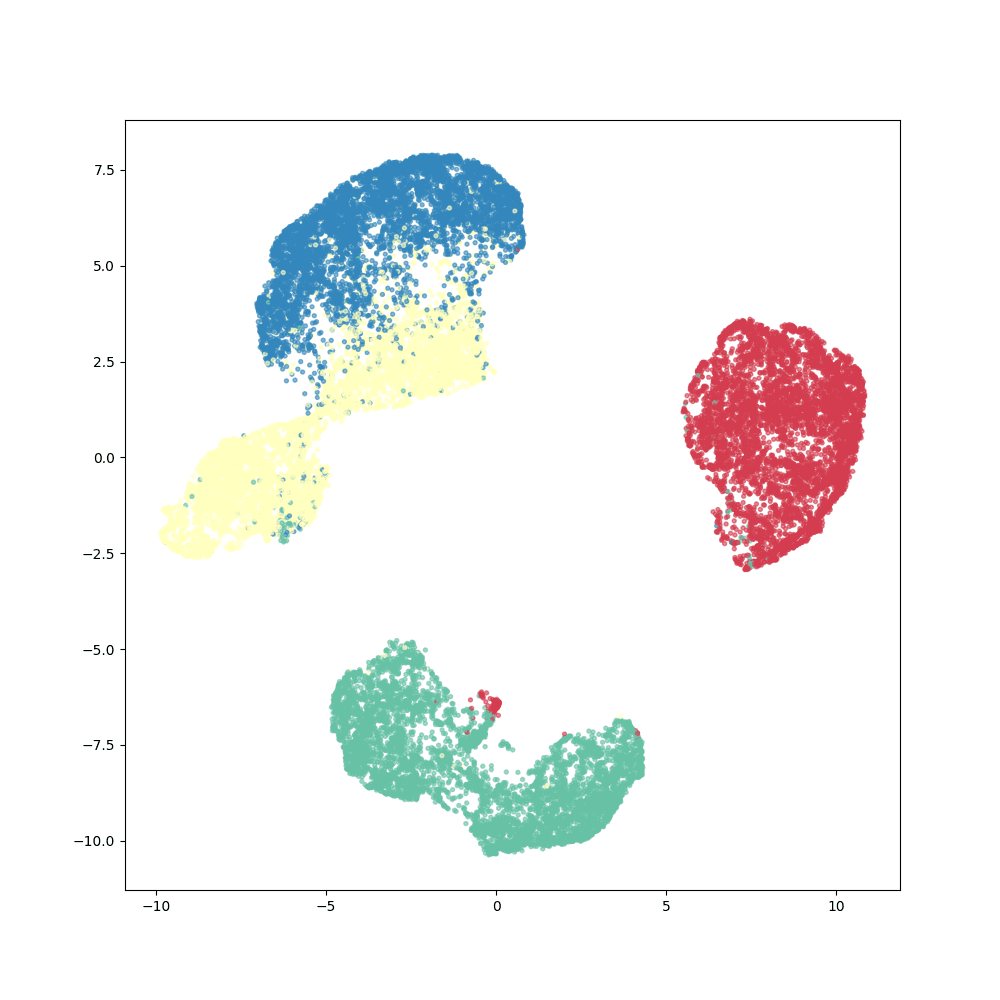} & \includegraphics[width=0.18\linewidth]{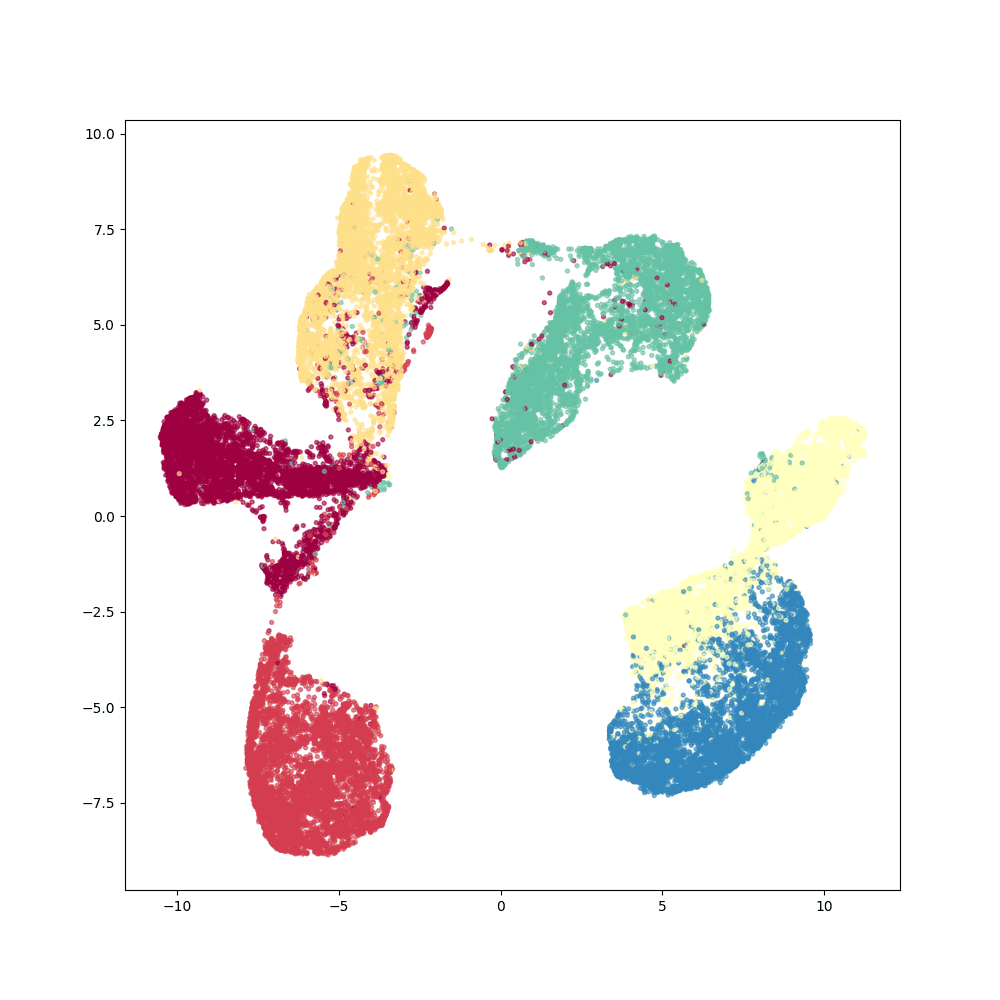} & \includegraphics[width=0.18\linewidth]{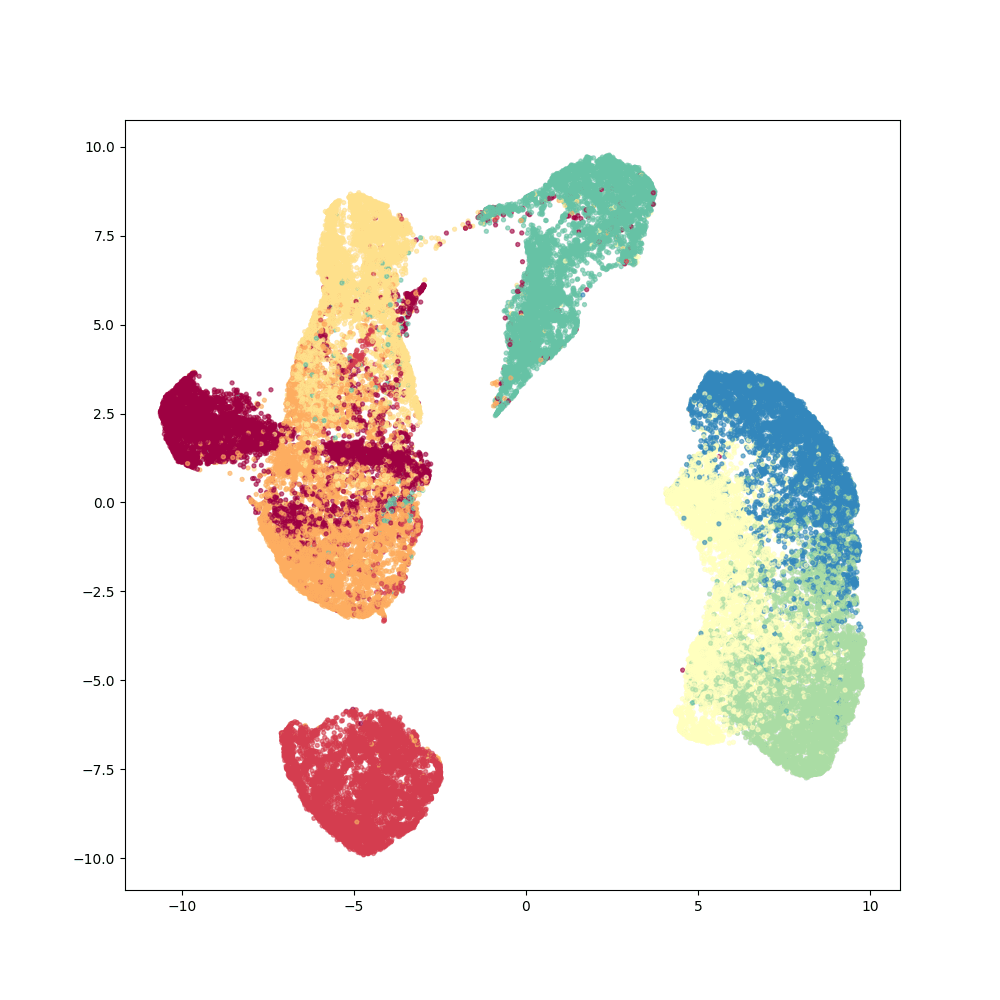} & \includegraphics[width=0.18\linewidth]{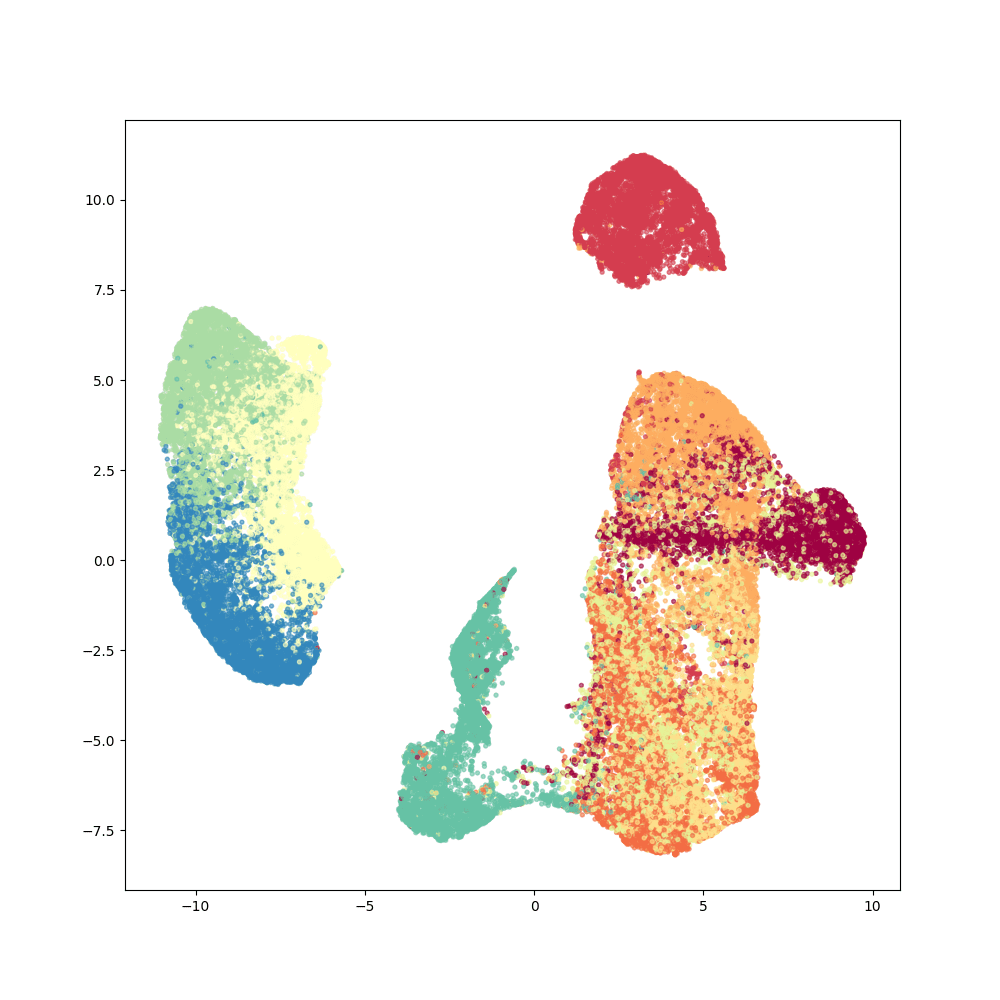} \\

     \end{tabular}

\caption{\label{fashion_biased}
Incremental visualization of the Fashion MNIST dataset using SONG, SONG + Reinit, Parametric t-SNE, t-SNE and UMAP, where two classes are added at a time.}
\end{figure*}

\begin{figure*}[ht]
\centering
\begin{tabular}{l c c c c c}
        & 2-classes & 4-classes & 6-classes & 8-classes & 10-classes \\
     \begin{turn}{90}SONG\end{turn}  & \includegraphics[width=0.18\linewidth]{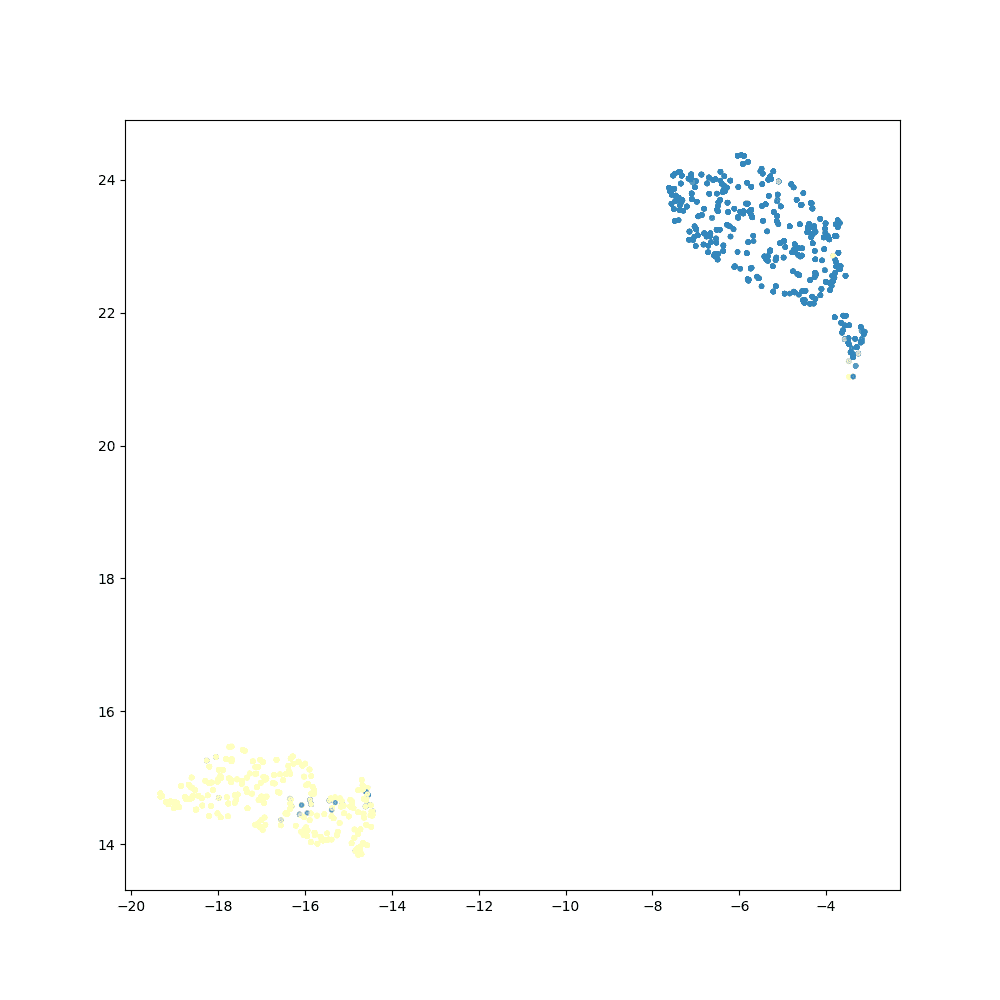} & \includegraphics[width=0.18\linewidth]{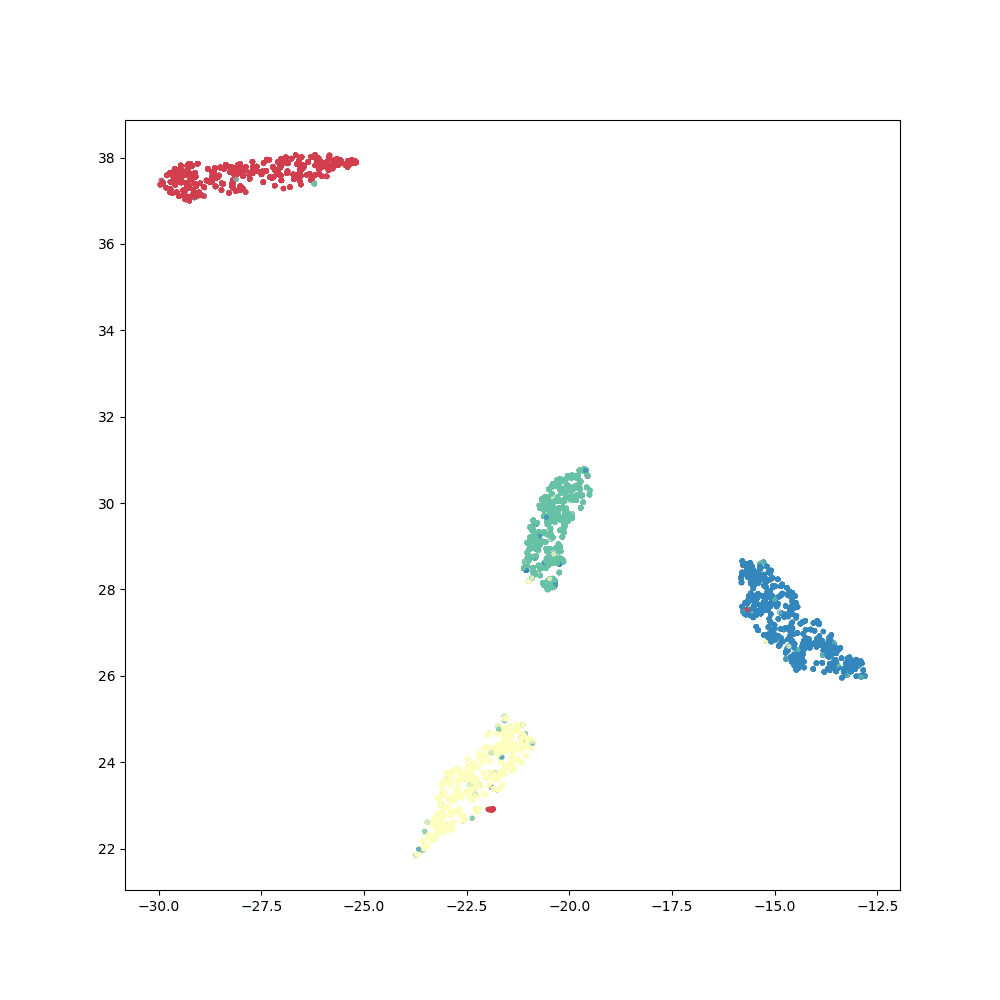} & \includegraphics[width=0.18\linewidth]{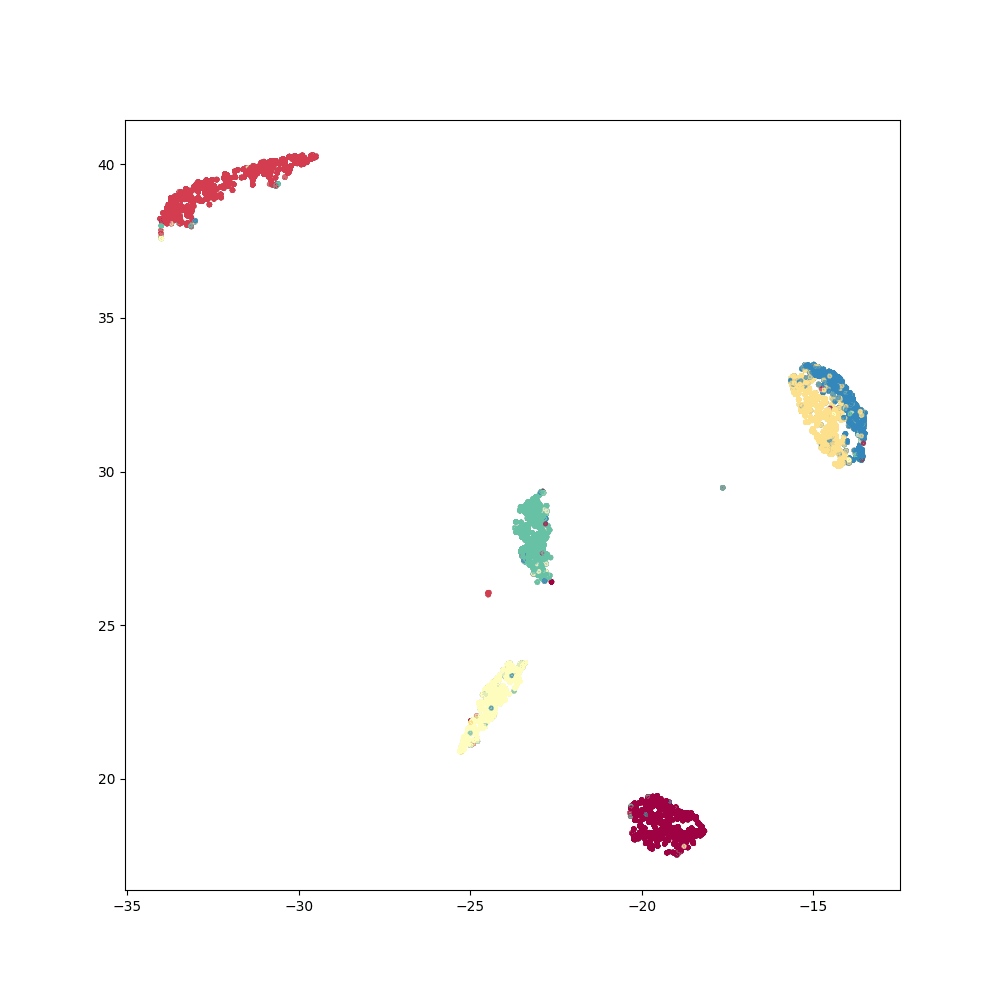} &  \includegraphics[width=0.18\linewidth]{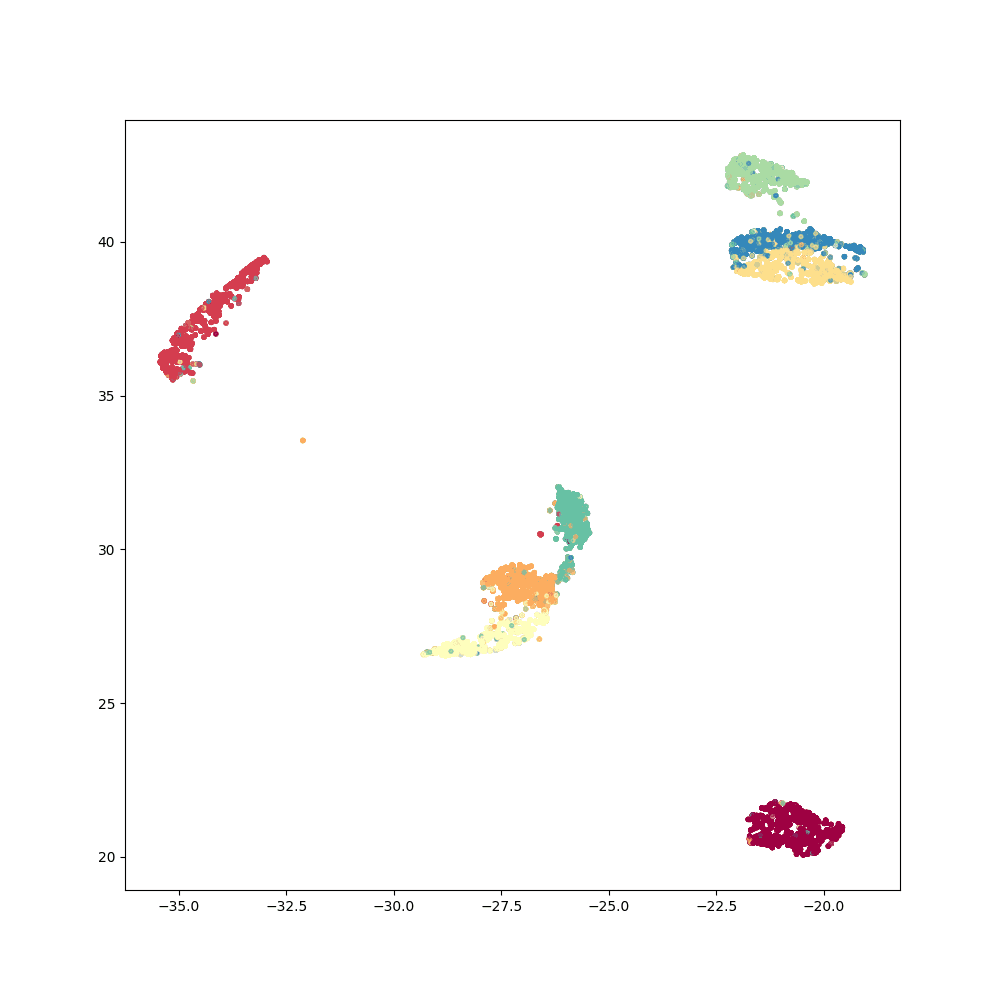} & \includegraphics[width=0.18\linewidth]{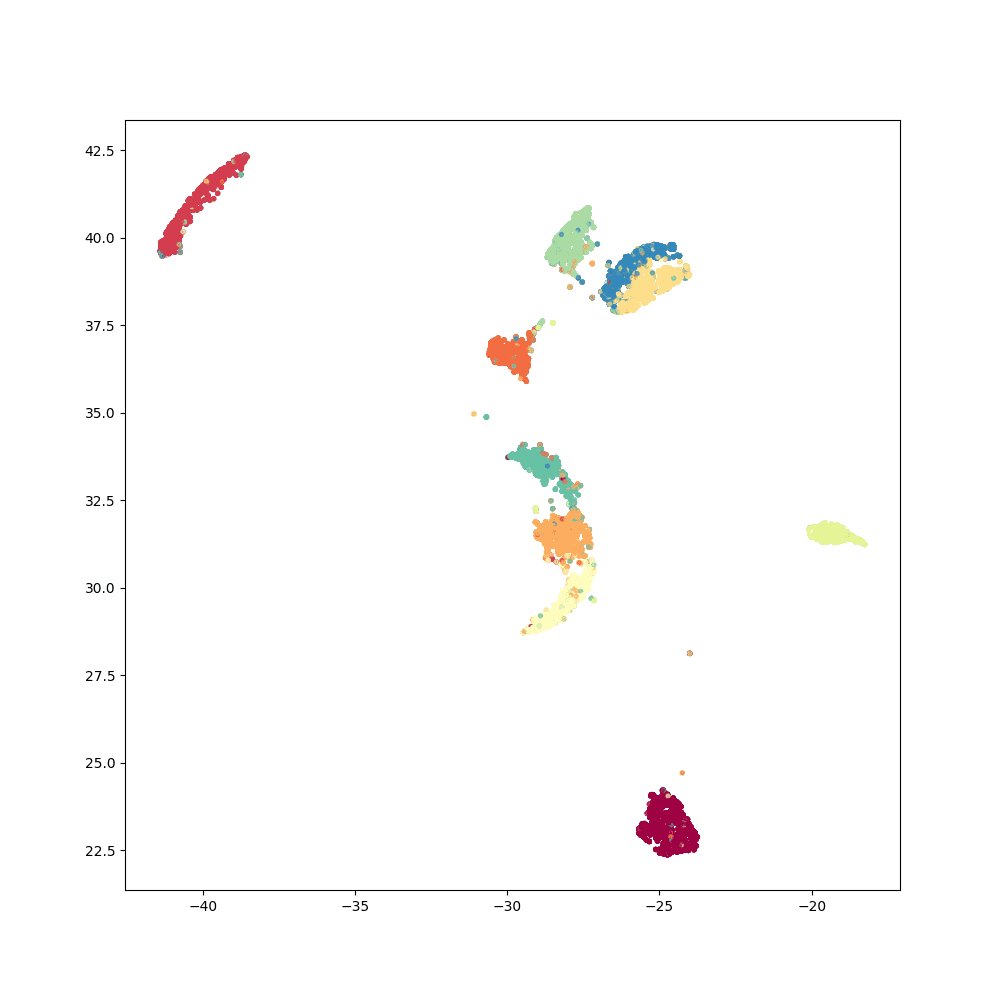} \\
     
     \begin{turn}{90}SONG + Reinit\end{turn} &\includegraphics[width=0.18\linewidth]{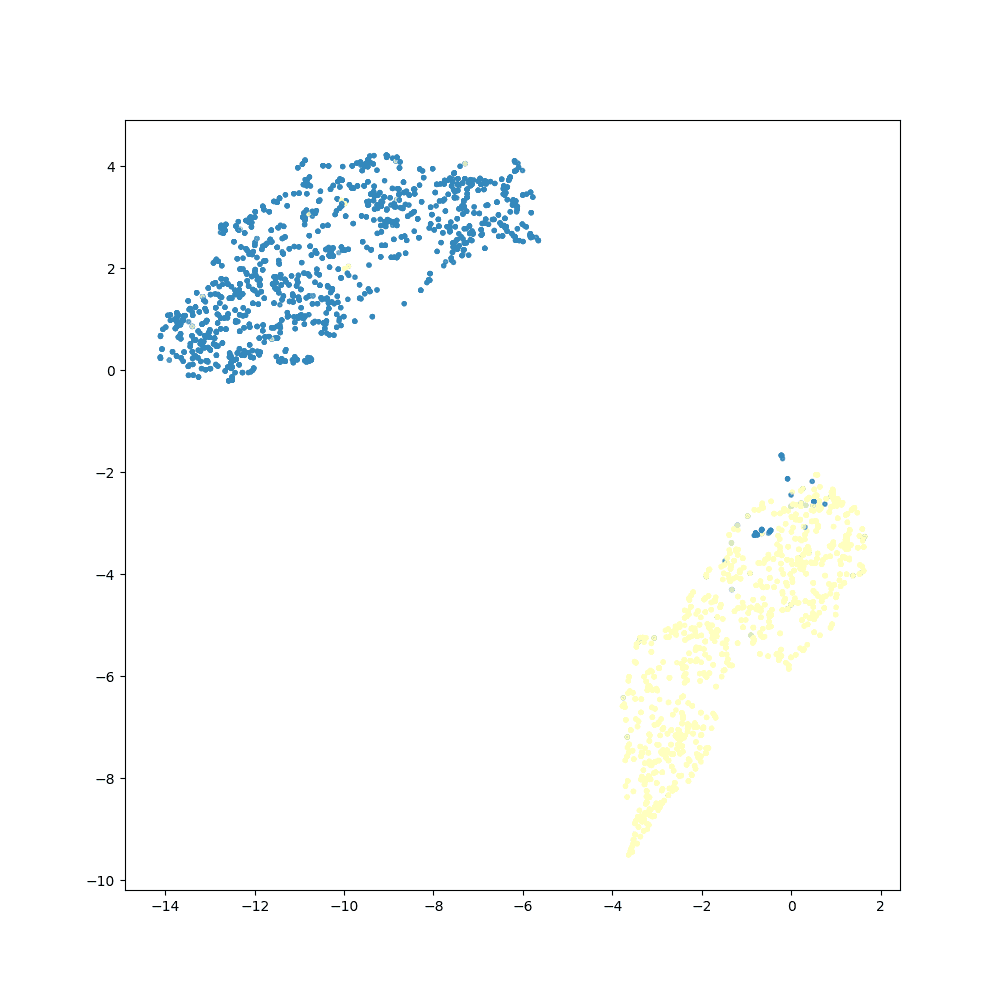} & \includegraphics[width=0.18\linewidth]{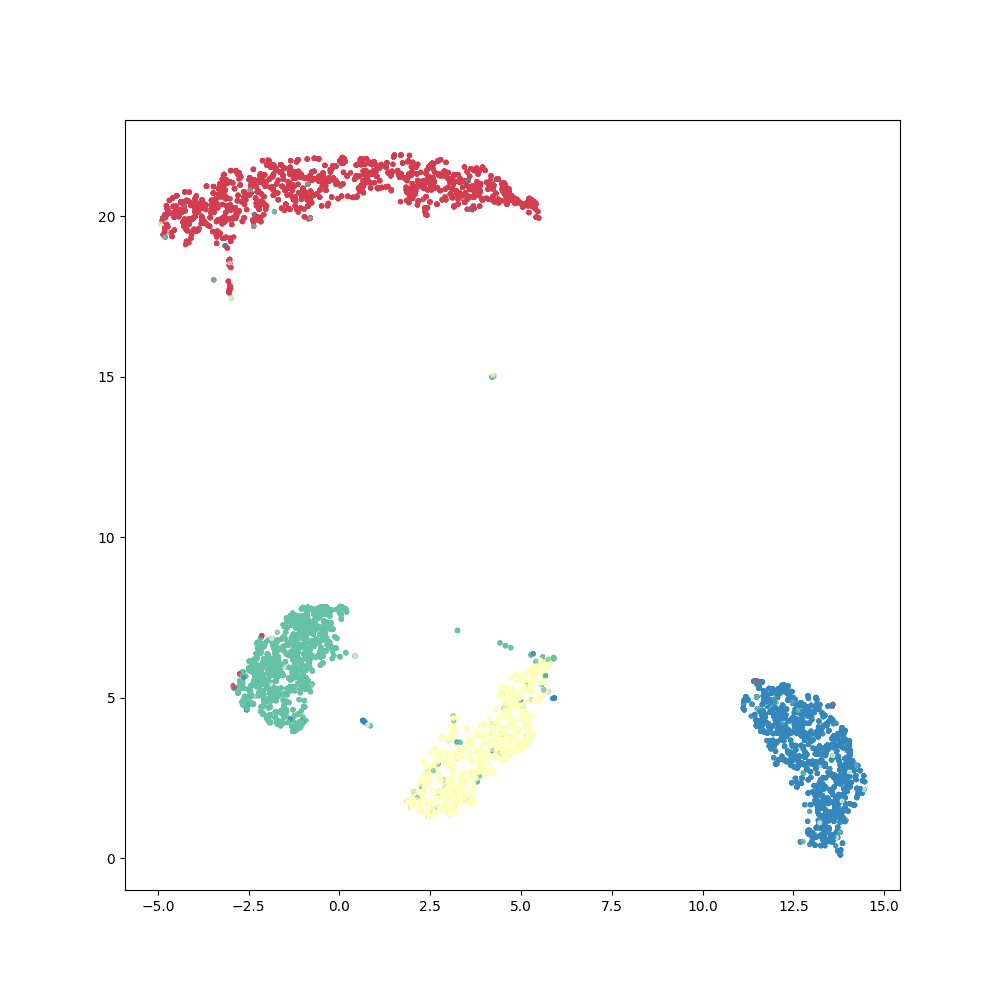} & \includegraphics[width=0.18\linewidth]{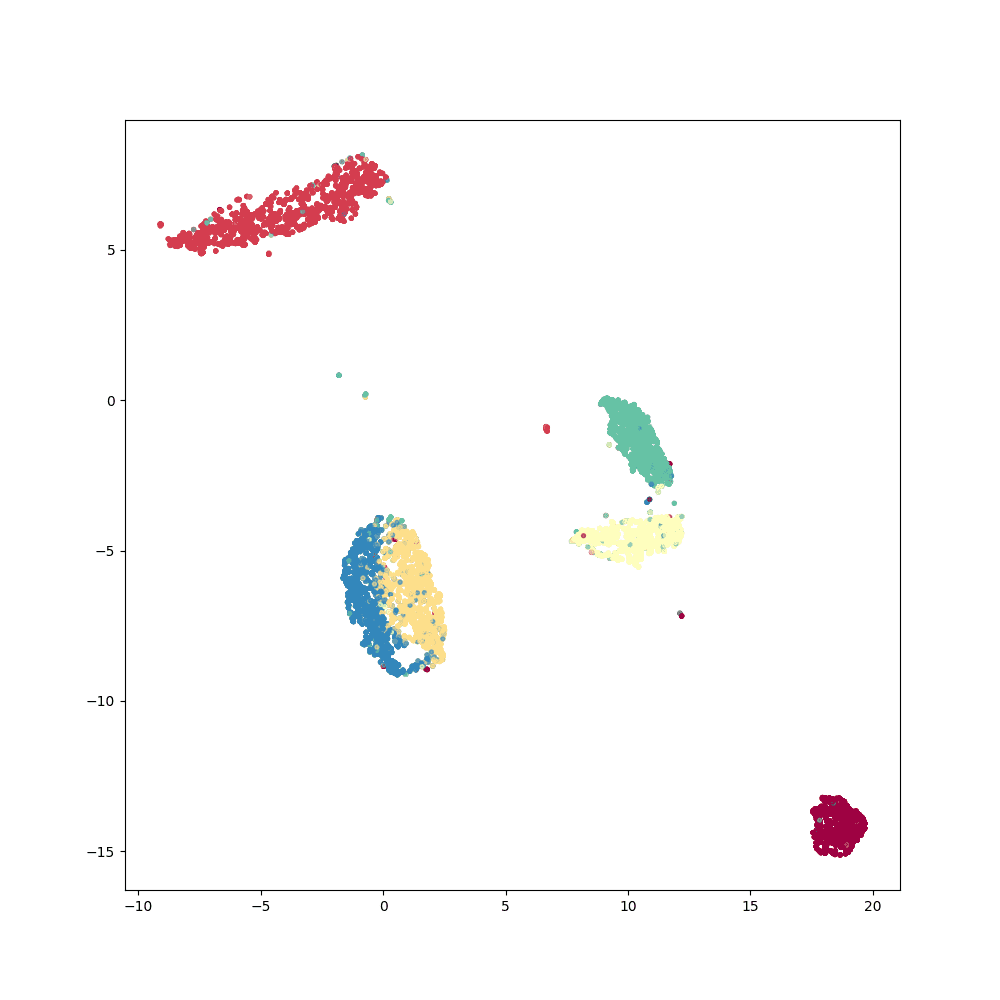} & \includegraphics[width=0.18\linewidth]{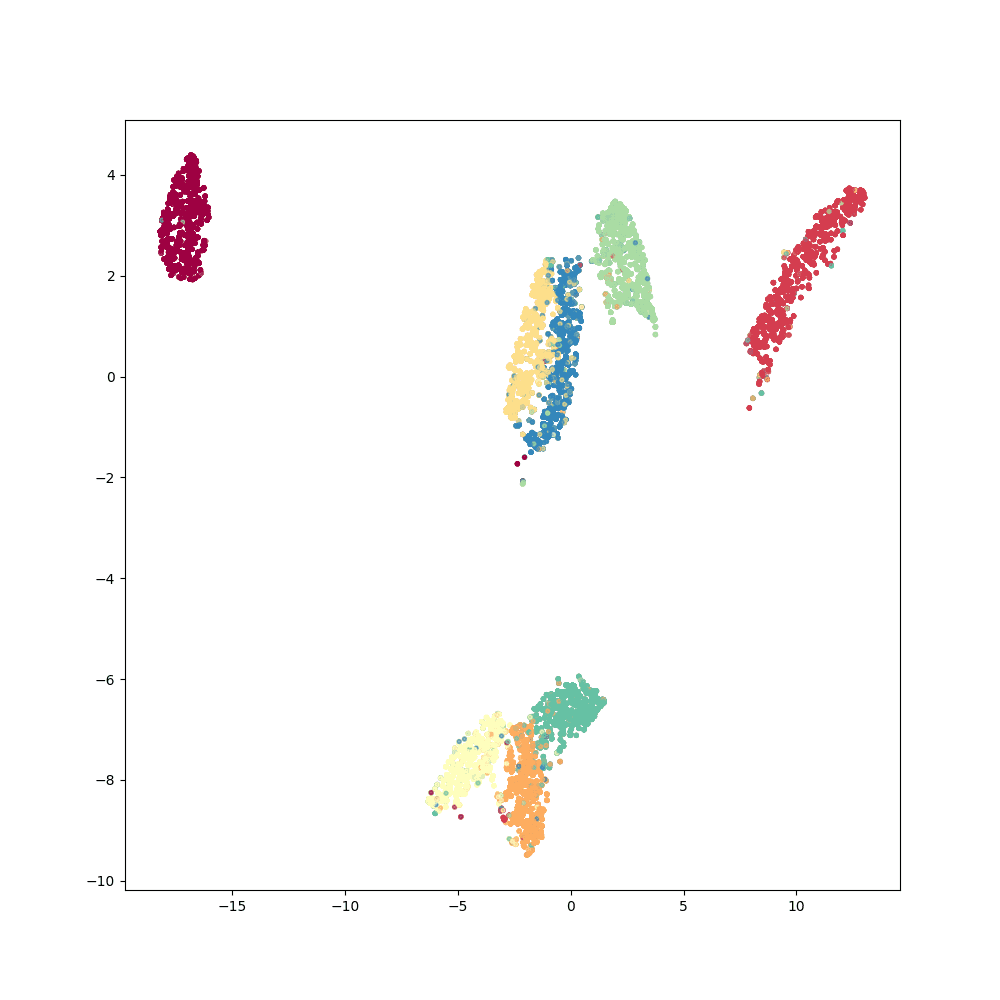} & \includegraphics[width=0.18\linewidth]{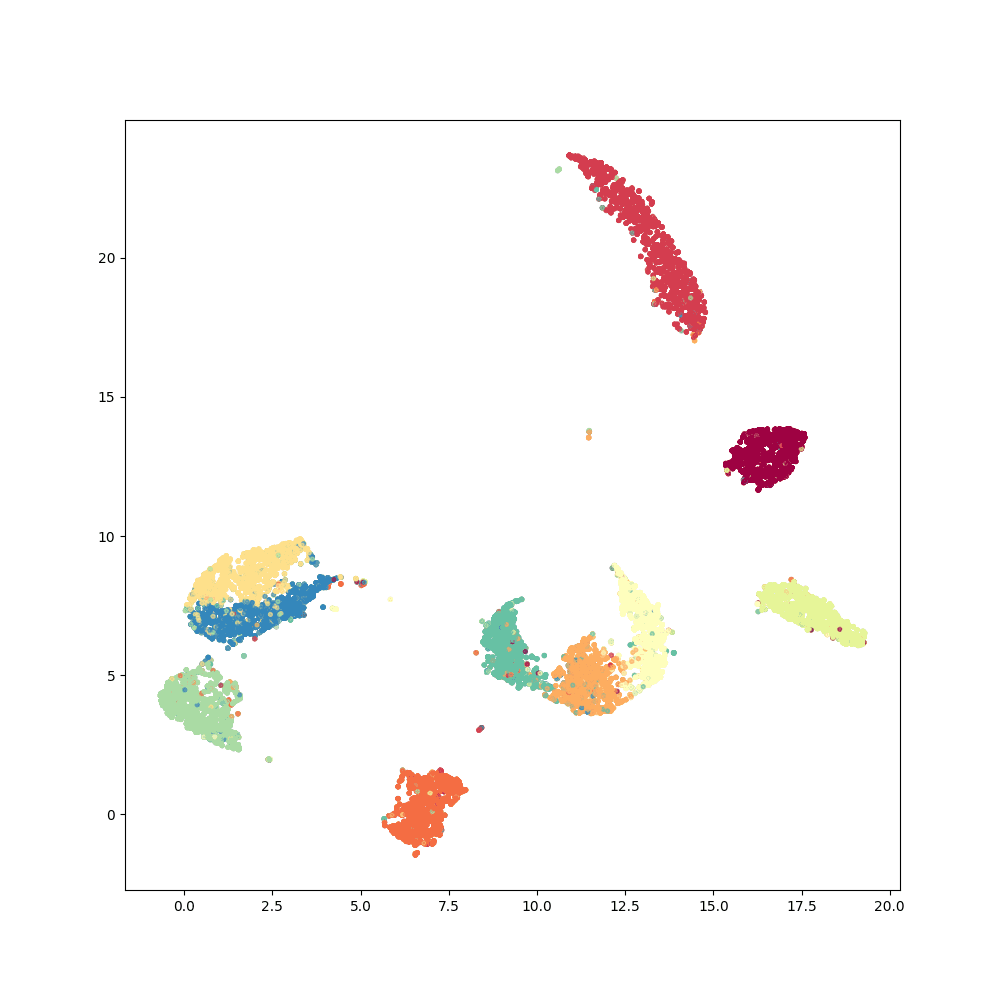} \\
     
     \begin{turn}{90}Parametric t-SNE\end{turn} &\includegraphics[width=0.18\linewidth]{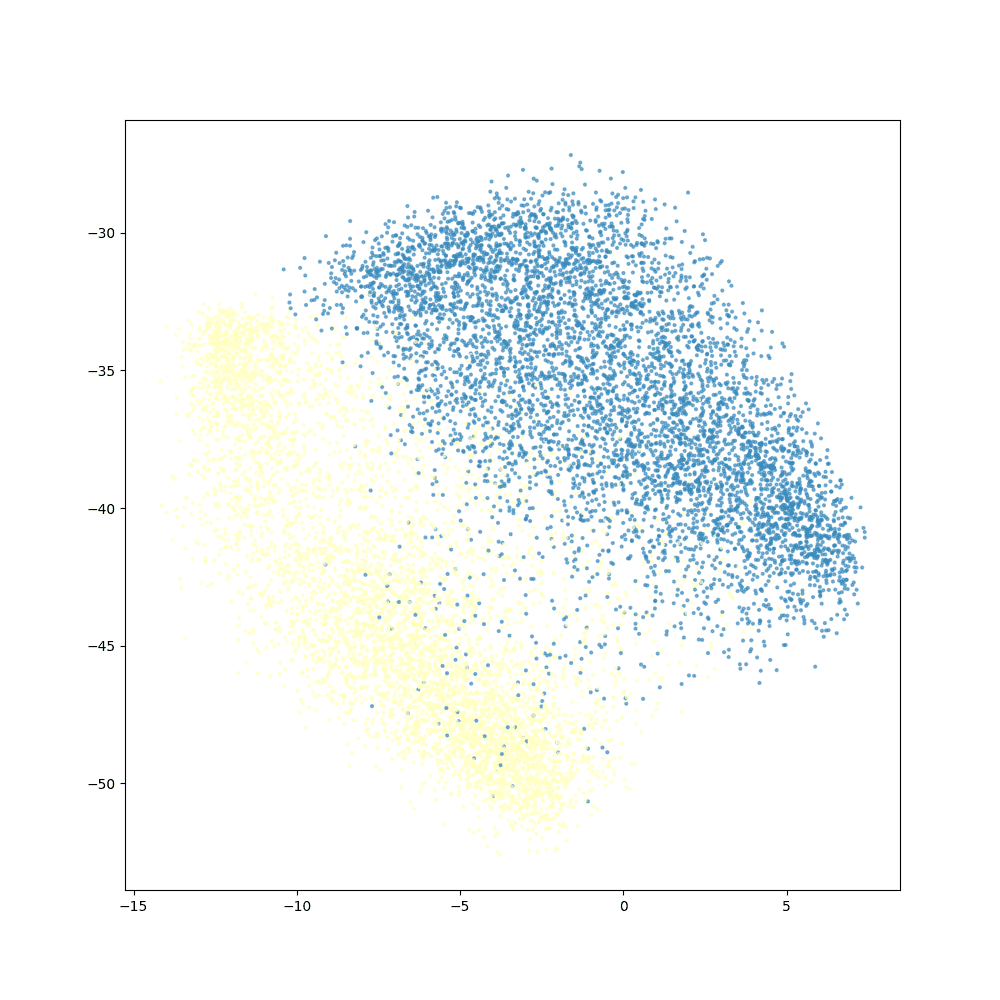} & \includegraphics[width=0.18\linewidth]{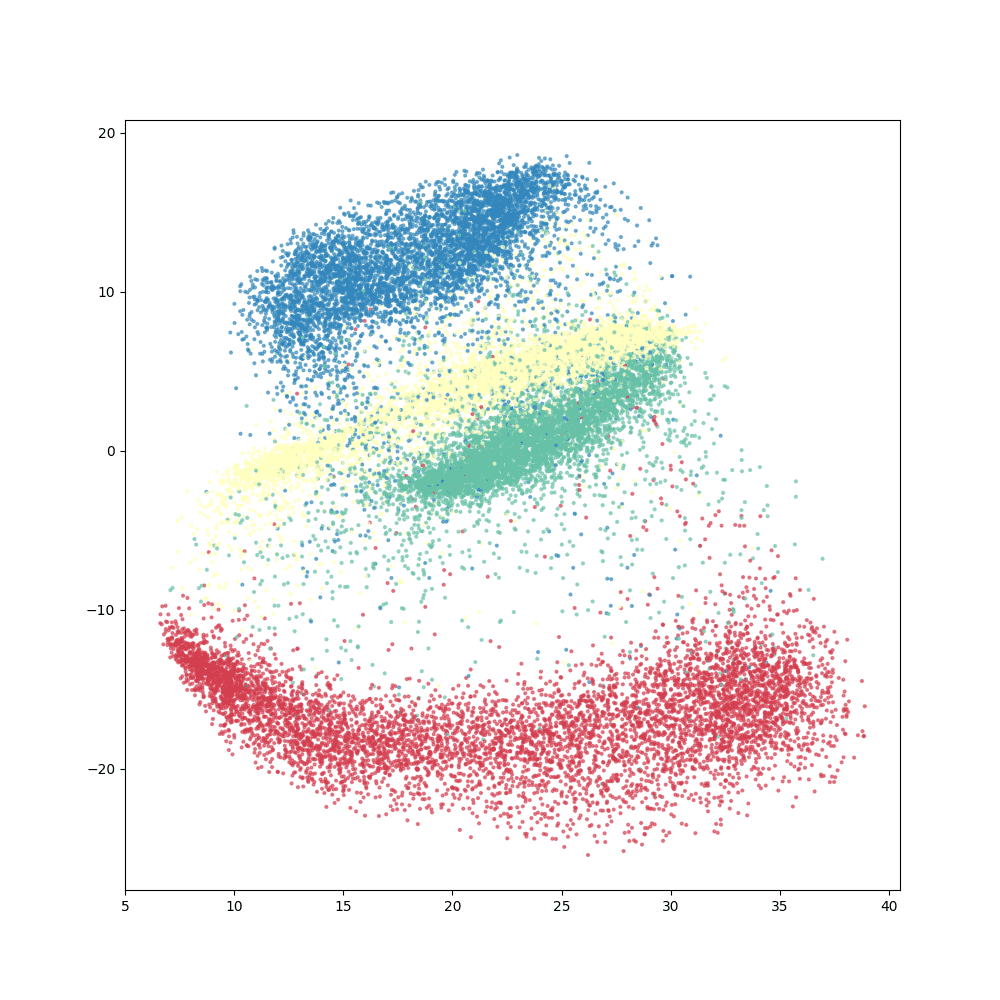} & \includegraphics[width=0.18\linewidth]{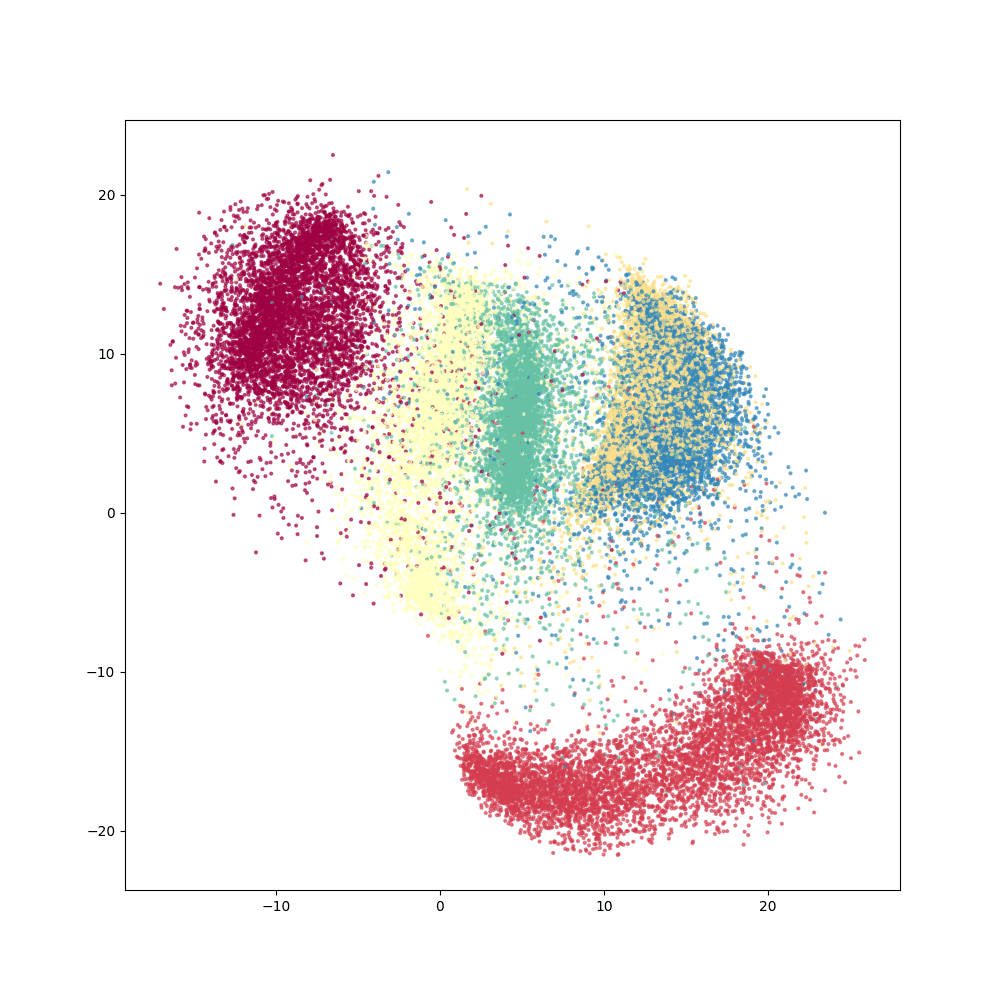} & \includegraphics[width=0.18\linewidth]{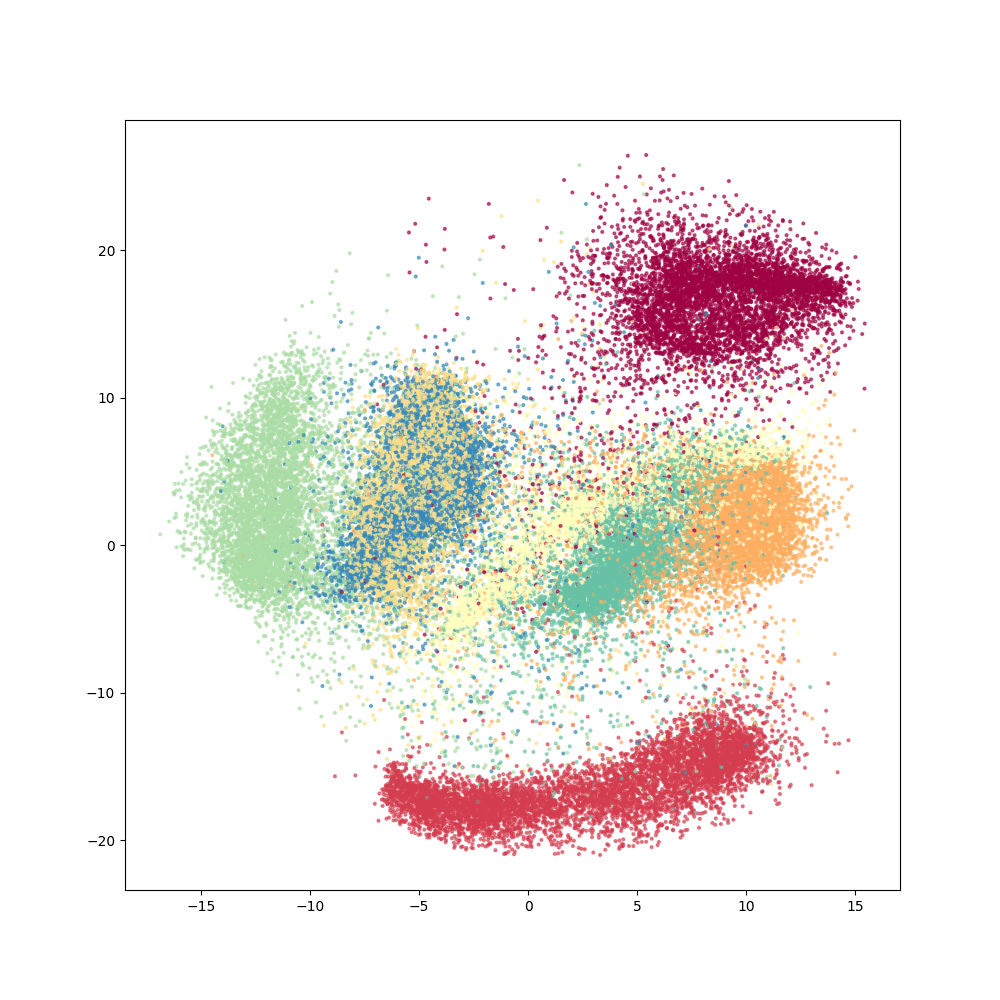} & \includegraphics[width=0.18\linewidth]{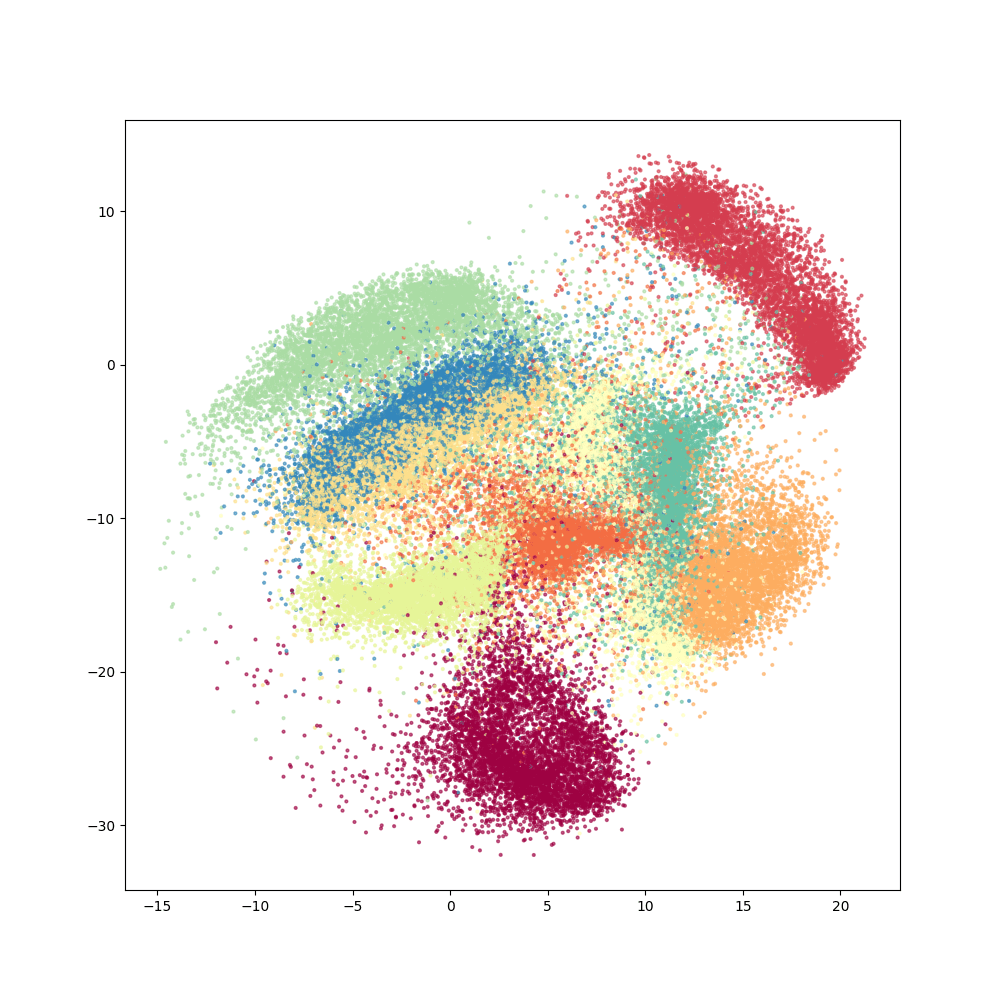} \\
     
     \begin{turn}{90}t-SNE\end{turn} &\includegraphics[width=0.18\linewidth]{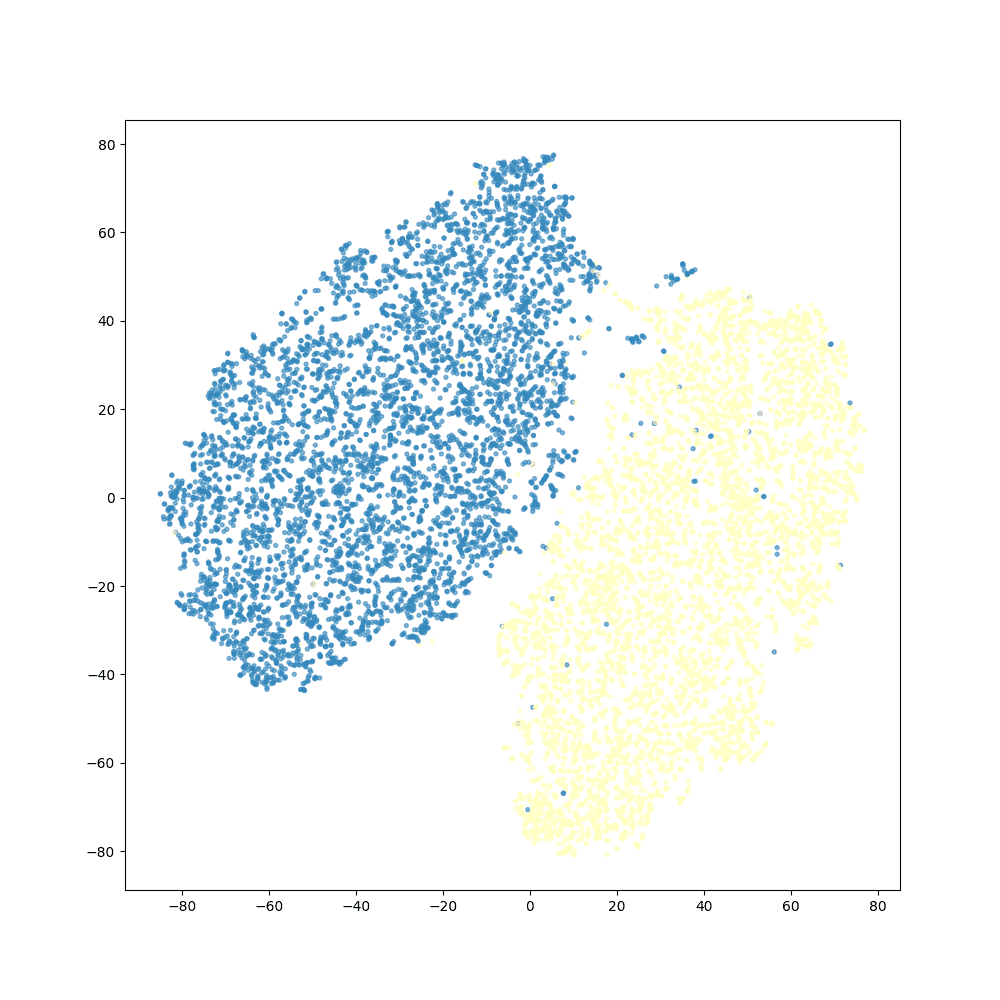} & \includegraphics[width=0.18\linewidth]{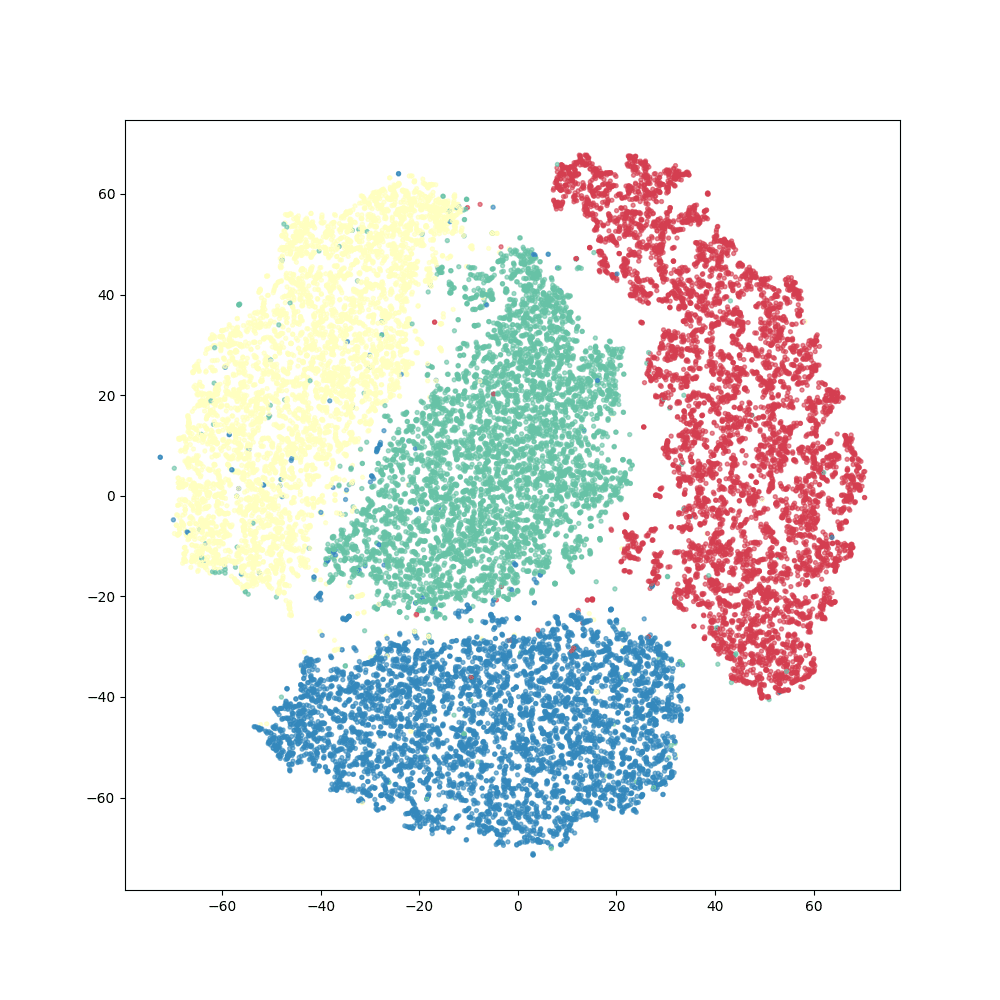} & \includegraphics[width=0.18\linewidth]{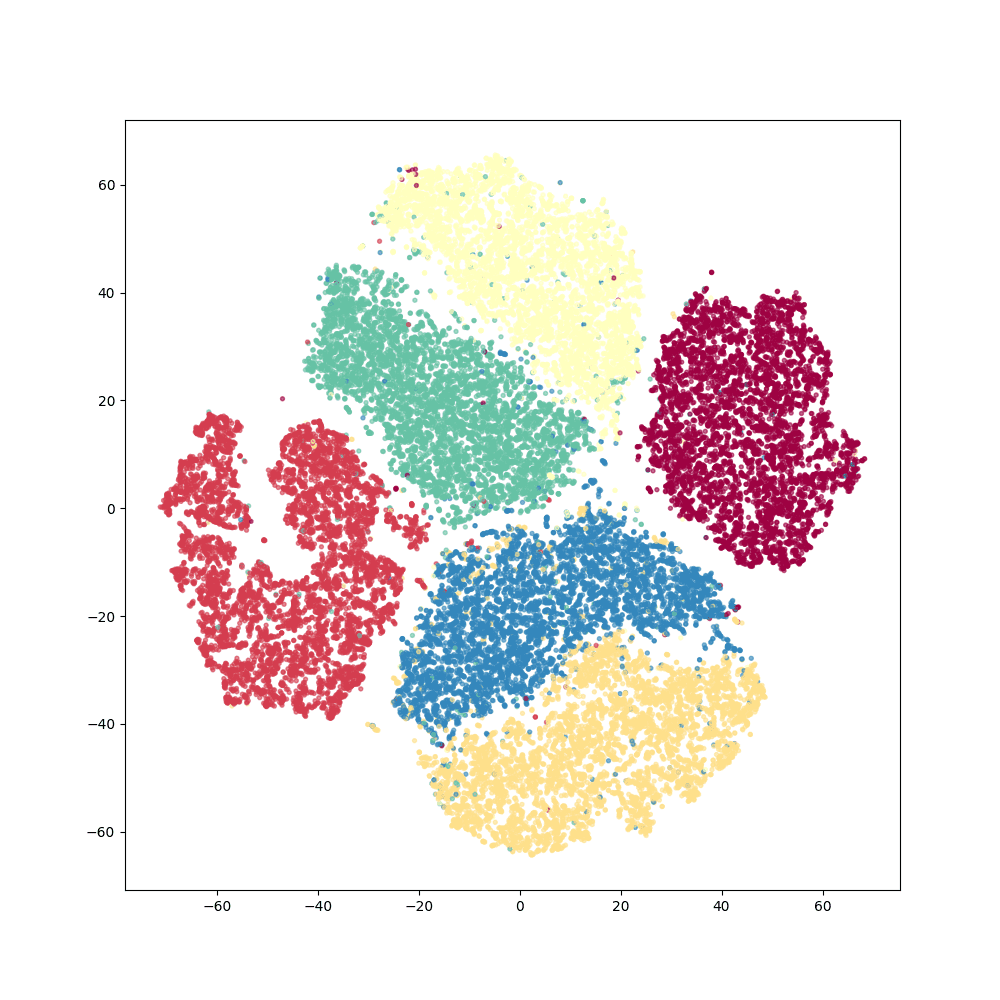} & \includegraphics[width=0.18\linewidth]{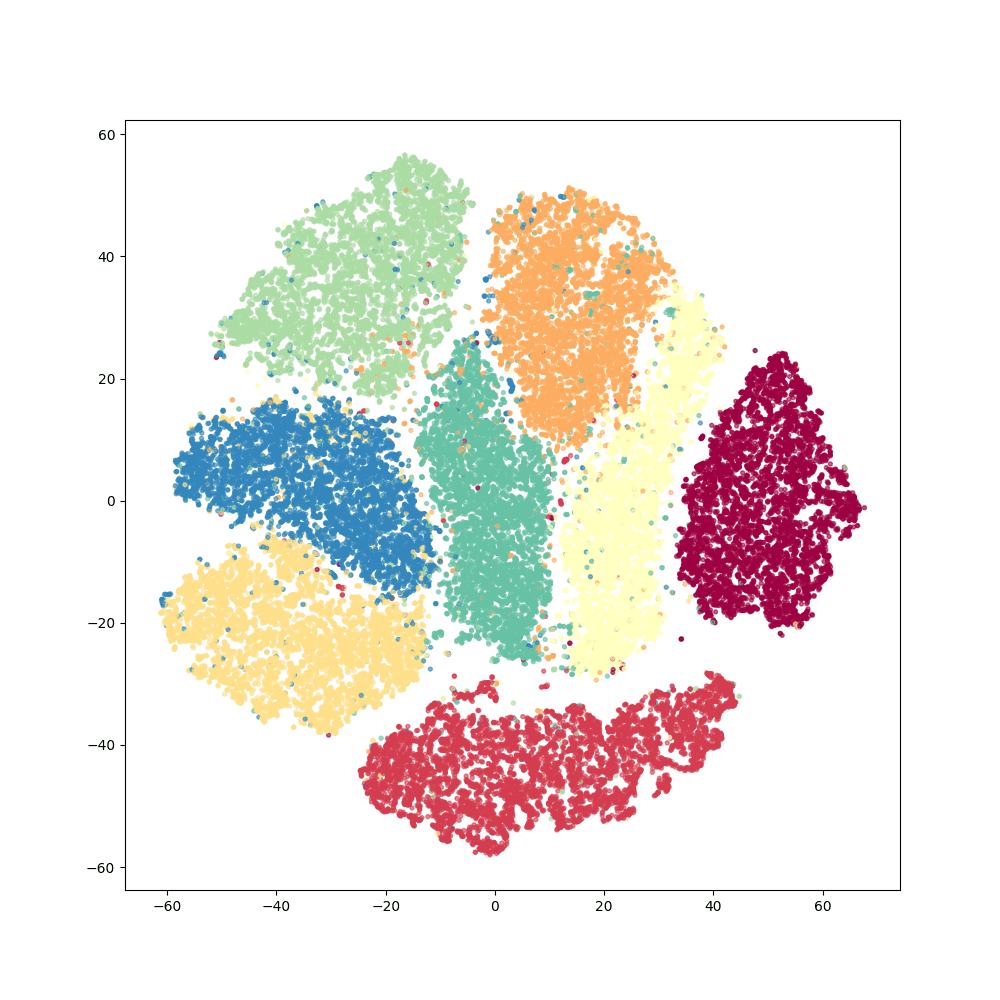} & \includegraphics[width=0.18\linewidth]{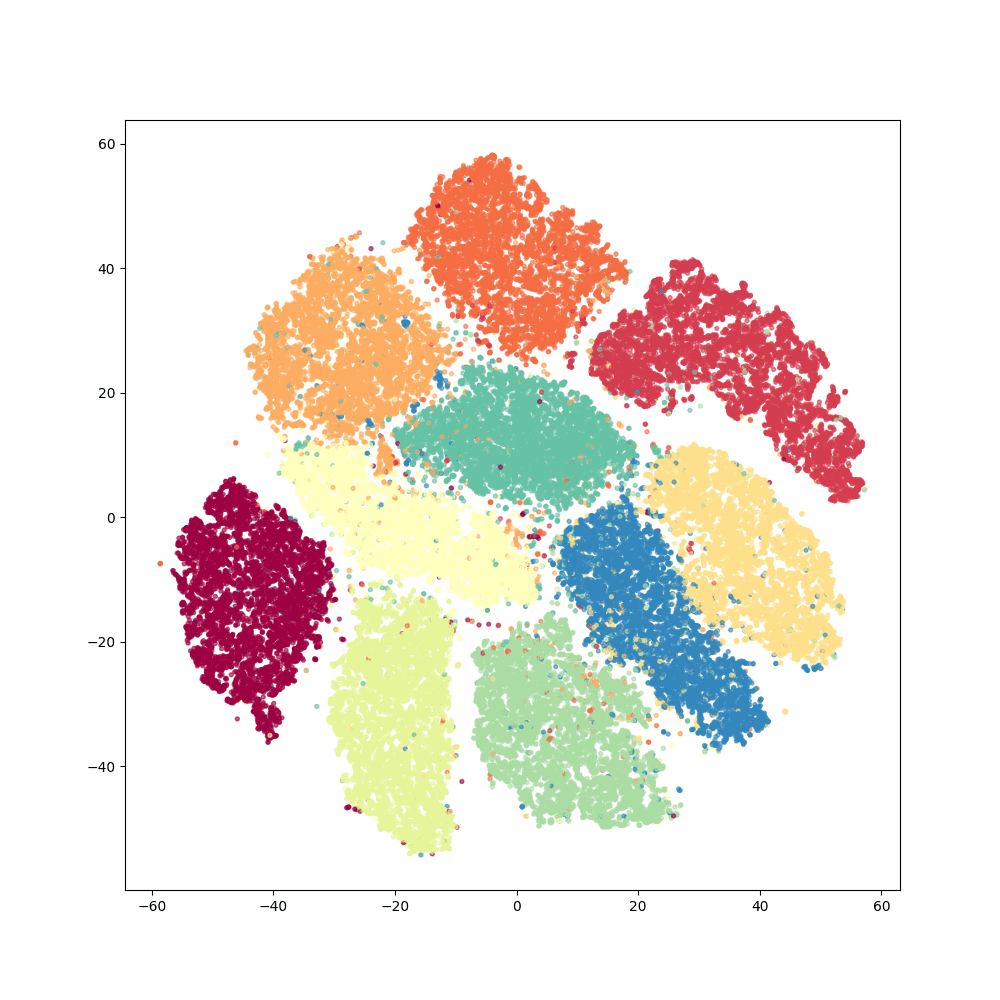} \\
     
     \begin{turn}{90}UMAP\end{turn} &\includegraphics[width=0.18\linewidth]{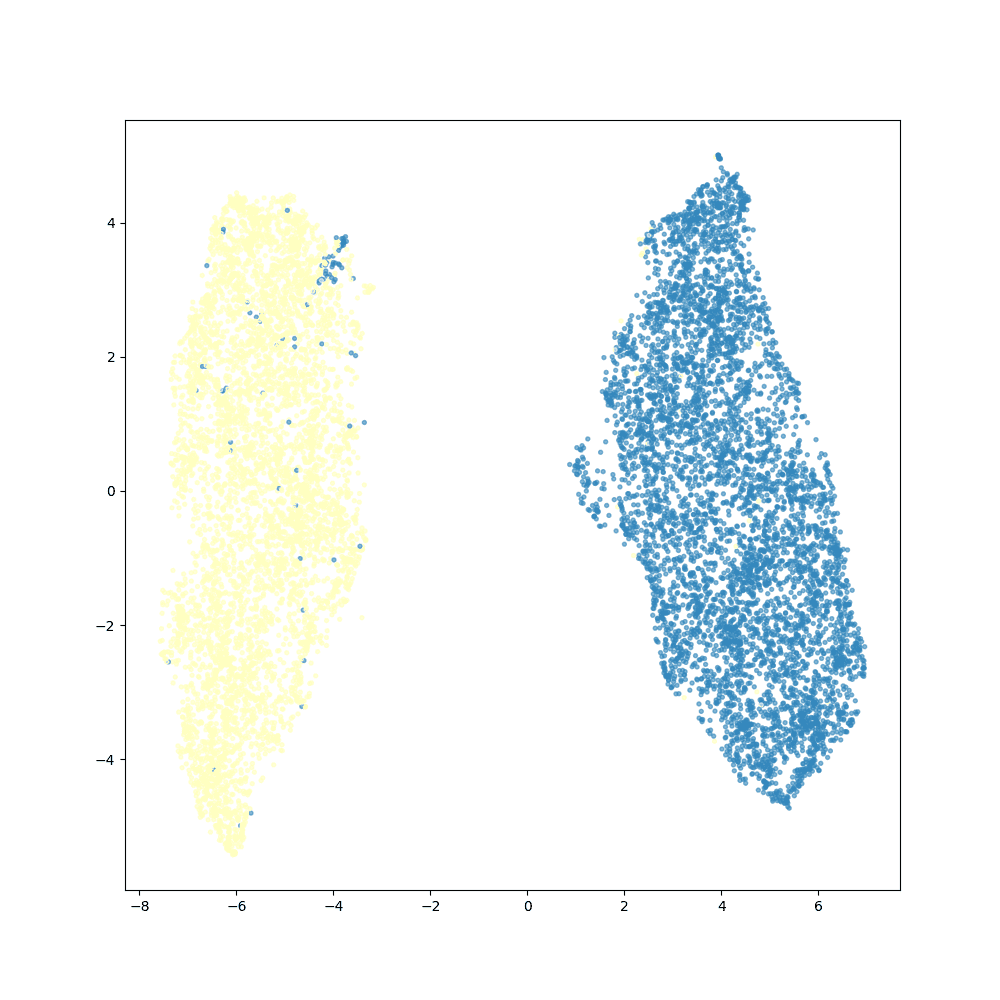} & \includegraphics[width=0.18\linewidth]{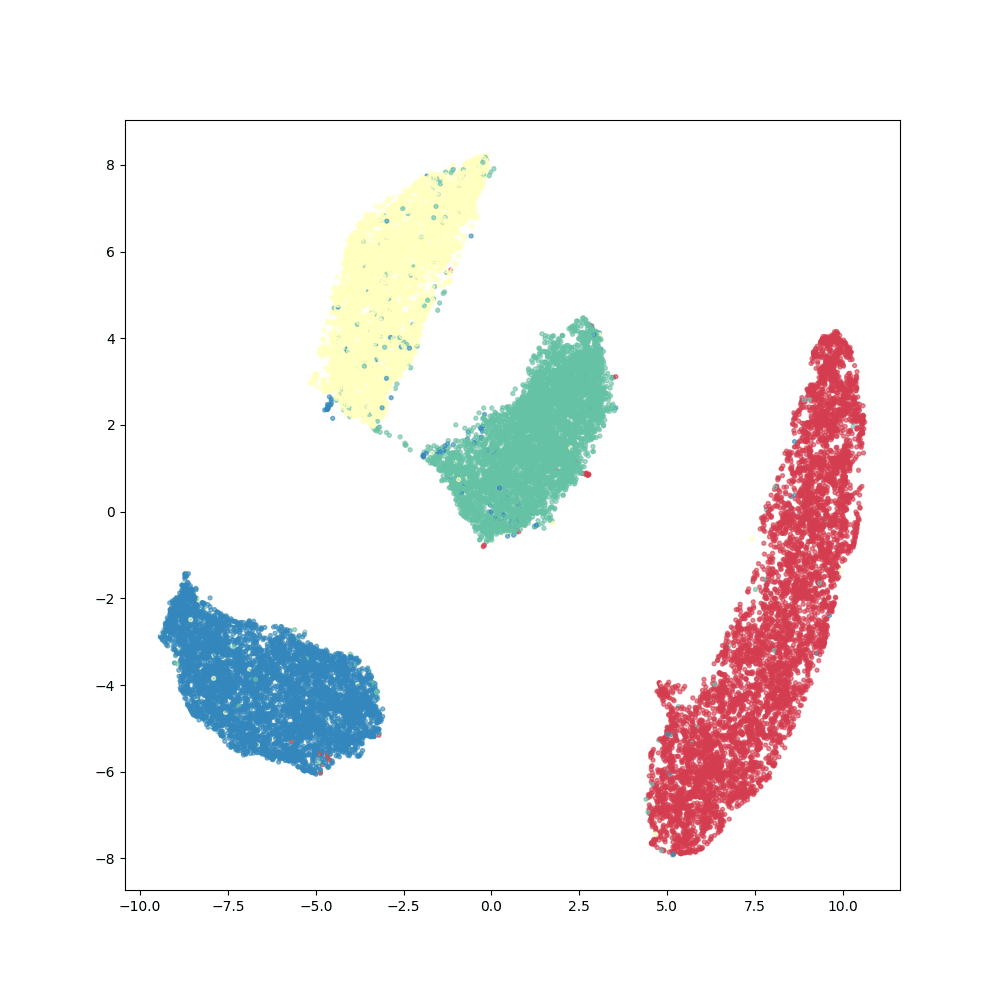} & \includegraphics[width=0.18\linewidth]{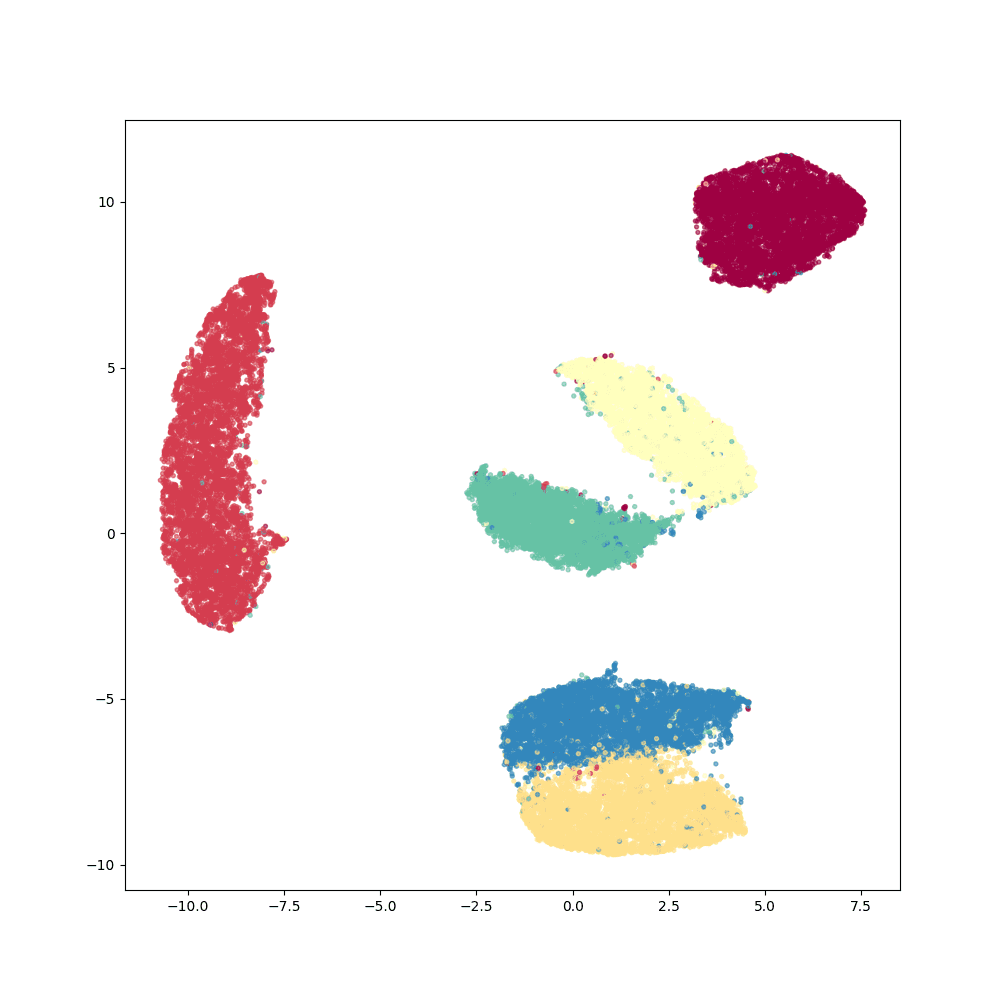} & \includegraphics[width=0.18\linewidth]{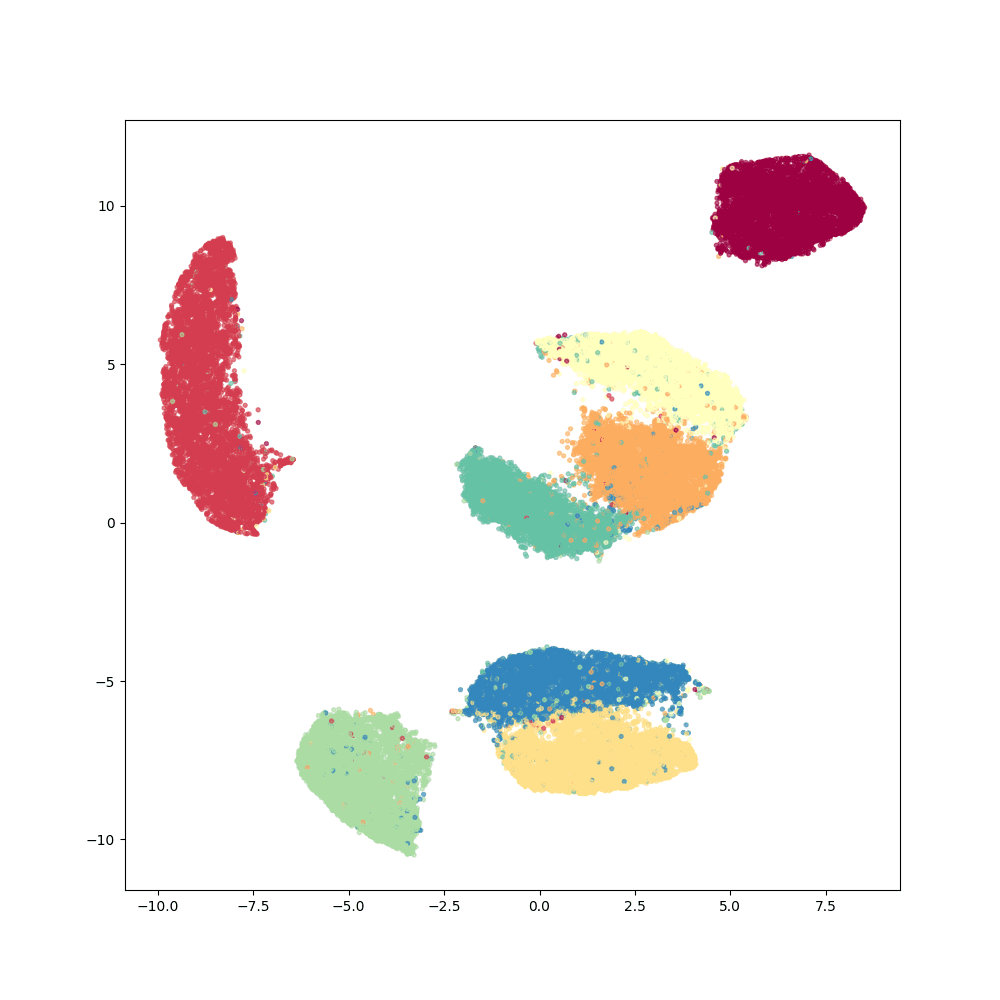} & \includegraphics[width=0.18\linewidth]{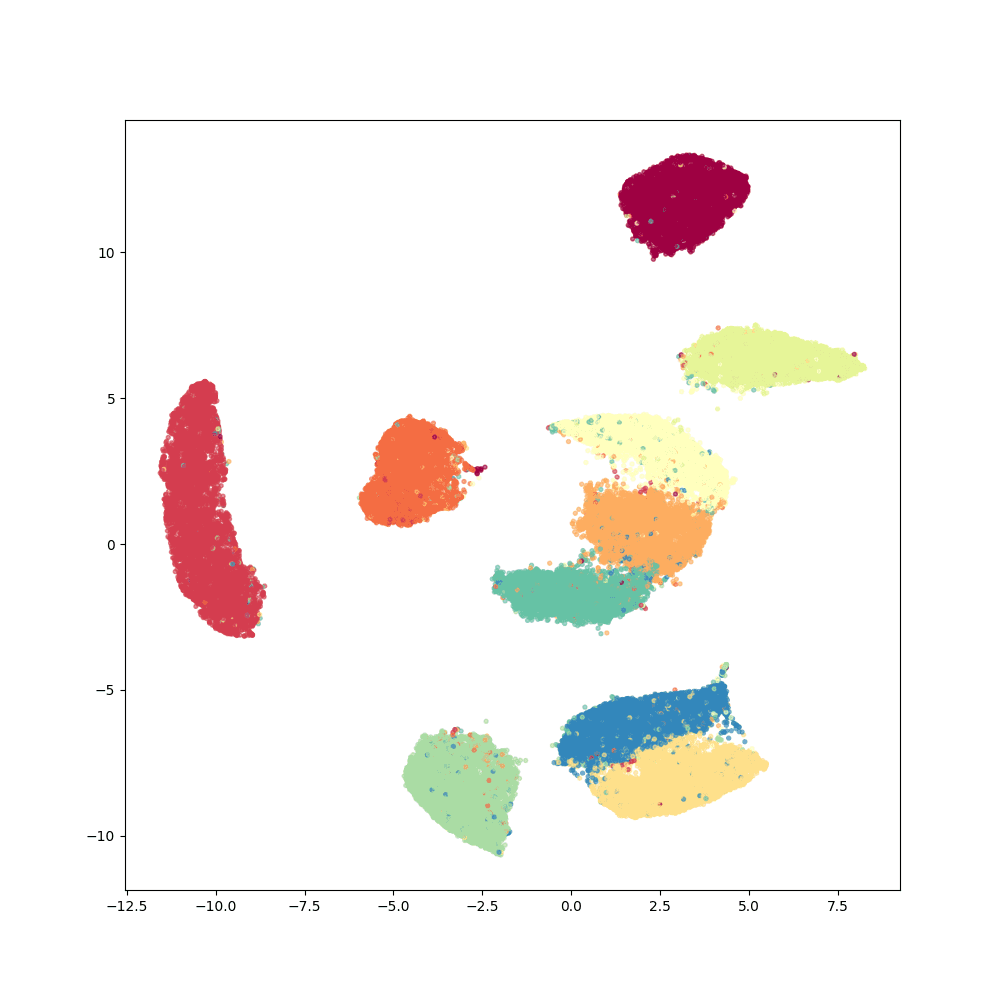} \\

     \end{tabular}
\caption{     \label{mnist_biased}
Incremental visualization of the MNIST dataset using SONG, SONG + Reinit, Parametric t-SNE, t-SNE and UMAP, where two classes are added at a time.}
\end{figure*}

\begin{table}[t]
\centering
\caption{AMI Scores on Heterogeneous Increments of Fashion MNIST and MNIST datasets. For fair comparison model-retaining methods (SONG and Parametric t-SNE) and model-reinitializing methods (SONG + Reinit, t-SNE and UMAP) are separated. The best AMI scores are highlighted.}
\label{Unbalanced-AMIS}
\begin{tabular}{p{1.8cm}|*{5}{p{0.6cm}}|*{5}{p{0.6cm}}}
\hline
                     & \multicolumn{5}{c|}{Fashion MNIST}                               & \multicolumn{5}{c}{MNIST}                                       \\
\textit{No. Classes} & \textit{2} & \textit{4} & \textit{6} & \textit{8} & \textit{10} & \textit{2} & \textit{4} & \textit{6} & \textit{8} & \textit{10} \\
\hline
SONG                 & \textbf{70.9}       & \textbf{86.1}       & \textbf{84}         & \textbf{71.2}       &\textbf{61.5}        & \textbf{88.4}       & \textbf{88.0}       & 7\textbf{9.2}       & \textbf{75.0}       & \textbf{81.0}        \\
Parametric t-SNE     & 50.3       & 80.6       & 76.1       & 60.2       & 57.8        & 39.2       & 58.8       & 60.8       & 56.8       & 59.3        \\
\hline
SONG + Reinit        & 70.9      & 76.8       & 78.4       & 69.1       & \textbf{61.0}        & 88.4       & 86.4       & 77.8       & 75.3       & 81.0        \\
t-SNE                & 14.2       & 56.0       & 59.9       & 57.3       & 56.3        & 89.3       & 67.0       & 72.1       & 71.2       & 73.8        \\
UMAP                 & 25.2       & 77.3       & 79.7       & 67.5       & 59.1        & 92.2       & 92.0       & 81.8       & 81.8       &\textbf{ 84.9}    \\
\hline
\end{tabular}
\end{table}

\textbf{Results}: For all three datasets: Wong (Fig. \ref{wong_biased}), Fashion MNIST (Fig. \ref{fashion_biased}) and MNIST (Fig. \ref{mnist_biased}), SONG shows the most stable placement of clusters when new data are presented, compared to parametric t-SNE and model-reinitalized methods, UMAP, t-SNE and SONG + Reinit. For the Wong dataset in Fig. \ref{wong_biased}, both SONG and Parametric t-SNE show similar cluster placements in the first two visualizations. However, Parametric t-SNE visualizations become unstable as more data are presented. In contrast, SONG provides consistently stable visualizations. On MNIST and Fashion MNIST datasets, the cluster placements provided by SONG have a noticeable change for the first two increments, but become more stable in their later increments. Parametric t-SNE shows a high level of cluster mixing in visualizations, which becomes more evident in later increments. Although UMAP shows similar relative placement of clusters at later increments for the MNIST and Fashion MNIST datasets, arbitrary rotations of the complete map are visible even for such visualizations. This may be due to UMAP using Spectral Embedding as the heuristic initialization instead of random initialization. Table \ref{Unbalanced-AMIS} summarizes the AMI Scores for the Fashion MNIST and MNIST datasets in heterogeneous incremental visualization scenarios.  Note that in Table \ref{Unbalanced-AMIS}, we have highlighted the best scores for each increment in the model-retaining methods. Since such incremental visualization for the model-reinitializing methods is not directly comparable with that of the model-retaining methods, we have only highlighted the winner for the complete dataset out of the model-reinitializing methods. In Table \ref{Unbalanced-AMIS}, SONG provides superior cluster purity than Parametric t-SNE, confirming our observations on the level of cluster mixing present in visualizations by Parametric t-SNE. SONG shows an average improvement of 14.98\% for Fashion MNIST and 49.73\% for MNIST in AMI compared to Parametric t-SNE. We observe that out of the non-parametric algorithms, SONG + Reinit is comparable but slightly inferior to UMAP, and superior to t-SNE. We also consider incremental visualizations with kernel t-SNE in Supplement Section 3.4. The results show that kernel t-SNE performs poorly when the new data is heterogeneous in distribution to the already observed data. 

For the Wong dataset (see Fig. \ref{wong_biased}), SONG + Reinit, Parametric t-SNE , t-SNE and UMAP all have drastic movement of clusters in consecutive visualizations. In t-SNE, we see a set of Gaussian blobs (possibly due to the Gaussian distribution assumption), with no discernible structure of cluster placement as visible in SONG and UMAP.  Parametric t-SNE shows stable placement of clusters in the first two visualizations. However, when more data are presented, we see a high level of mixed clusters in the visualization. 

In the Fashion MNIST visualizations (Fig.\ref{fashion_biased}), we see drastic re-arrangement of placement when using SONG + Reinit, Parametric t-SNE, t-SNE and UMAP. We emphasize that SONG does not show rotations, as seen in visualizations provided by UMAP. 

In Fig. \ref{mnist_biased}, the hierarchy of clusters is more preserved in SONG and UMAP than in t-SNE and Parametric t-SNE for the MNIST dataset. We expect in low-dimensional embedding space, the distances between clusters should vary as not all pairs of clusters are equally similar to each other, e.g., ``1'' should be more similar to ``7" than to ``3" or to ``5". In the results of t-SNE, however, the clusters are separated by similar distances, thereby the results do not provide information about the varying degrees of similarity between clusters. For both UMAP and SONG, the distances separating the clusters vary as expected. Although parametric t-SNE shows similar placement of clusters with rotations or flips in the last two visualizations, the level of cluster mixing is relatively high.

\subsection{Visualization of Data with Homogeneous Increments}
\label{unbiased_exp}
In this section, we further examine how each method performs when the incrementally added data proportionally represent all classes and clusters, using the same three datasets: Wong, Fashion-MNIST and MNIST. 

\textbf{Setup}:  For Wong dataset, we first sample 10k random cells, and increment this sample to 20k, 50k and 327k with new randomly sampled cells. We select these numbers to investigate whether the incremental inference of topology can be achieved starting from a small number of samples. Similarly, for Fashion MNIST and MNIST datasets, we begin with a sample of 12k random images, and increment this sample to 24k, 48k and 60k images.

Since Fashion-MNIST and MNIST have known ground-truths, for each visualization, we again conduct a k-means clustering to investigate the separability of clusters in the visualization, and use AMI to evaluate the cluster quality.

Furthermore, for Fashion MNIST and MNIST datasets, we develop a metric called the \textit{consecutive displacement of $\mathbf{Y}$} (CDY), to quantify the preservation of cluster placement in two consecutive incremental visualizations. CDY is defined as follows. Initially, we apply each algorithm to 6000 randomly sampled images, and iteratively add 6000 more images to the existing visualization until we have presented all images in a dataset. At the $t$-th iteration, we record the visualizations of the existing data (without the newly added data) before and after the training with the 6000 new images, namely $\mathbf{Y}^{(t-1)}$ and $\mathbf{Y}^{(t)}$. Next we calculate the CDY of a point $\mathbf{y}_i$ in the existing visualization  $\mathbf{Y}^{(t-1)}$ as:
\[ \text{CDY}(\mathbf{y}_i) =  \Vert \mathbf{y}_i^{(t)} -\mathbf{y}_i^{(t-1)} \Vert \]
We record the average and standard deviations of CDY calculated for all points in the visualization. We note that the lack of a ground-truth to give us information about an accurate placement of clusters renders a similar analysis for the Wong dataset prohibitive.

\begin{figure*}[ht]
\centering
\begin{tabular}{l c c c c}
     & 10000 & 20000 & 50000 & 327000 \\
     \begin{turn}{90}SONG\end{turn} & \includegraphics[width=0.2\linewidth]{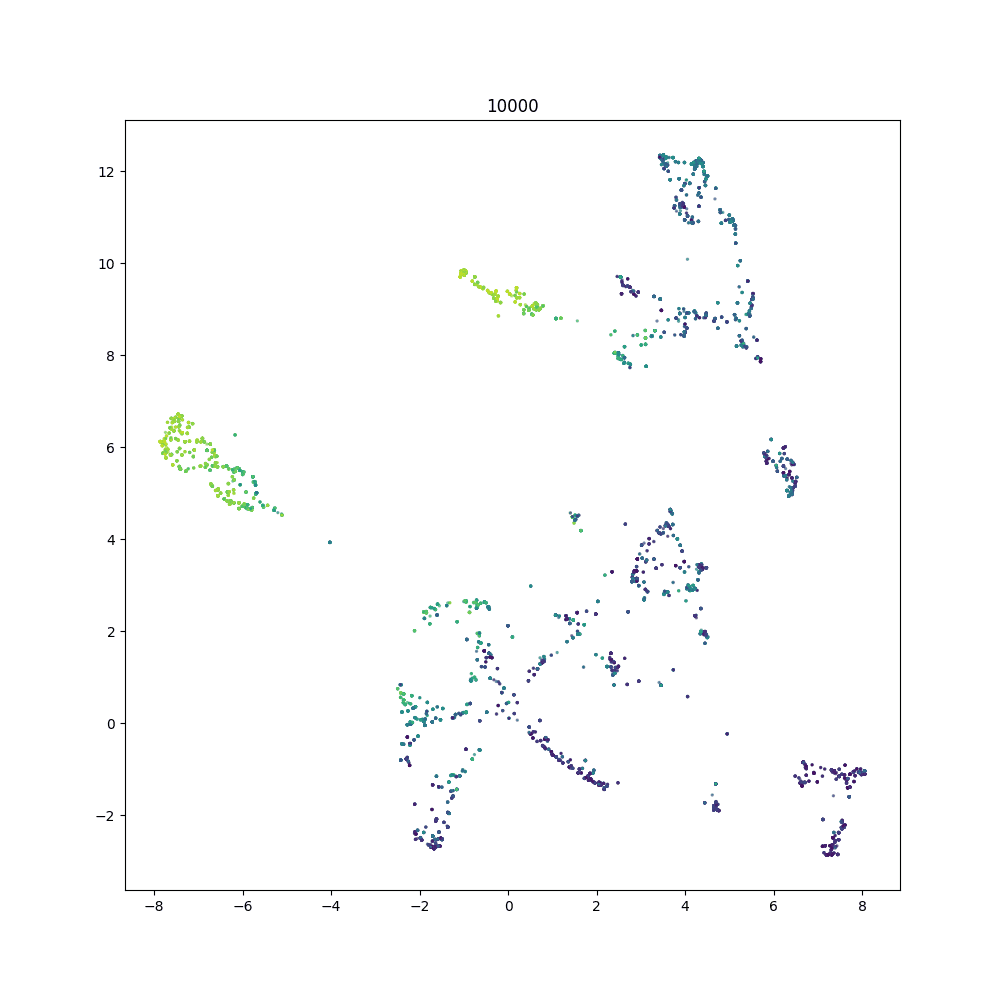} & \includegraphics[width=0.2\linewidth]{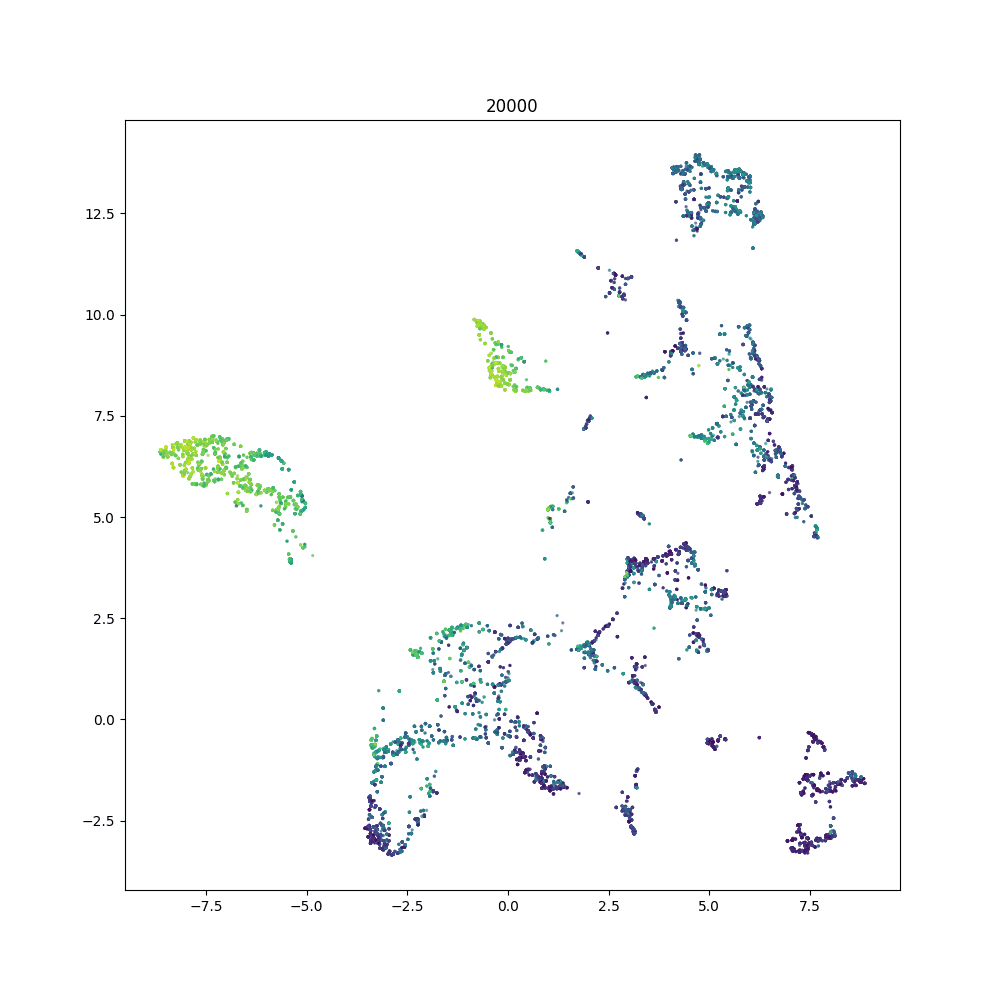} & \includegraphics[width=0.2\linewidth]{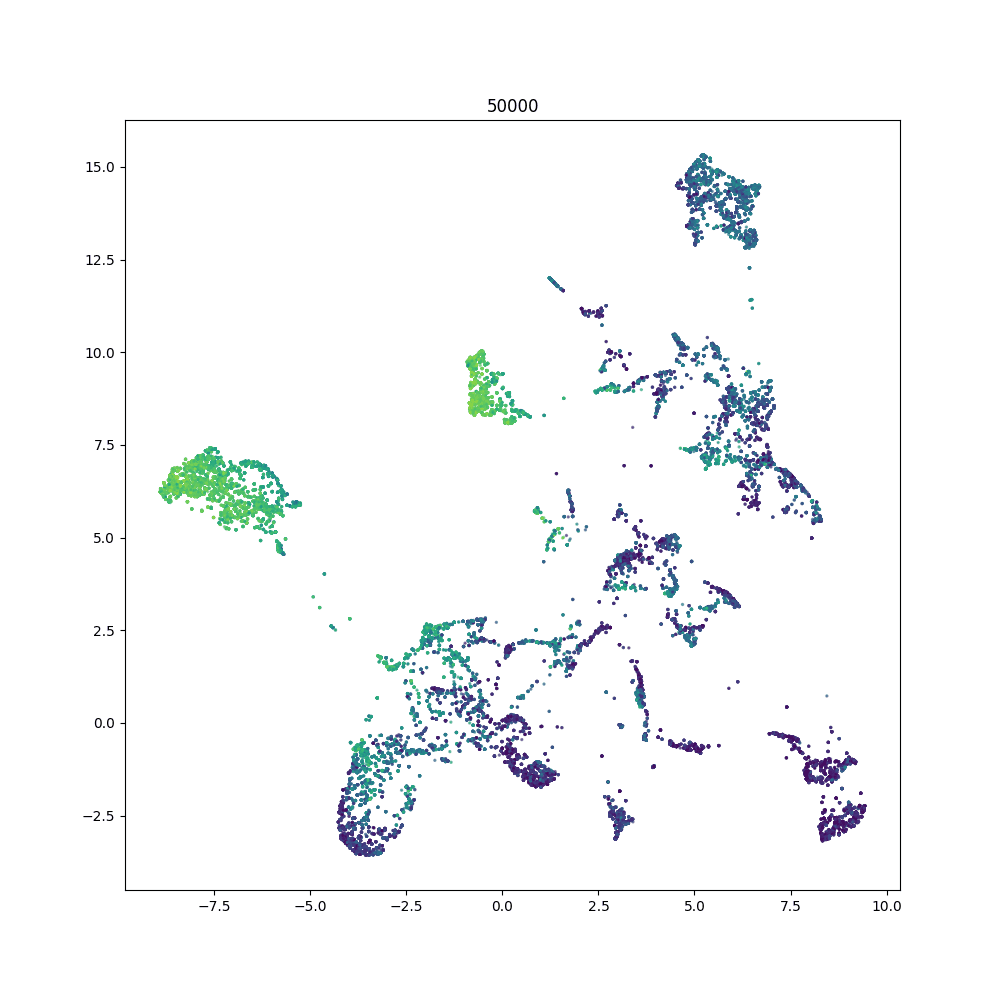} & \includegraphics[width=0.2\linewidth]{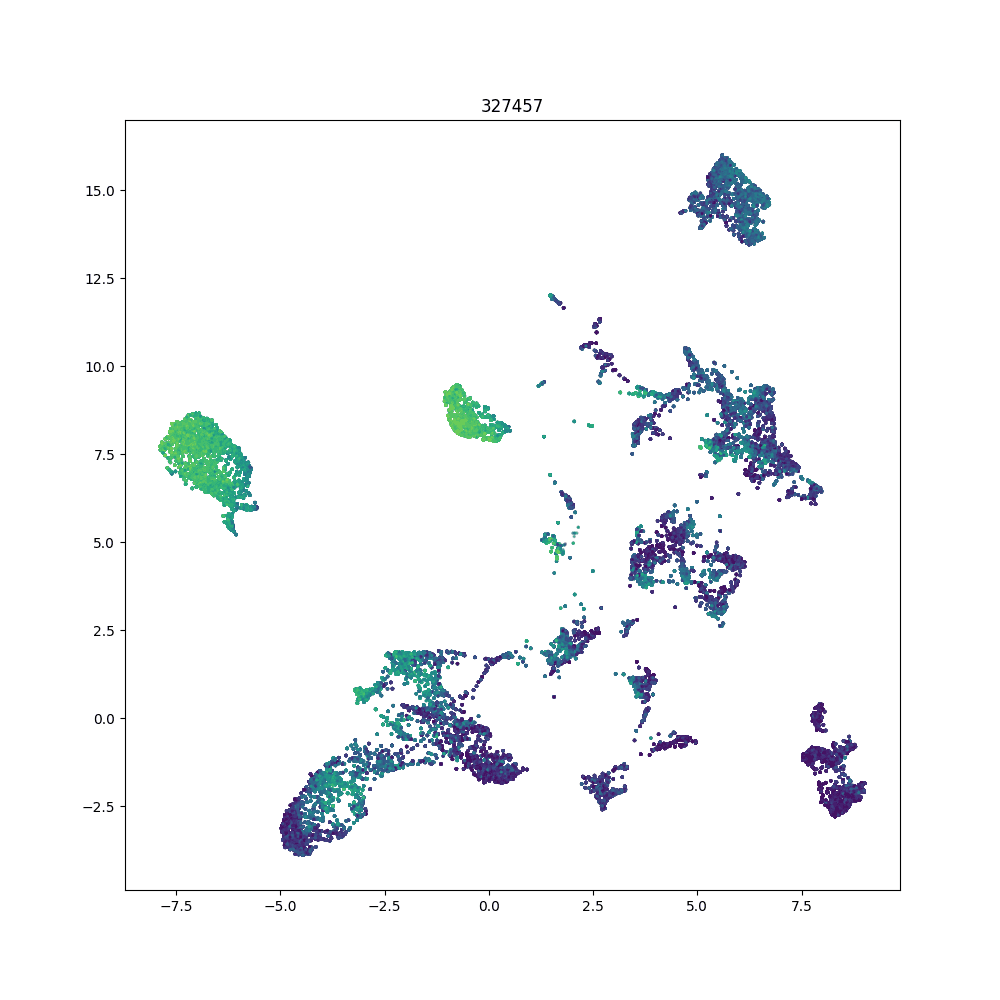} \\
     
     \begin{turn}{90}SONG + Reinit\end{turn} & \includegraphics[width=0.2\linewidth]{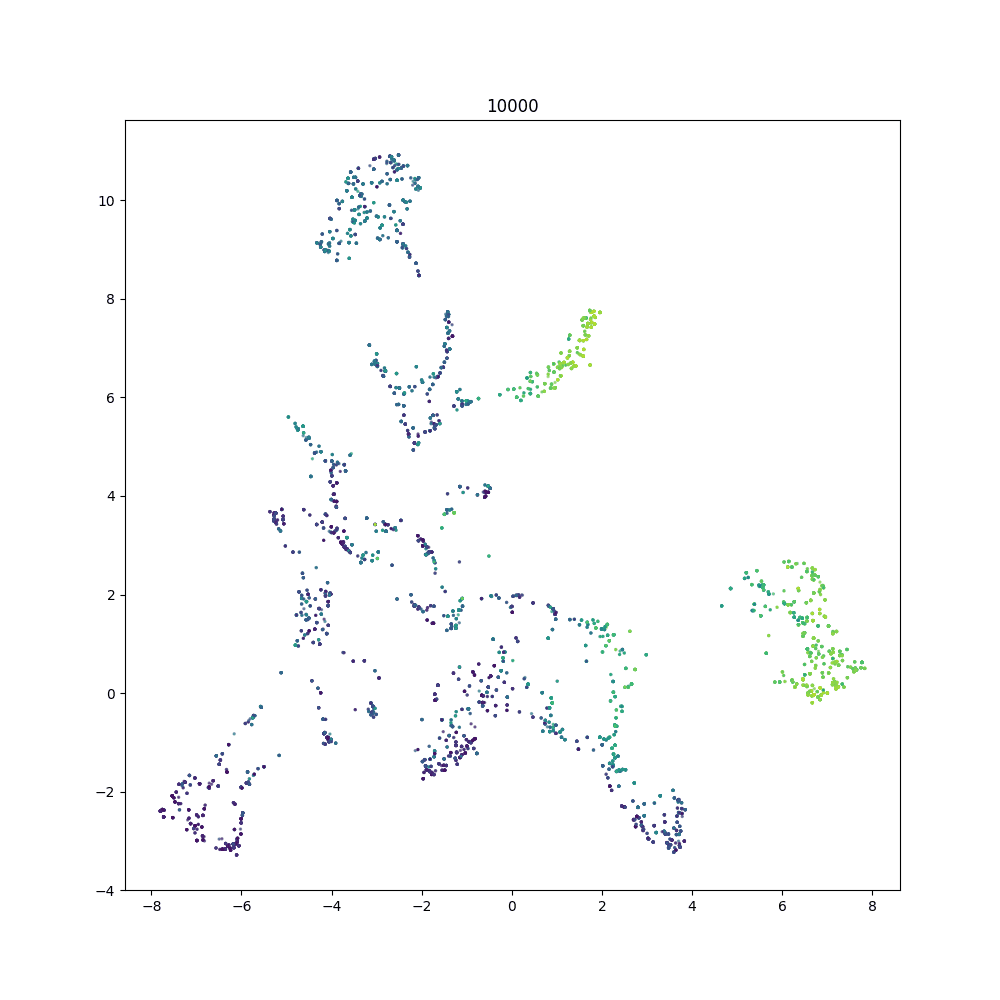} & \includegraphics[width=0.2\linewidth]{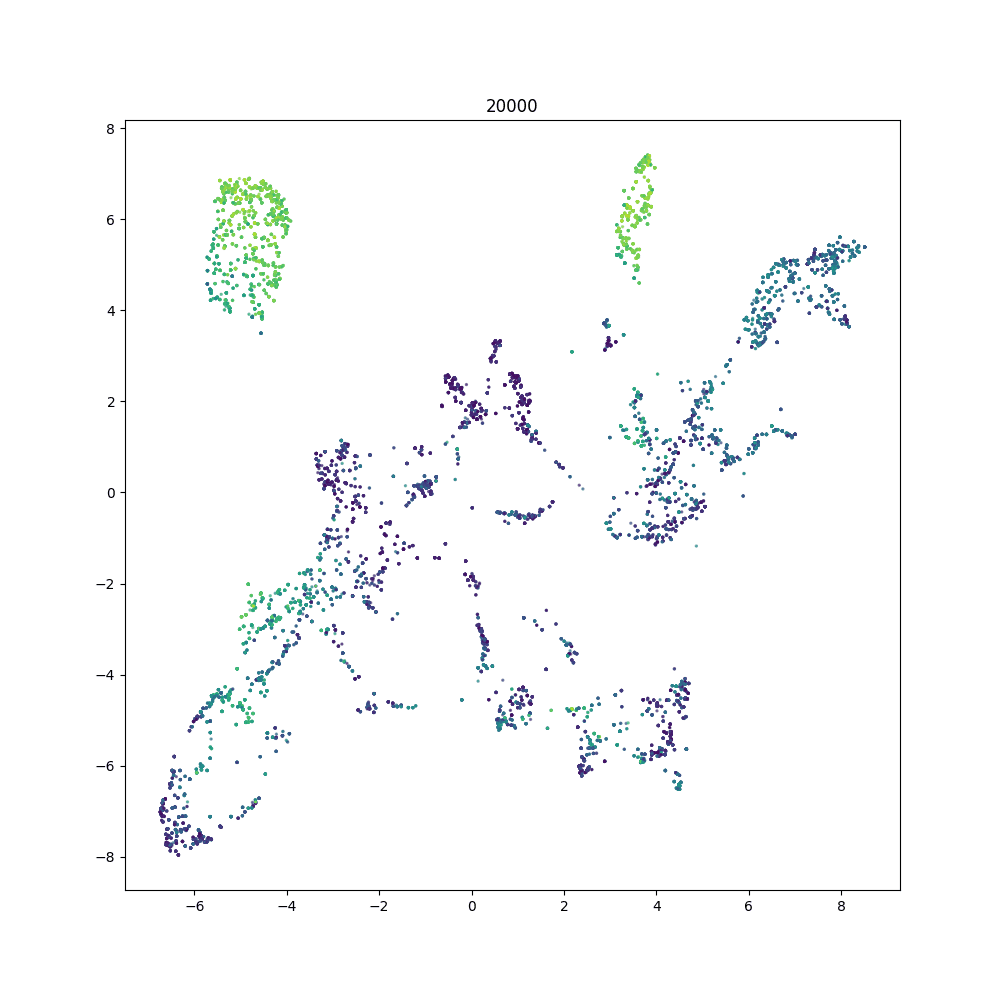} & \includegraphics[width=0.2\linewidth]{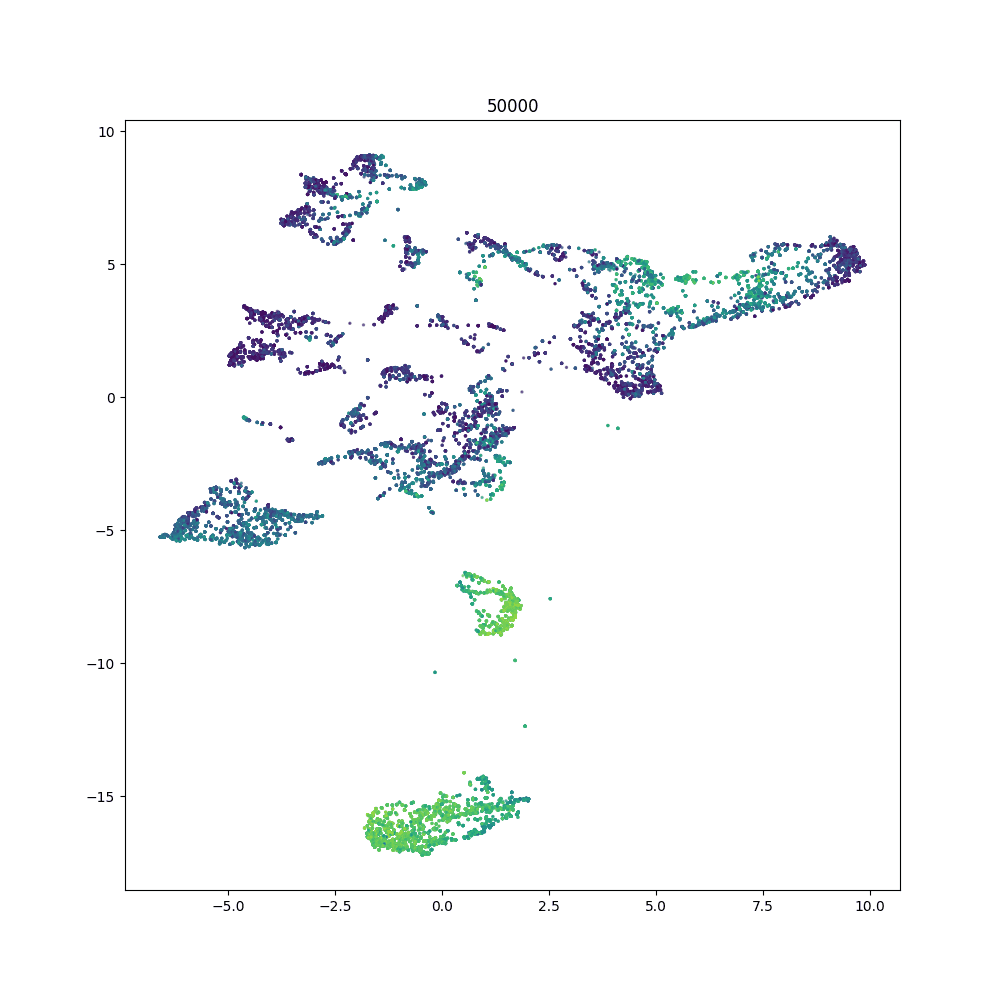} & \includegraphics[width=0.2\linewidth]{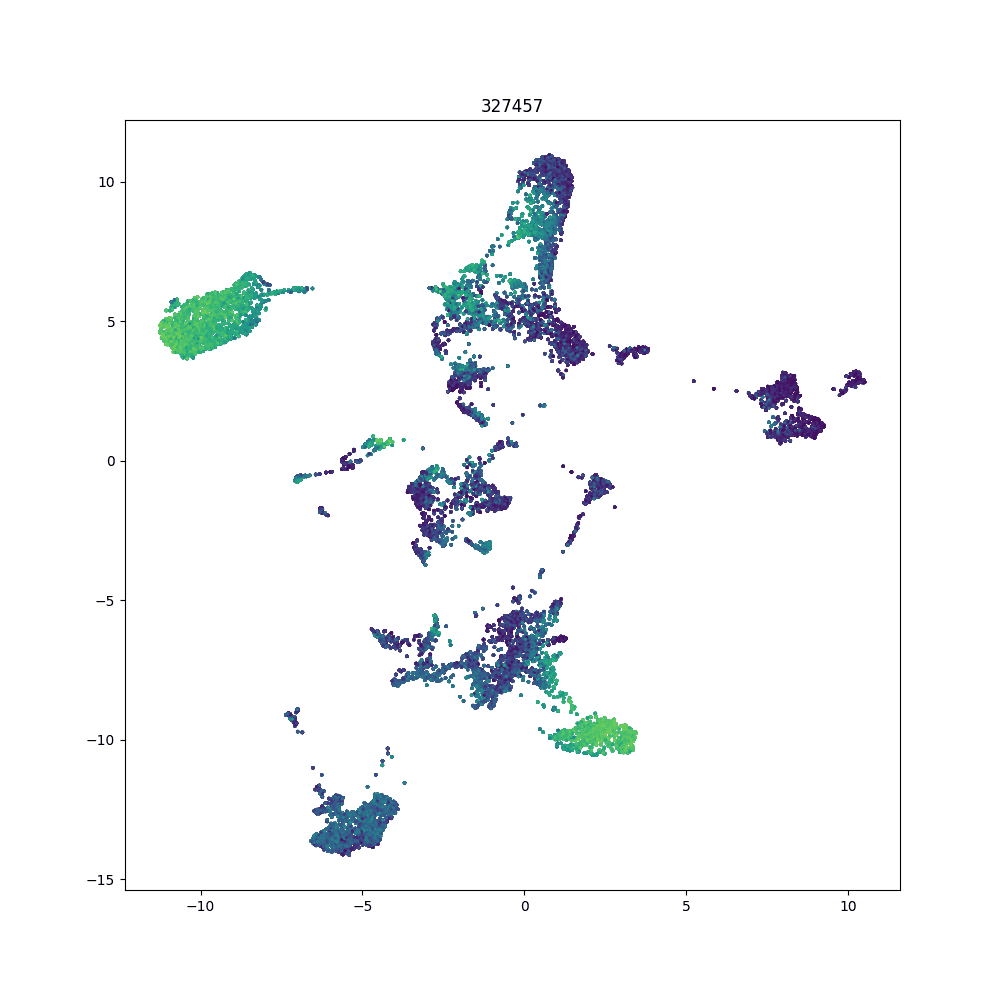} \\
     
     \begin{turn}{90}Parametric t-SNE\end{turn} & \includegraphics[width=0.2\linewidth]{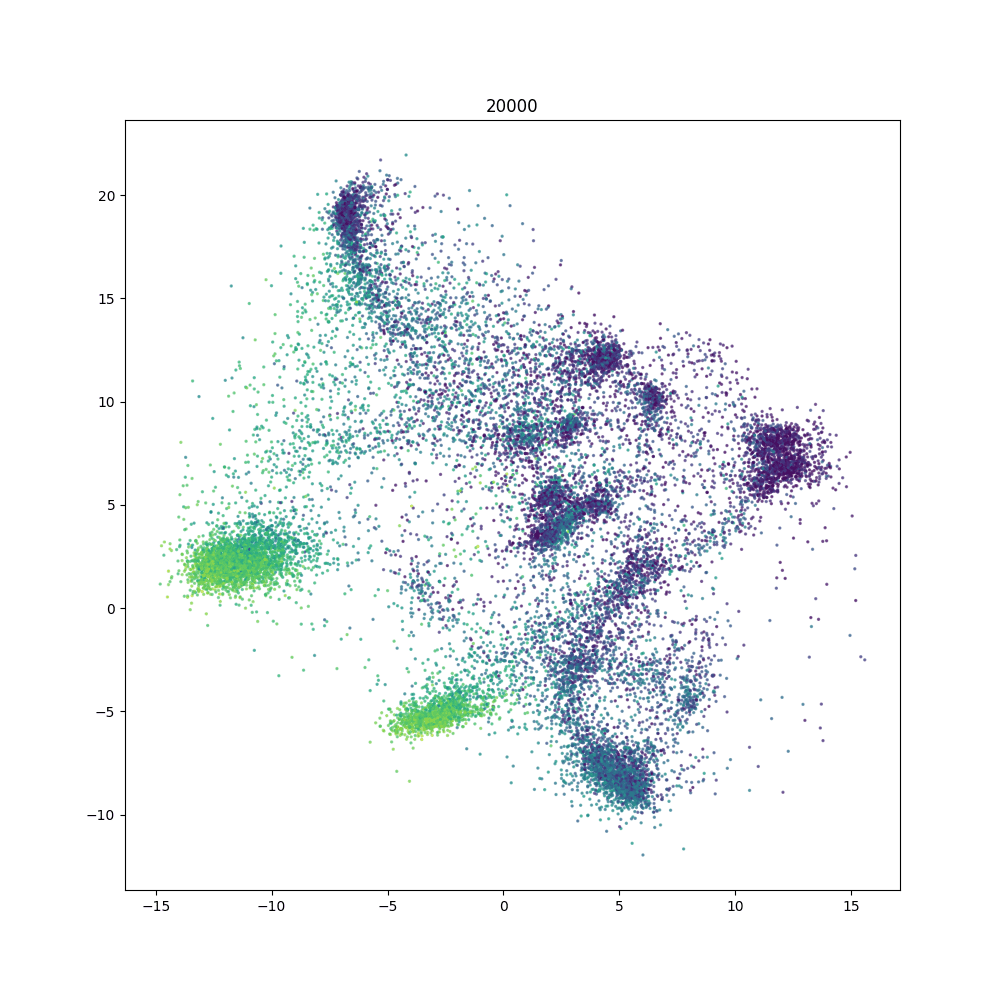} & \includegraphics[width=0.2\linewidth]{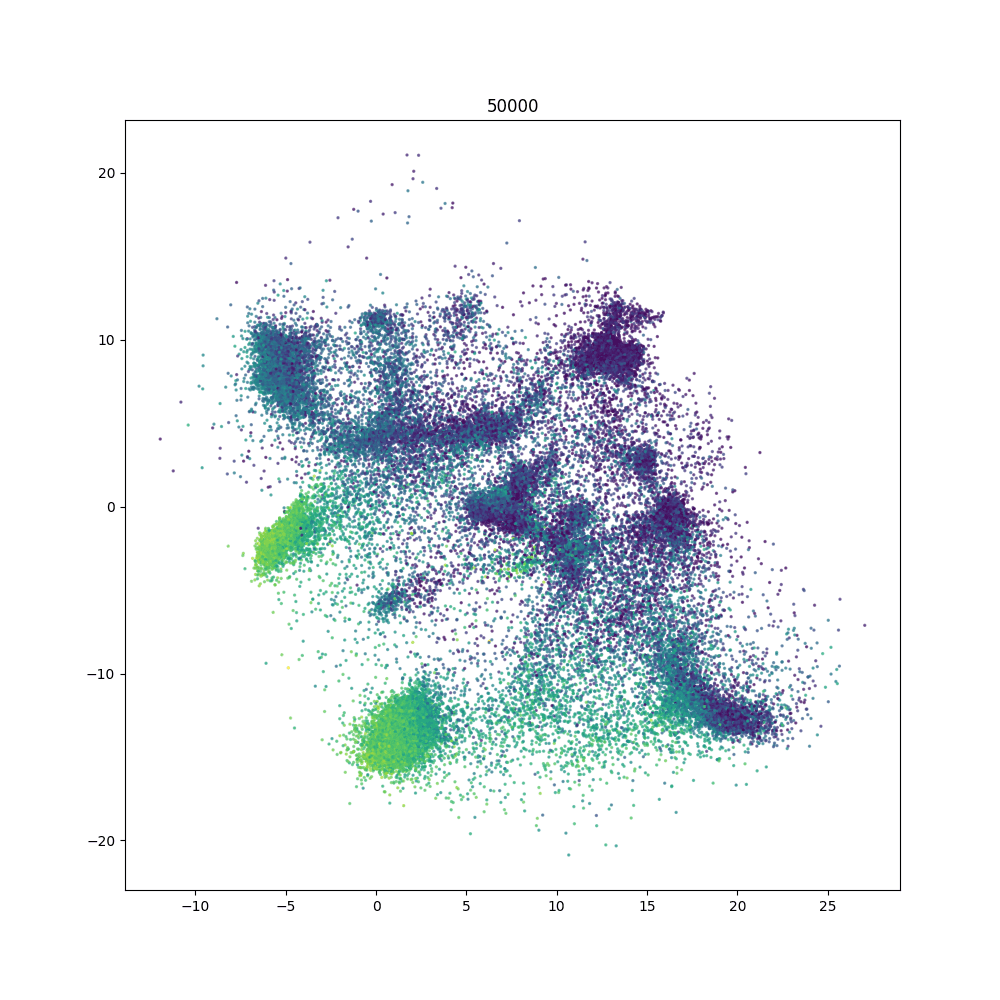} & \includegraphics[width=0.2\linewidth]{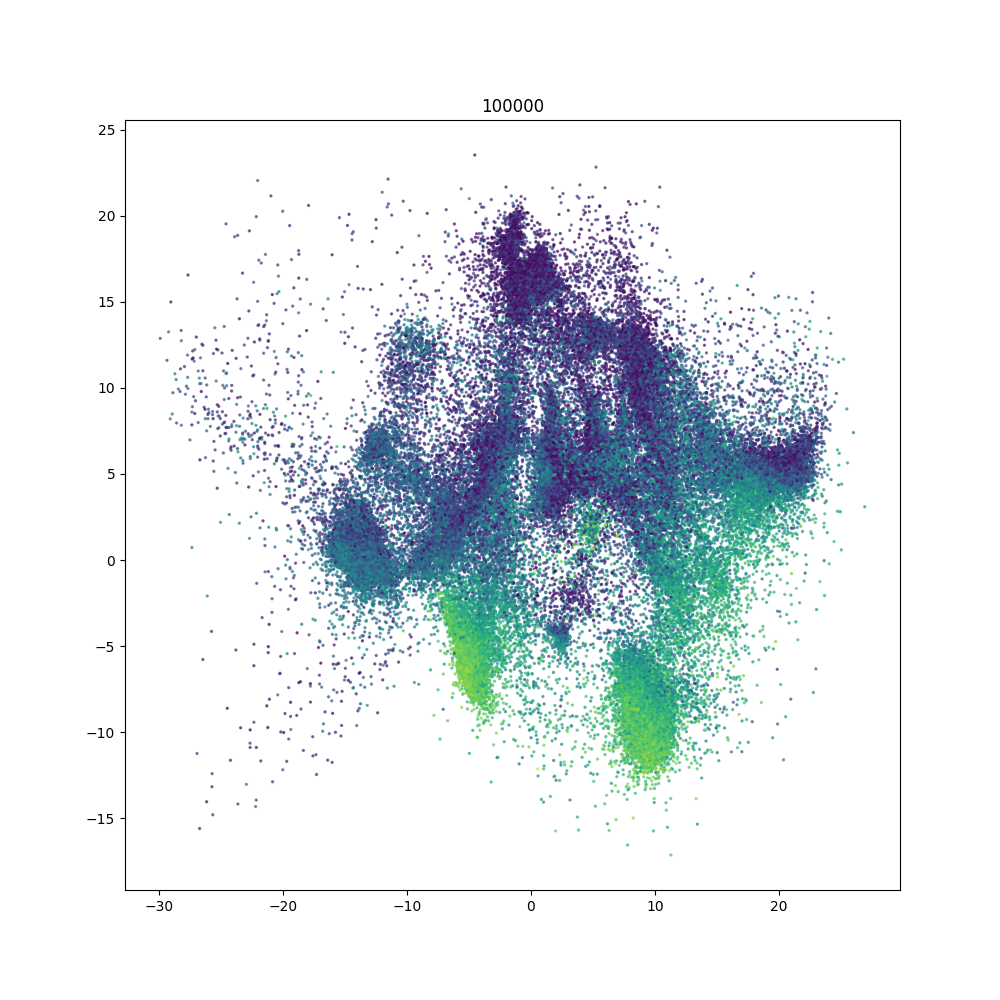} & \includegraphics[width=0.2\linewidth]{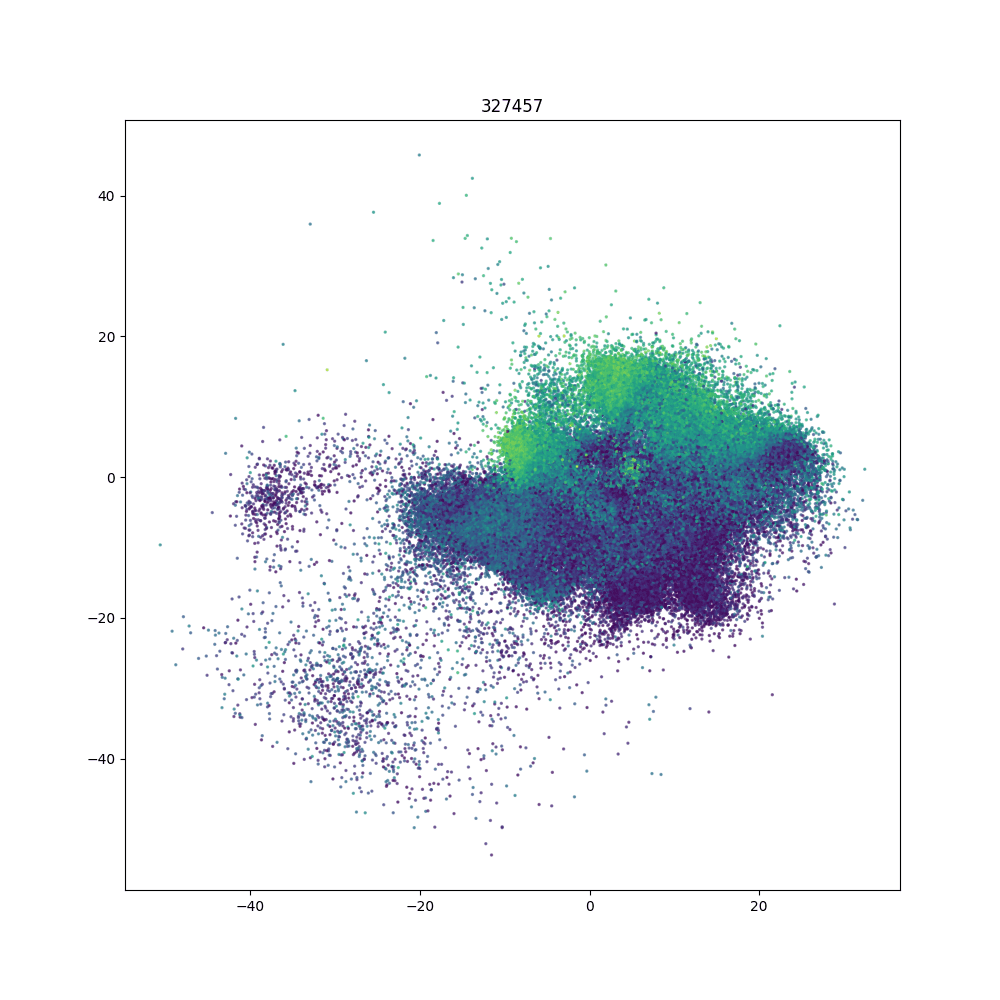} \\
     
     \begin{turn}{90}UMAP\end{turn} & \includegraphics[width=0.2\linewidth]{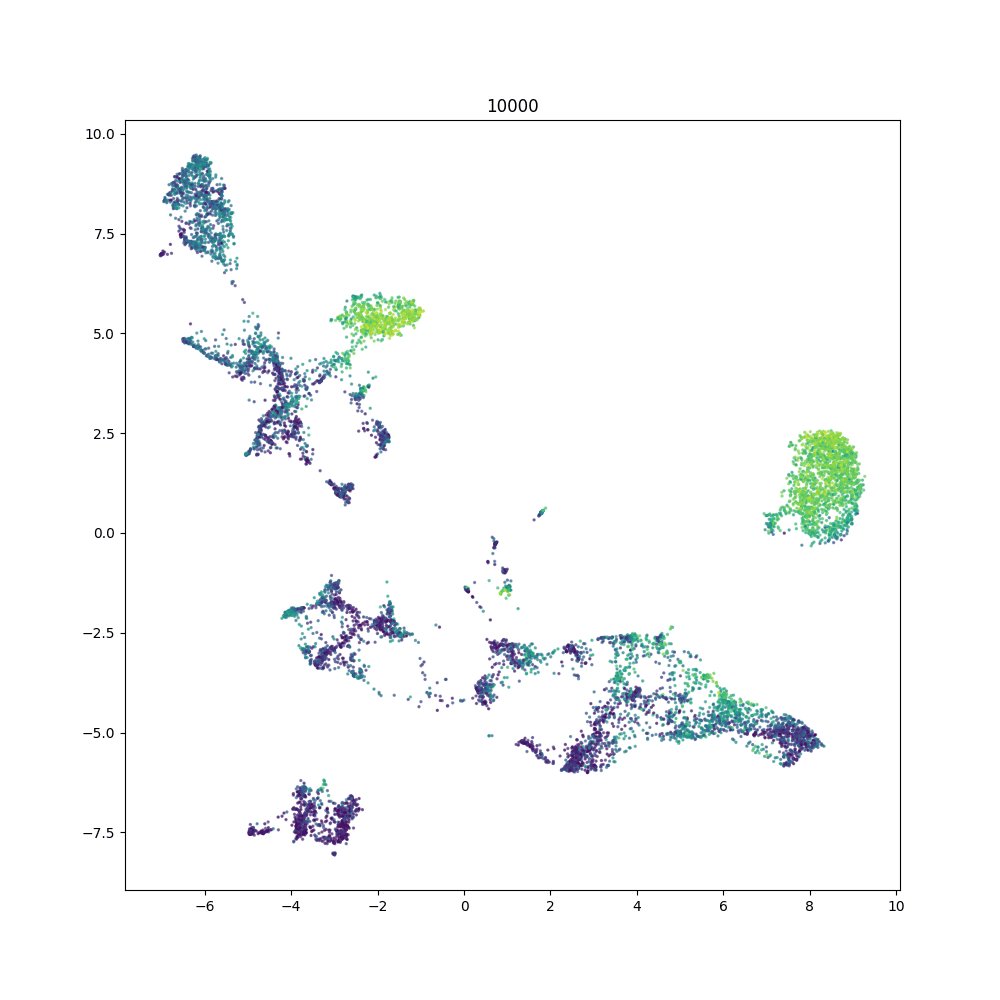} & \includegraphics[width=0.2\linewidth]{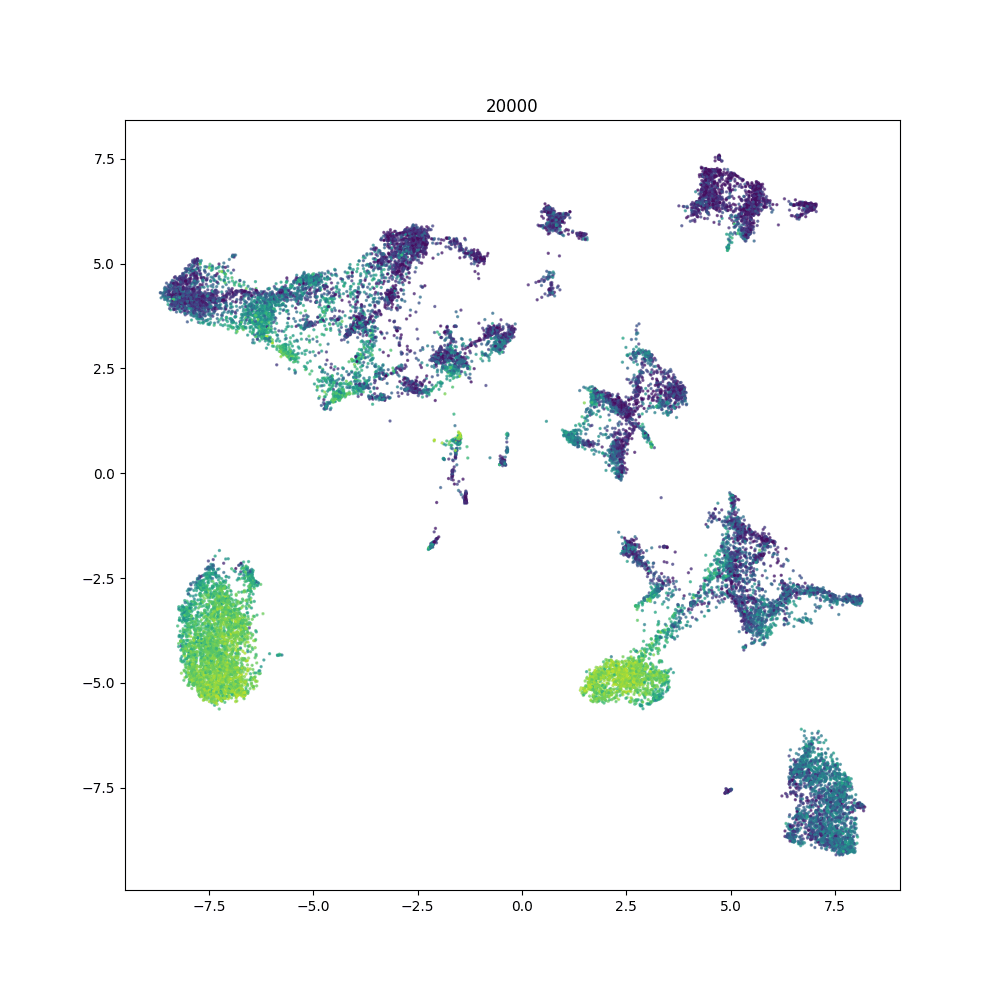} & \includegraphics[width=0.2\linewidth]{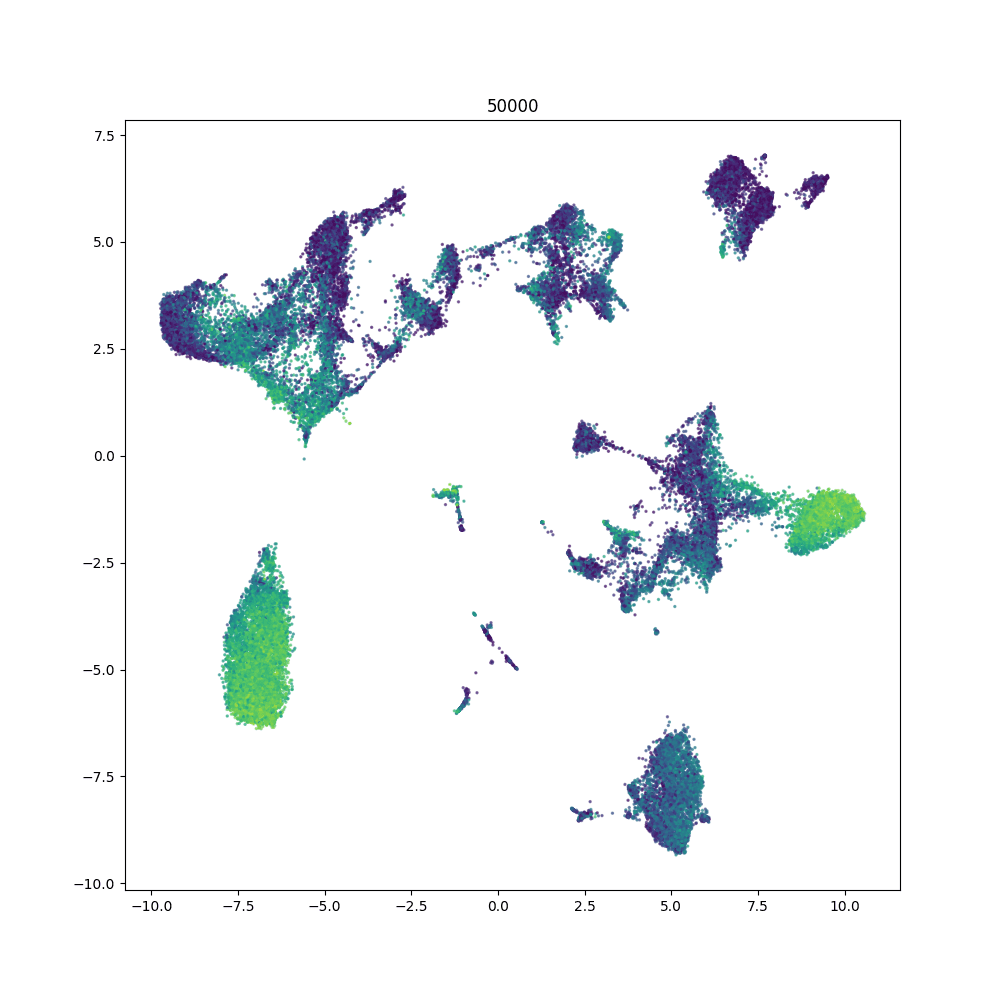} & \includegraphics[width=0.2\linewidth]{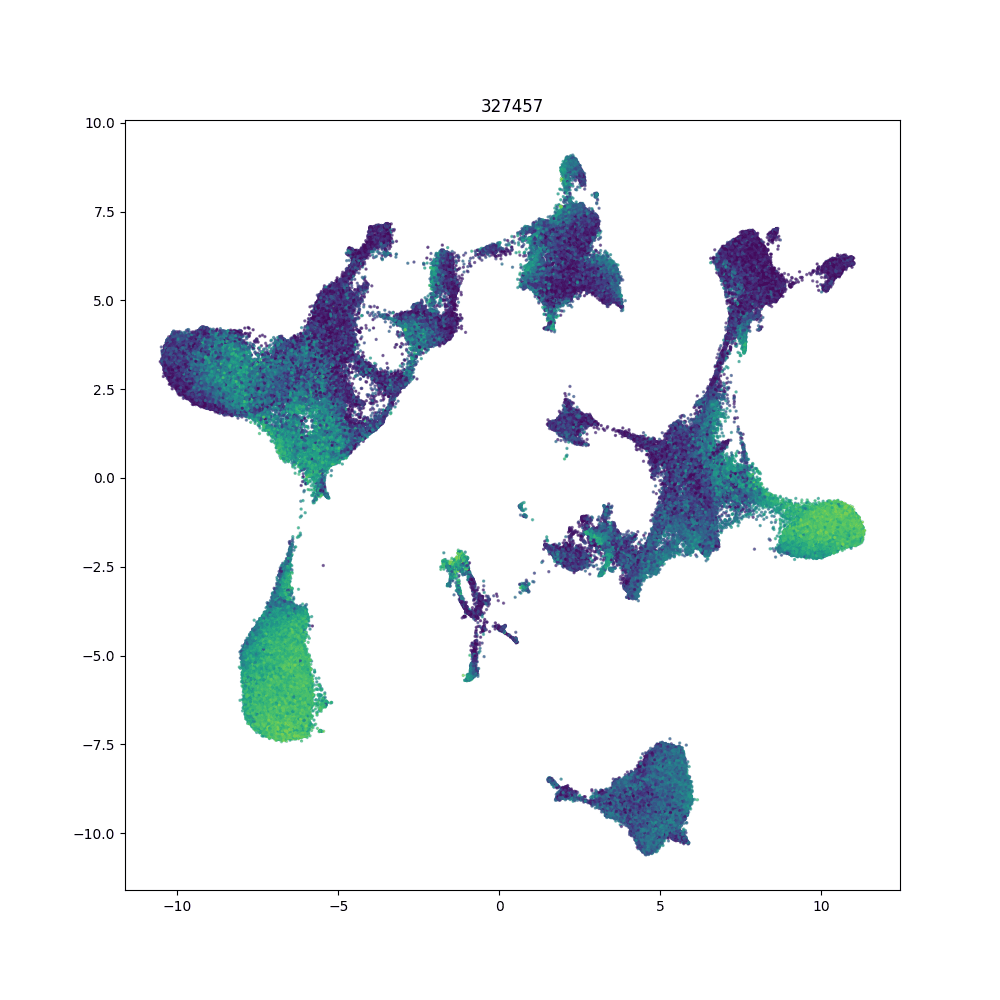} \\
     
     \begin{turn}{90}t-SNE\end{turn} & \includegraphics[width=0.2\linewidth]{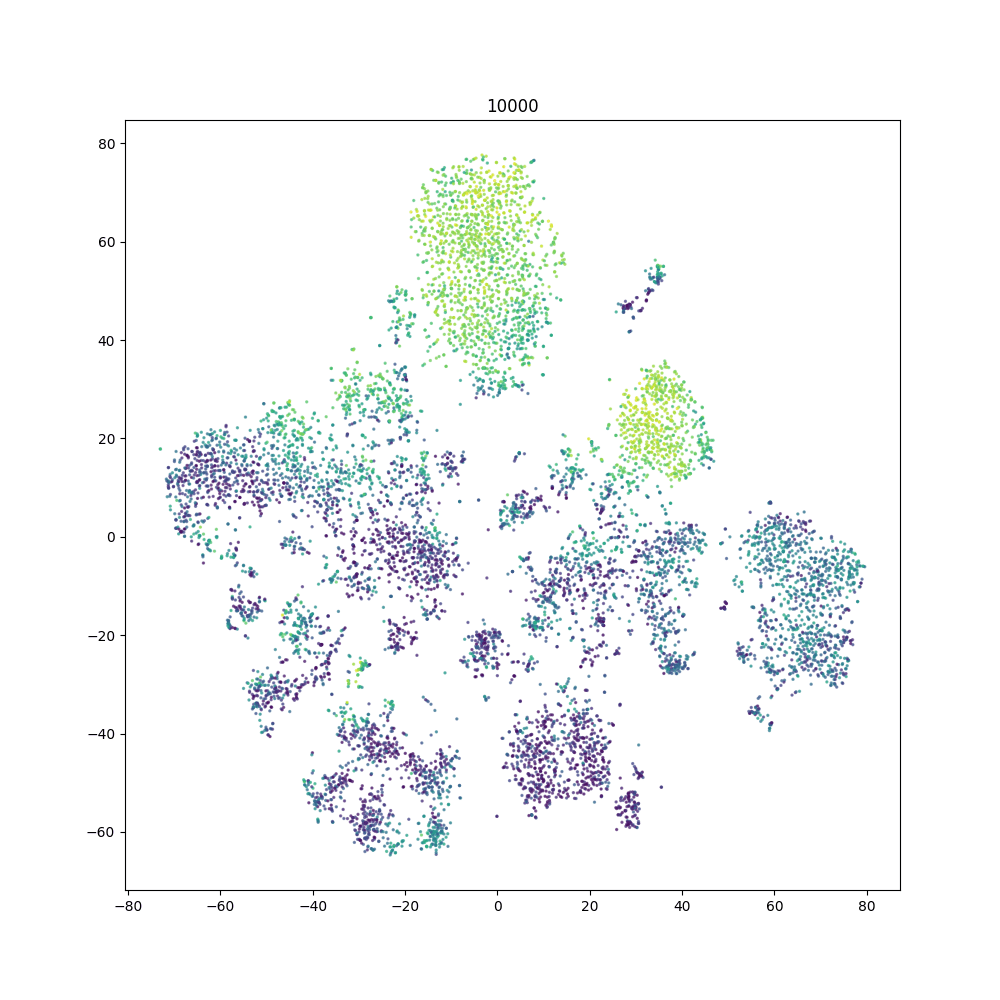} & \includegraphics[width=0.2\linewidth]{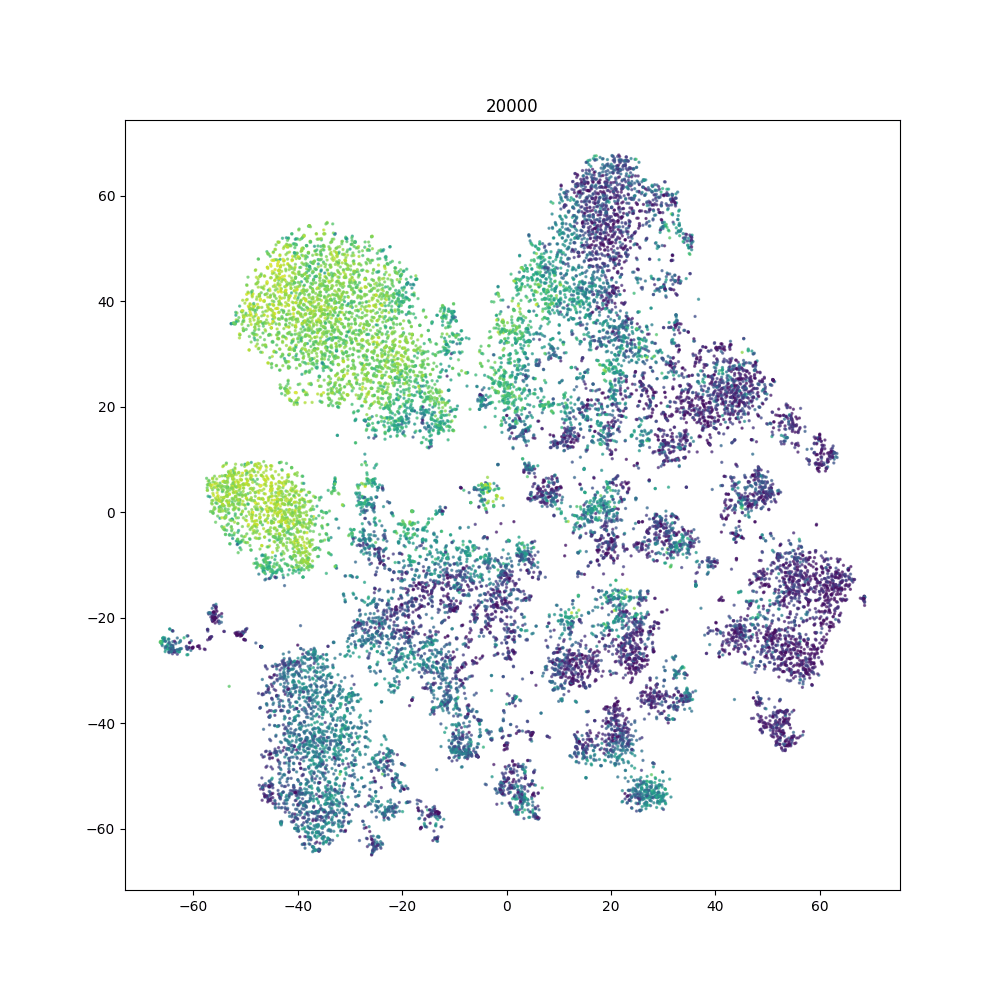} & \includegraphics[width=0.2\linewidth]{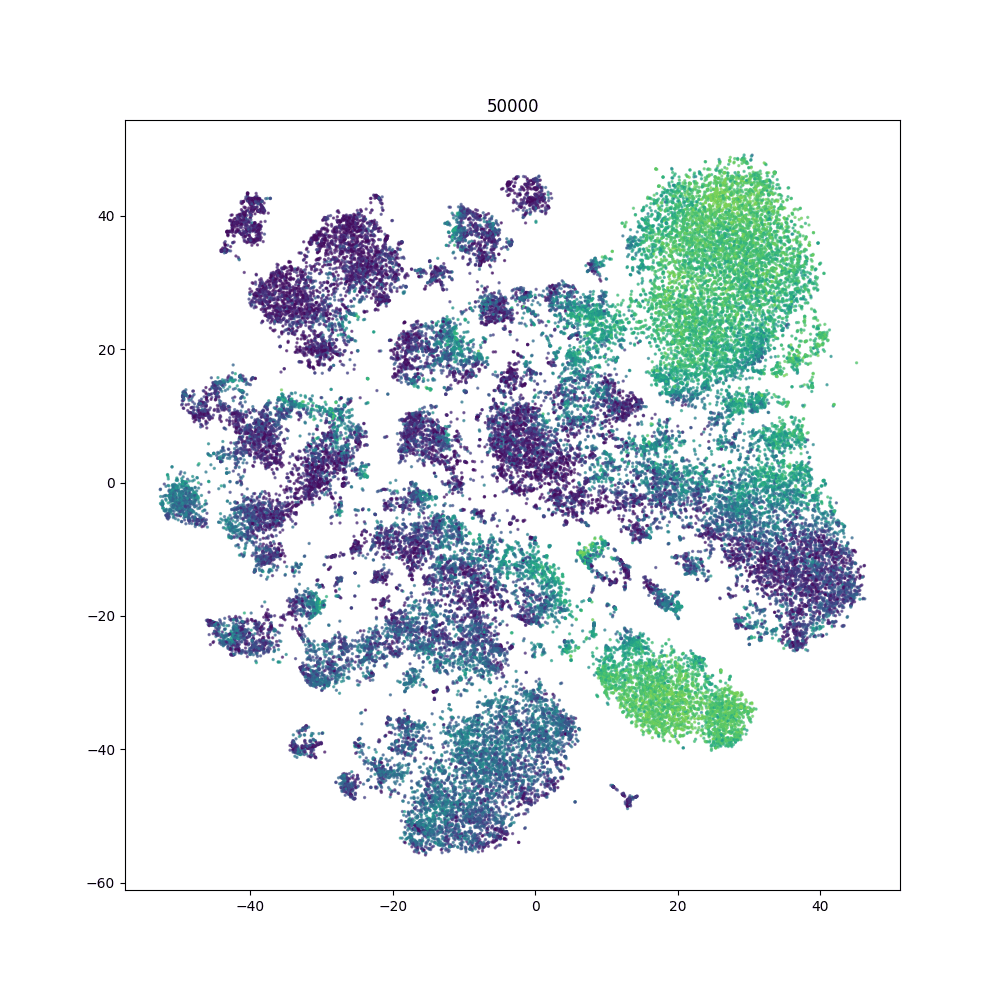} & \includegraphics[width=0.2\linewidth]{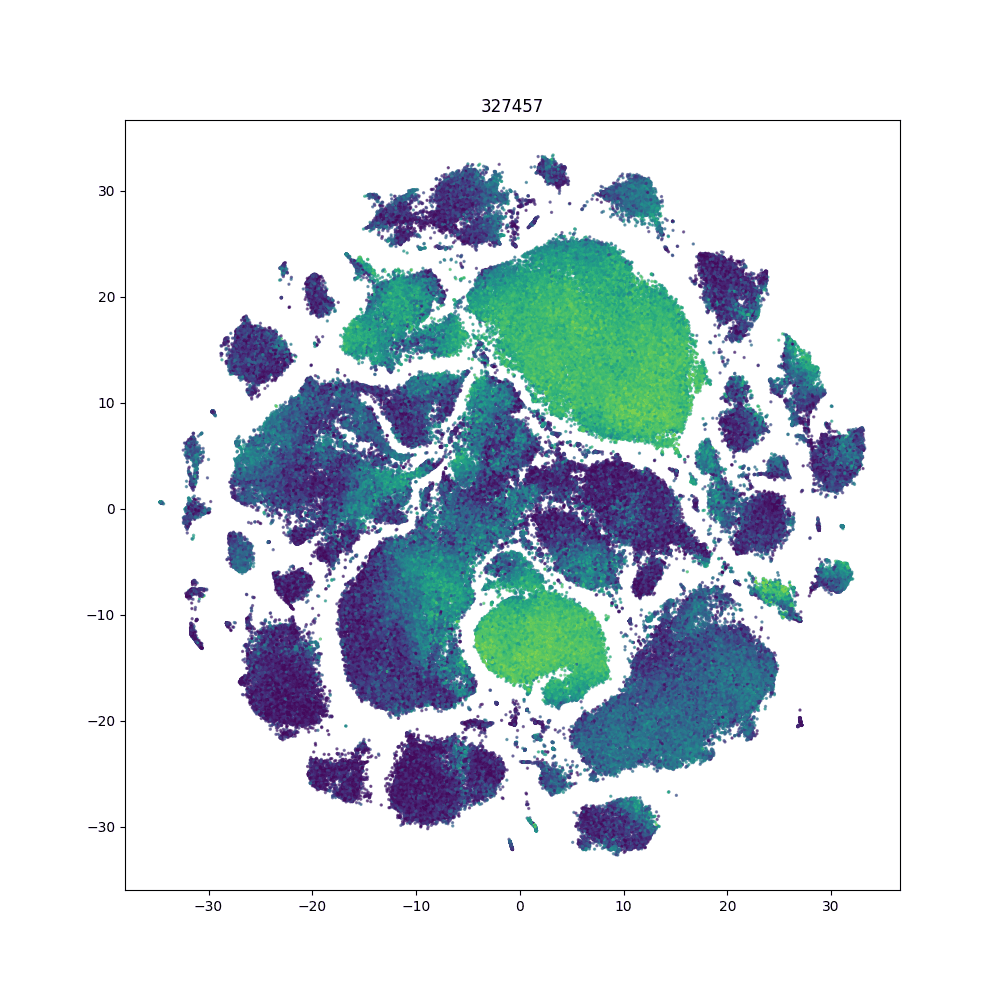} 
\end{tabular}

\caption{\label{wong_unbiased}Random samples of varying sizes from the Wong dataset presented to SONG, Parametric t-SNE, SONG + Reinit , UMAP and t-SNE  incrementally.  }

\end{figure*}

\begin{table}[t]
\caption{AMI scores of the visualized homogeneous increments of the Fashion MNIST and MNIST datasets. In SONG and Parametric t-SNE, a trained model from one intermediate dataset is updated and used to visualize the next dataset. SONG + REINIT, t-SNE, and UMAP are reinitialized and retrained at each increment. }
\label{Balanced-AMIS}
\centering
\begin{tabular}{p{2cm} | p{0.8cm} p{0.8cm} p{0.8cm} p{0.8cm} | p{0.8cm} p{0.8cm} p{0.8cm} p{0.8cm}}
\toprule
                 & \multicolumn{4}{c|}{Fashion MNIST} & \multicolumn{4}{c}{MNIST} \\
                 & 12k    & 24k    & 48k    & 60k    & 12k  & 24k  & 48k  & 60k  \\
\midrule
SONG             &\textbf{ 59.6}   & \textbf{60.1}   & \textbf{58.1}   & \textbf{59.5}   & \textbf{74.8} & \textbf{79.4} &\textbf{ 80.7 }&\textbf{ 81.9} \\
Parametric t-SNE & 51.2   & 55.8   & 57.2   & 54.8   & 44.9 & 53.3 & 57   & 61.2 \\
\hline
SONG-Reinit      & 59.6   & 58.8   & 61.2   & \textbf{61}     & 74.8 & 79.8 & 80.9 & 84   \\

t-SNE            & 58.9   & 58     & 59.4   & 53.5   & 77.5 & 74.1 & 78.6 & 77.4 \\
UMAP             & 60.1   & 59.5   & 58.7   & 59     & 76.6 & 80.8 & 83.2 & \textbf{84.9} \\
\bottomrule
\end{tabular}
\end{table}

\begin{figure*}
\centering
\subfigure[Fashion MNIST]{\includegraphics[width=0.49\linewidth]{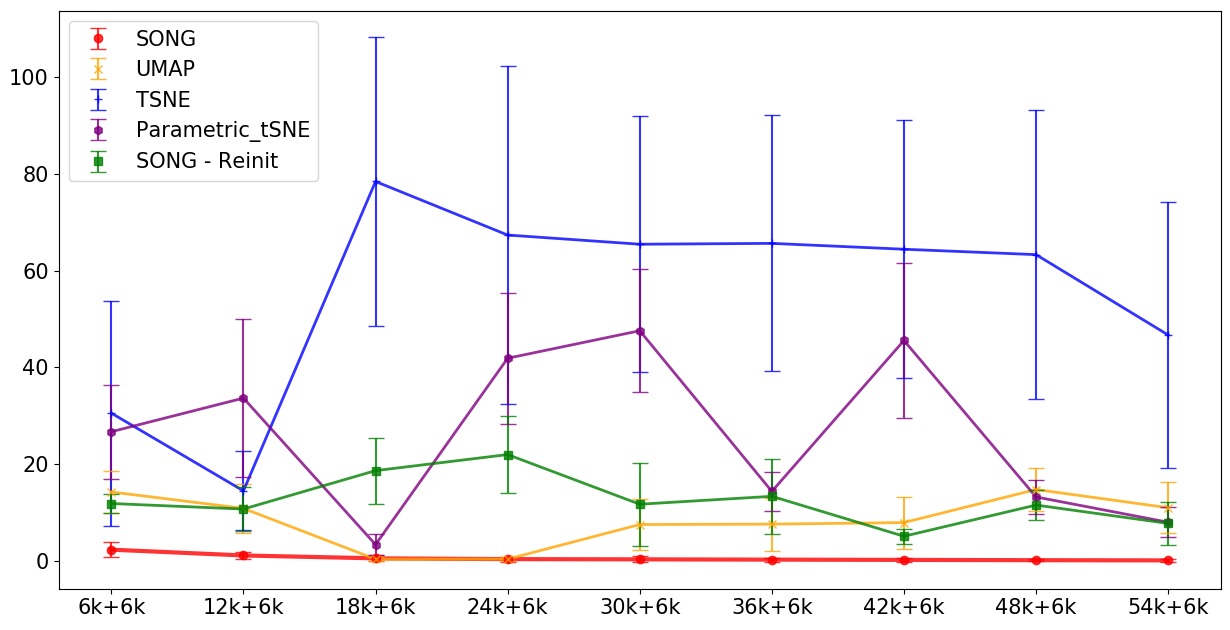}}
\subfigure[MNIST]{\includegraphics[width=0.49\linewidth]{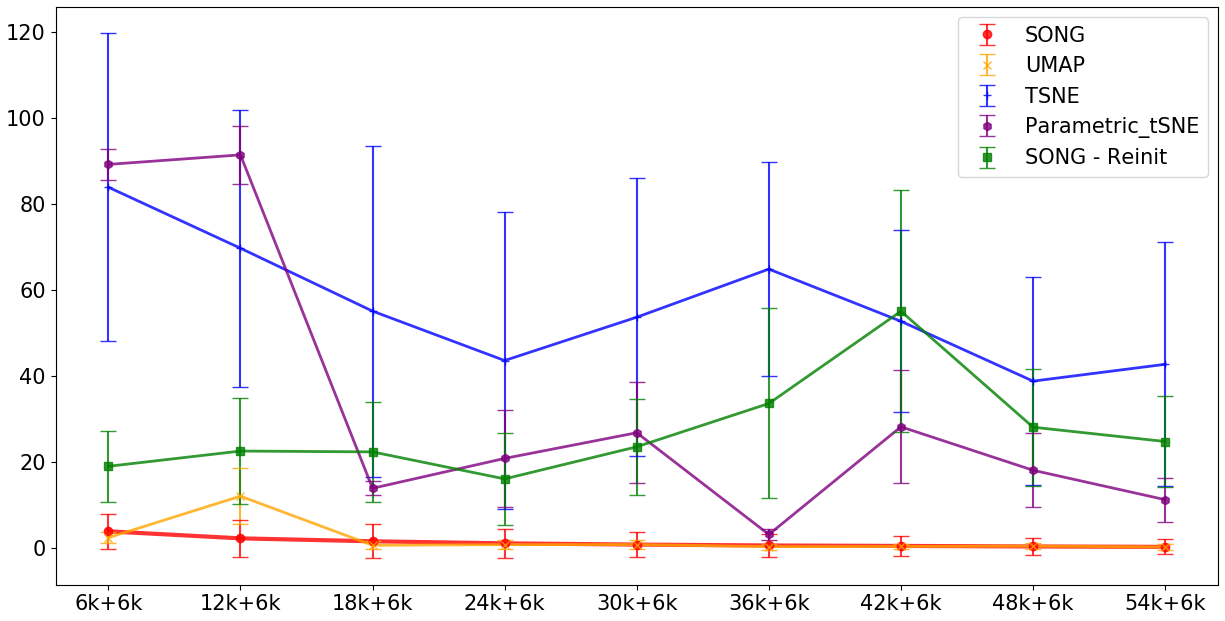}}
\centering
\caption{The average Consecutive Displacement of Y for each established point after subsequent presentation of 6000 images to each algorithm. }
\label{movement-curve}
\end{figure*}

\textbf{Result}: Among compared methods, SONG shows the highest stability in cluster placement when new data are presented, as shown in Fig.\ref{wong_unbiased} for the Wong dataset, and Fig. \ref{movement-curve} for the MNIST and Fashion-MNIST datasets. Furthermore, SONG shows good quality in the clusters inferred in the output embedding as per Table \ref{Balanced-AMIS}. In Table \ref{Balanced-AMIS}, we have highlighted the best scores for each increment in the model-retaining methods. However, for the model-reinitializing methods, we have highlighted the winner for the complete dataset. Compared to Parametric t-SNE, SONG has an average improvement of accuracy by 8.36\% on Fashion MNIST and 42.26\% on MNIST. Out of three model-reinitialized methods, UMAP has similar placements of clusters in consecutive visualizations, but shows rotations of the entire visualization in early to mid intermediate representations on the Wong dataset. UMAP stabilizes towards the last increments. However, for the three datasets, this stabilization happens at different stages. t-SNE shows arbitrary placement of clusters at each intermediate representation, making t-SNE not as good as UMAP or SONG for incremental visualizations. 

On Wong dataset, UMAP and SONG have more similar cluster placements in the homogeneous increment cases (see Fig. \ref{wong_unbiased}) than that of the heterogeneous increment cases (see Fig. \ref{wong_biased}). This increased similarity may be due to the fact that initial intermediate data samples in homogeneous cases more accurately represent the global structure of the data than that of heterogeneous cases. However, SONG shows no rotations in the visualization when data is augmented; in contrast, UMAP shows different orientations of similar cluster placements.

Table \ref{Balanced-AMIS} shows that the AMI scores for SONG on the MNIST dataset have increased as we present more data to the SONG algorithm. For the Fashion MNIST dataset, the AMI scores for SONG remain relatively low throughout the increments. This difference between trends may be due to the higher level of mixing of classes present in the Fashion MNIST dataset than in the MNIST, which makes it more difficult to separate the classes in Fashion MNIST into distinct clusters despite having more data. Table \ref{Balanced-AMIS} further shows that SONG provides visualizations of comparable quality to UMAP and superior to t-SNE and parametric t-SNE. We note that SONG generally produces lower AMIs than SONG + Reinit, possibly because SONG attempts to preserve the placement of points in existing visualizations which may cause some structural changes warranted by new data to be neglected. Neglecting such changes may explain the slight drop of performance in the incremental visualization vs the visualization of the complete dataset. However, we see that the incremental scores of SONG are not considerably worse than that of SONG + Reinit where SONG is trained from scratch at each increment on existing data and the newly presented data.

In Fig. \ref{movement-curve}, SONG has the lowest CDY values for both MNIST and Fashion MNIST throughout the increments. SONG also shows small standard deviations, showing that the CDYs for all points are indeed limited. In contrast, t-SNE has the largest displacements and standard deviations. Surprisingly, the average CDYs for each increment in Parametric t-SNE is relatively higher than heuristically reinitialized UMAP. We note that parametric t-SNE has a low standard deviation of displacement compared to t-SNE. Given a completely random re-arrangement of clusters would cause high standard deviation, this implies that parametric t-SNE produces translational or rotational displacements while keeping the cluster structure intact. Notably, UMAP does fairly well compared to other methods in terms of cluster displacement by having the second lowest average displacement and standard deviation. However, we see that in addition to having a smaller movement of points, SONG shows a strict decrease in displacement when more data are presented. SONG + Reinit shows comparatively large average CDYs as well as large standard deviations of CDYs, implying large movements between consecutive visualizations.

\subsection{Tolerance to Noisy and Highly Mixed Clusters}
We explore how well SONG performs in the presence of noisy data which we simulated as a series of datasets with high levels of cluster mixing and large cluster standard deviations. 

\label{noise_exp}
\textbf{Setup}: 
We compare SONG against UMAP and t-SNE on a collection of 32 randomly generated Gaussian Blobs datasets using 8 different cluster standard deviations (4, 8, 10, 12, 14, 16, 18, 20) and 4 different numbers of clusters (10, 20, 50, 100) for each standard deviation. These datasets have a dimensionality of 60. In addition, for each algorithm, we calculate the Adjusted Mutual Information (AMI) score for the visualizations compared to the known labels of the Gaussian clusters.

\begin{table*}[ht]
\centering
\caption{The AMI scores for different Gaussian Blobs configurations in 60 Dimensions. Each configuration of Gaussian Blobs have different numbers of clusters and different cluster standard deviations. Having a large number of clusters and a large standard deviation increases the probability of mixing of clusters.}
\begin{tabular}{p{0.8cm} p{1.6cm} |p{0.8cm} p{0.8cm} p{0.8cm} p{0.8cm} |p{0.8cm} p{0.8cm} p{0.8cm} p{0.8cm} |p{0.8cm} p{0.8cm} p{0.8cm} p{0.8cm} }
\toprule
 &  & \multicolumn{4}{c}{SONG} & \multicolumn{4}{|c}{UMAP} & \multicolumn{4}{|c}{t-SNE} \\
& \textit{No. Clusters} & \textit{10} & \textit{20} & \textit{50} & \textit{100} & \textit{10} & \textit{20} & \textit{50} & \textit{100} & \textit{10} & \textit{20} & \textit{50} & \textit{100} \\
\hline
\multirow{8}{*}{\begin{turn}{90}Cluster Std. Deviation\end{turn}} & 4 & \textbf{100} & \textbf{100} & \textbf{100} & \textbf{100} & \textbf{100} & \textbf{100} & \textbf{100} & \textbf{100} & \textbf{100} & \textbf{100} & \textbf{100} & \textbf{100} \\
 & 8 & \textbf{99.9} & \textbf{99.6} & \textbf{99.2} & \textbf{99.1} & \textbf{99.9} & 99.2 & 98.4 & 97.1 & 99.7 & 99.2 & 98.0 & 96.0 \\
 & 10 & \textbf{97.6} & \textbf{95.4} &\textbf{ 91.4} & \textbf{89.2} & 96.8 & 92.8 & 84.4 & 67.9 & 95.7 & 90.4 & 80.3 & 68.4 \\
 & 12 & \textbf{90.0} & \textbf{84.1} & \textbf{75.2} & \textbf{65.8} & 86.5 & 75.4 & 51.0 & 33.2 & 82.7 & 68.5 & 48.5 & 34.5 \\
 & 14 & \textbf{77.6} & \textbf{66.6} & \textbf{52.2} & \textbf{39.7} & 70.9 & 51.3 & 24.8 & 19.6 & 65.8 & 43.5 & 23.0 & 21.3 \\
 & 16 & \textbf{62.8} & \textbf{49.2} & \textbf{31.2} & \textbf{21.2} & 54.2 & 30.3 & 13.4 & 15.8 & 46.7 & 23.4 & 12.8 & 17.5 \\
 & 18 & \textbf{50.0} & \textbf{35.2} & \textbf{16.7} & \textbf{16.0} & 36.8 & 17.2 & 8.7 & 14.5 & 32.9 & 13.7 & 8.7 & 15.6 \\
 & 20 & \textbf{38.1} & \textbf{22.6} & \textbf{10.2} & \textbf{14.6} & 25.7 & 10.3 & 6.84 & 13.6 & 22.7 & 9.1 & 7.0 & 14.5 \\
\bottomrule
\end{tabular}

\label{noise-levels}
\end{table*}

\begin{figure*}
\centering
\begin{tabular}{c c c}
     SONG & UMAP & t-SNE  \\
     \includegraphics[width=0.3\linewidth]{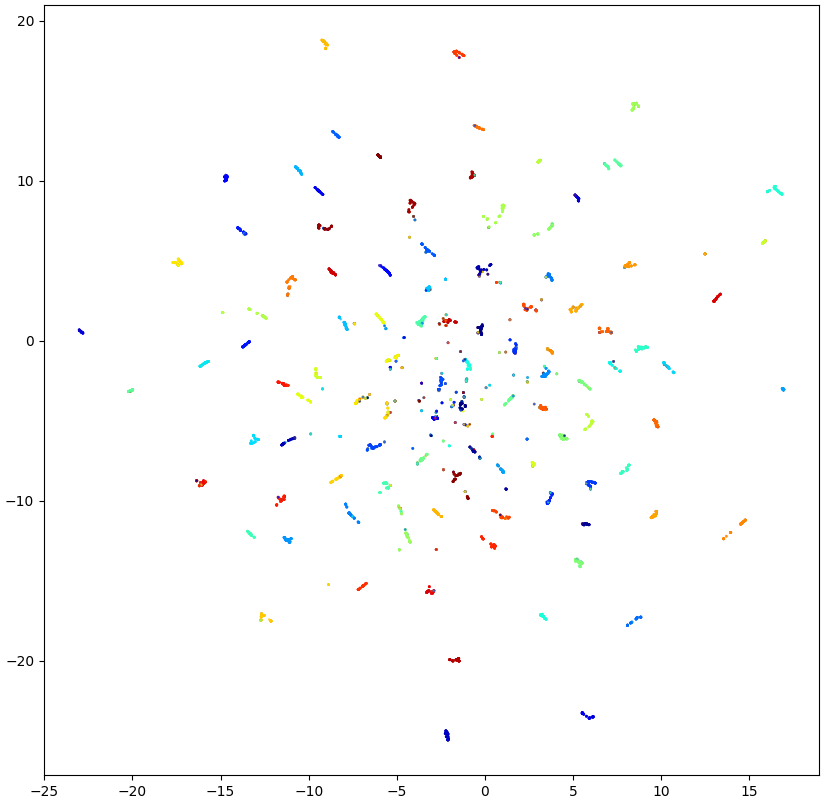}& \includegraphics[width=0.3\linewidth]{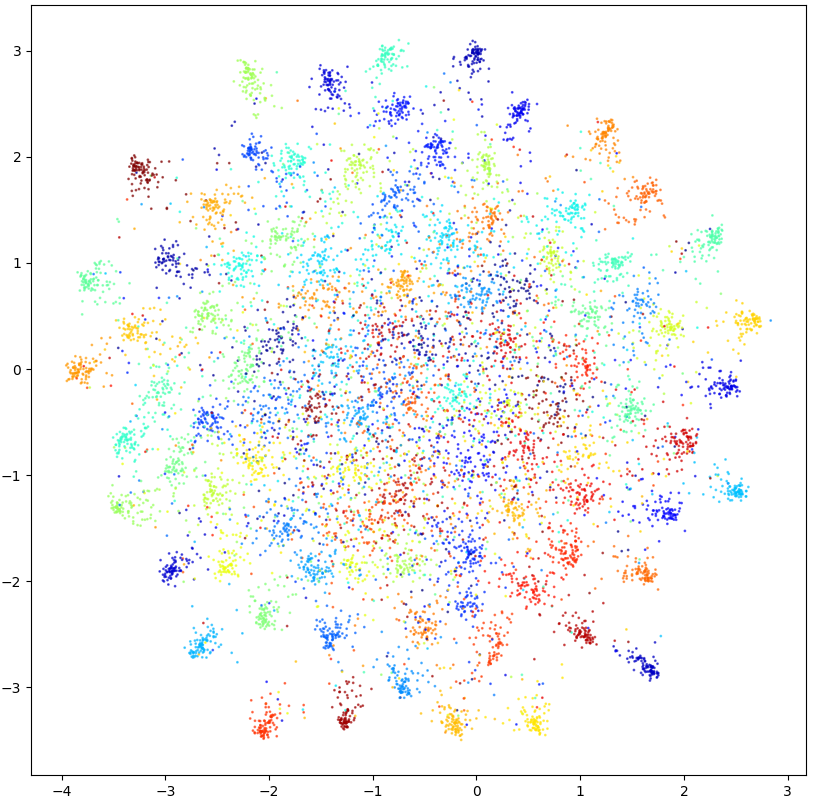}& \includegraphics[width=0.3\linewidth]{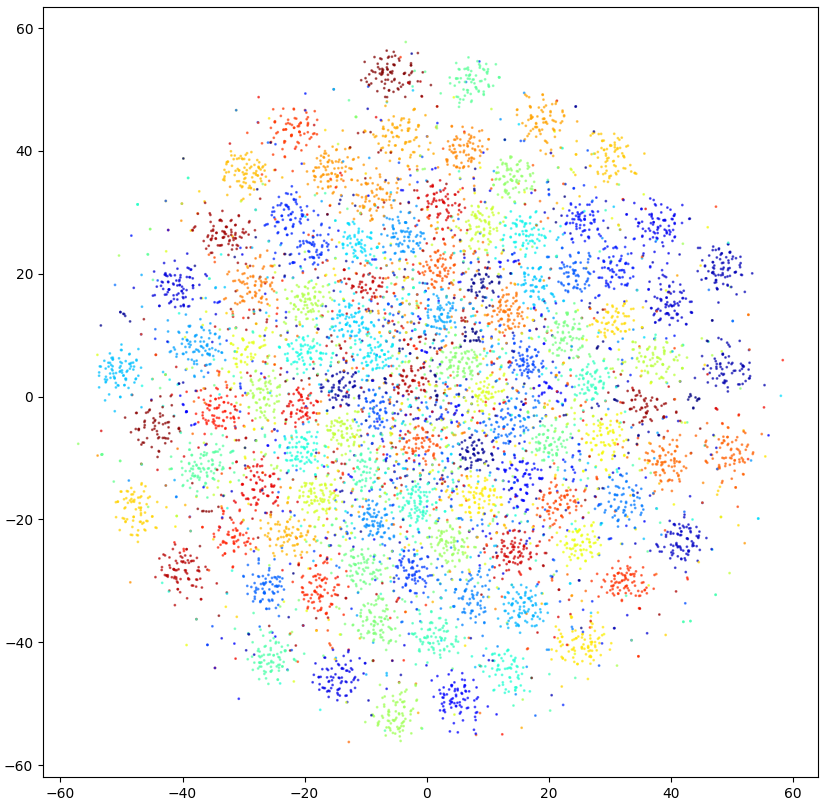}
\end{tabular}
\caption{\label{noise}The visualizations of a dataset having 100 random Gaussian clusters, each cluster having a standard deviation of 10, and 60 dimensions using the three methods SONG, UMAP and t-SNE}
\end{figure*}

\textbf{Result}: Table \ref{noise-levels} shows that SONG has the highest accuracy for discerning mixing clusters in all cases. The resulting visualizations for one of these datasets for SONG, UMAP and t-SNE are provided in Fig.\ref{noise}. We observe that the cluster representations in the visualizations by SONG are more concentrated than that of both UMAP and t-SNE. 

We refer to the Supplement Section 3.1 for an extended set of visualizations, where we additionally changed the dimensionality to observe how it affects these observations. In this extended study, we test the visualization performance of the three algorithms on an additional 125 datasets. These datasets are generated by simulating data corresponding to 5 cluster standard deviations (1, 2, 3, 4, 10), 5 numbers of clusters (3, 4, 20, 50, 100) and 5 numbers of dimensions (3, 15, 45, 60, 120). In these visualizations, consistent to our observations in Table \ref{noise-levels}, we observe that SONG has better separation of clusters while UMAP and t-SNE show fuzzy cluster boundaries when the level of cluster mixing increases. 

\subsection{Qualitative Topology Preservation of SONG}
\label{tp_exp}
To qualitatively examine the capability of SONG to preserve specific topologies in the input data, we used the COIL-20 dataset \cite{nene1996columbia}, which is frequently used to assess the topology preservation of visualization methods \cite{maaten2008visualizing}\cite{mcinnes2018umap}.

\textbf{Setup}: We compare SONG with UMAP and t-SNE on visualizing the COIL-20 dataset, which has 20 different objects, each photographed at pose intervals of 5-degrees, resulting in 1440 images in total. Each image has 4096 pixels. As preprocessing, we reduced the COIL-20 dataset down to its first 300 principle components. Because of the rotating pose angles, we expect to see 20 circular clusters in our visualization, where each cluster represents a different object. 

\begin{figure*}[ht]
    \centering
    \subfigure[t-SNE]{
        \frame{\includegraphics[width=0.25\textwidth]{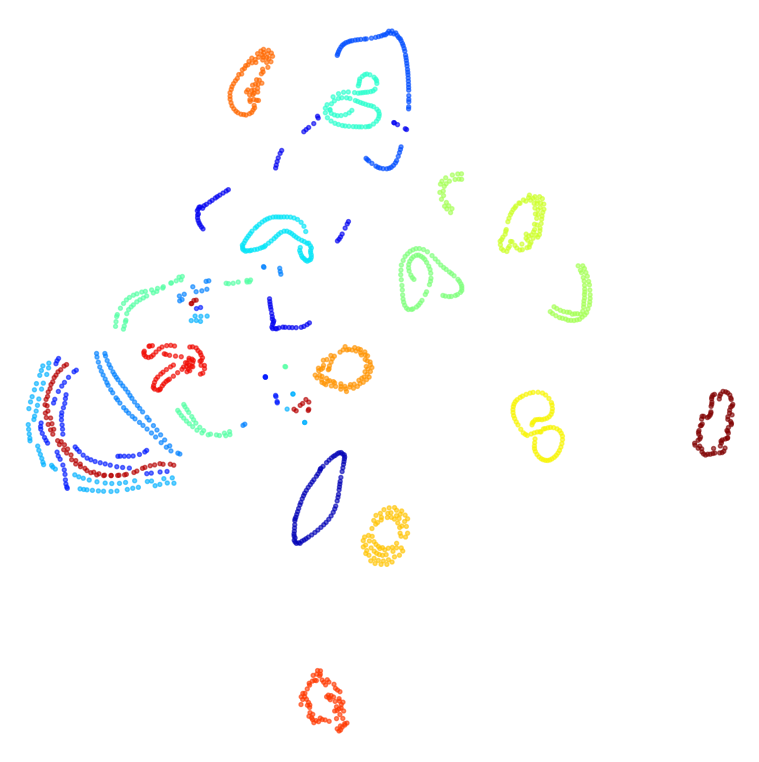}} }
        \subfigure[UMAP]{
        \frame{\includegraphics[width=0.25\textwidth]{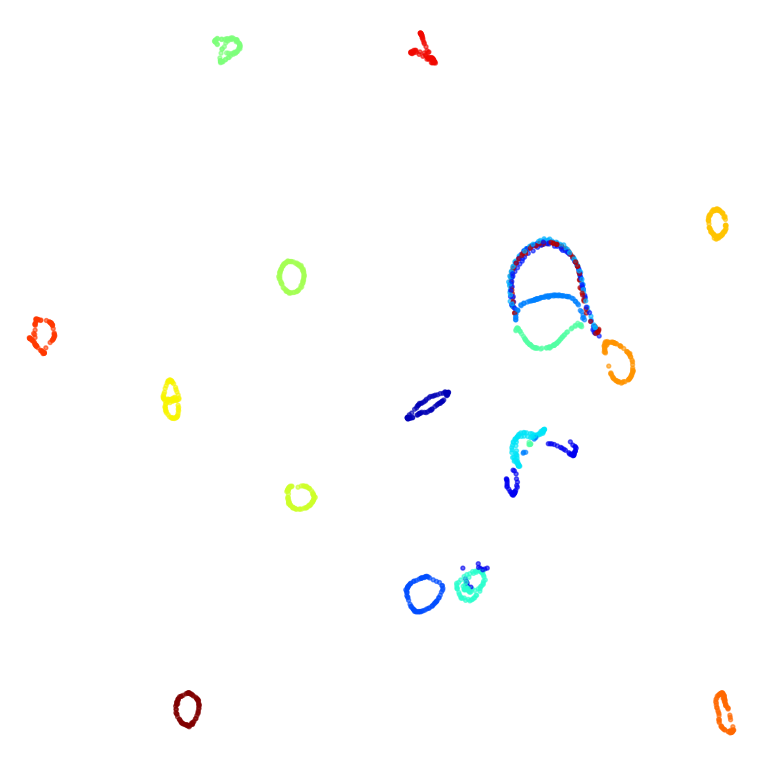}}}  
        \subfigure[SONG]{
        \frame{\includegraphics[width=0.25\textwidth]{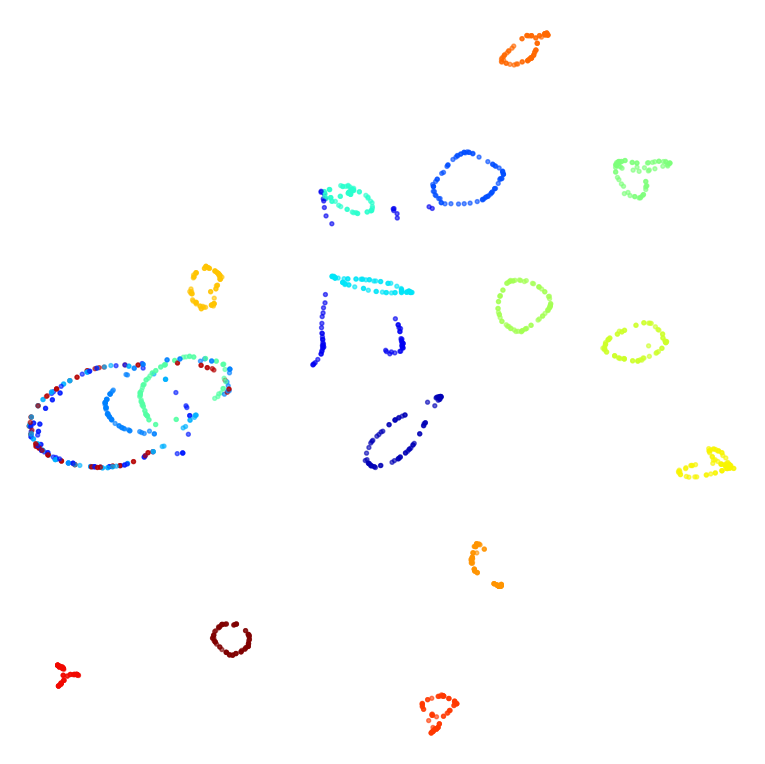}}}  
   
    \caption{COIL-20 dataset when reduced using the three algorithms a)t-SNE, b)UMAP and c)SONG. Both UMAP and SONG preserve the circular topologies, even in clusters where the classes are not well separated, to a greater degree than t-SNE. }\label{COIL20}
\end{figure*}

\textbf{Result}: The separation of circular clusters in SONG is similar to UMAP as shown in  Fig. \ref{COIL20}. For highly inseparable clusters, SONG and UMAP preserve the circular topologies better than t-SNE, where t-SNE shows an arch-like shape instead of circular structures.

\subsection{Running Time Comparison with t-SNE and UMAP}

In our Supplement Section 3.2, we show that SONG is faster for the same dataset configurations than t-SNE. However, as SONG needs to recalculate the pairwise distances between two sets of high-dimensional vectors ($\mathbf{X}$ and $\mathbf{C}$) multiple times for self-organization, SONG has a performance bottleneck which renders it slower than UMAP.

\section{Conclusions and Future Work}

In this work, we have presented a parametric nonlinear dimensionality reduction method called Self-Organizing Nebulous Growths (SONG) that can provide topology-preserving visualizations of high-dimensional data, while allowing new data to be mapped into existing visualizations without complete reinitialization. In our experiments, we presented SONG with both heterogeneous (Section \ref{biased_exp}) and homogeneous (Section \ref{unbiased_exp}) data increments, and observed that in both cases, SONG is superior to parametric t-SNE in preserving cluster placements when incorporating new data. Additionally, SONG's cluster visualization quality is on par with UMAP and superior to non-parametric t-SNE. We also showed that SONG is robust to noisy and highly mixed clusters (Section \ref{noise_exp}), and that SONG is capable of preserving specific topologies (Section \ref{tp_exp}) inferred from the input. 

The main merit of SONG is its usefulness in visualizing large datasets where considerable heterogeneity is present. This heterogeneity may be due to undesired batch effects \cite{gandolfo2018rle} or genuine variation in the populations of data. Consequently, SONG may be a promising tool for large-scale benchmarking projects that require coordination and curation of highly heterogeneous data, such as the Human Cell Atlas \cite{regev2017science}. 

However, SONG has a few limitations to be addressed in future work. First, SONG has a higher computational complexity than UMAP because in SONG, the high-dimensional parametric graph in the input space needs to be recalculated several times, whereas in UMAP, the high-dimensional KNN graph is constructed only once. We have empirically determined for the datasets considered in this paper that 8-10 recalculations of the graph are sufficient to provide a comparable approximation with UMAP. One possible direction of minimizing this graph reconstruction bottle-neck is to use batch gradient descent instead of stochastic gradient descent at later stages of learning. In our implementation, we chose stochastic gradient descent in an attempt to obtain an optimal visualization quality as the batch versions of self-organizing algorithms are prone to sub-optimal solutions \cite{fort2002advantages}. This may be viable at later stages of training when the graph is relatively stable and unchanging compared to the earlier stages. It should be noted that SONG still is less complex than t-SNE because SONG uses a negative sampling trick where we do not compute pairwise embedding distances globally (see Supplement Section 3.2).

Second, the current version of SONG cannot adjust the trade-off between the following two aspects of cluster placement: 
\begin{enumerate*}
    \item preserving the already inferred topological representations and
    \item adapting to represent new data which potentially alter the existing topology in high-dimensional space.
\end{enumerate*}
This trade-off is made evident by the discrepancy between the AMI scores of SONG and SONG + Reinit calculated on the incremental data visualizations. Besides, in the Wong dataset (Fig. \ref{wong_biased} and Fig. \ref{wong_unbiased}), the topology of the visualizations obtained using SONG is different from that obtained using SONG + Reinit, which may be explained by SONG's preference to preserve the topology of the visualizations of previous increments. Future work would explore the introduction of an `agility' parameter to regulate the aforementioned trade-off.

At last, throughout our manuscript, we discuss SONG as an unsupervised learning method. Another research direction is to use SONG in semi-supervised learning incorporating known or partially known labels of data to enhance cluster quality and separation of clusters. 

\section*{Acknowledgements}
The authors thank Mr. Rajith Vidanaarachchi and Ms. Tamasha Malepathirana for proof-reading. This work is partially funded by ARC DP150103512. 

\bibliographystyle{unsrt}
\bibliography{ref.bib}

\begin{thebibliography}{10}

\bibitem{fisher2012trust}
Danyel Fisher, Igor Popov, Steven Drucker, et~al.
\newblock Trust me, i'm partially right: incremental visualization lets
  analysts explore large datasets faster.
\newblock In {\em Proceedings of the SIGCHI Conference on Human Factors in
  Computing Systems}, pages 1673--1682. ACM, 2012.

\bibitem{montoya2003geo}
Lorena Montoya.
\newblock Geo-data acquisition through mobile {GIS} and digital video: an urban
  disaster management perspective.
\newblock {\em Environmental Modelling \& Software}, 18(10):869--876, 2003.

\bibitem{zhang2008visualization}
Heng Zhang, Benyu Zhang, Teresa Mah, Dong Zhuang, Jeremy Tantrum, and Ying Li.
\newblock Visualization application for mining of social networks, May~1 2008.
\newblock US Patent App. 11/555,279.

\bibitem{maaten2008visualizing}
Laurens van~der Maaten and Geoffrey Hinton.
\newblock {V}isualizing {D}ata using t-{SNE}.
\newblock {\em Journal of machine learning research}, 9(Nov):2579--2605, 2008.

\bibitem{mcinnes2018umap}
Leland McInnes, John Healy, and James Melville.
\newblock {UMAP}: {U}niform {M}anifold {A}pproximation and {P}rojection for
  dimension reduction.
\newblock {\em arXiv preprint arXiv:1802.03426}, 2018.

\bibitem{van2009learning}
Laurens Van Der~Maaten.
\newblock Learning a parametric embedding by preserving local structure.
\newblock In {\em Artificial Intelligence and Statistics}, pages 384--391,
  2009.

\bibitem{kohonen1982self}
Teuvo Kohonen.
\newblock Self-{O}rganized formation of topologically correct feature maps.
\newblock {\em Biological cybernetics}, 43(1):59--69, 1982.

\bibitem{hinton2003stochastic}
Geoffrey~E Hinton and Sam~T Roweis.
\newblock {S}tochastic {N}eighbor {E}mbedding.
\newblock In {\em Advances in neural information processing systems}, pages
  857--864, 2003.

\bibitem{tenenbaum2000global}
Joshua~B Tenenbaum, Vin De~Silva, and John~C Langford.
\newblock A global geometric framework for nonlinear dimensionality reduction.
\newblock {\em science}, 290(5500):2319--2323, 2000.

\bibitem{roweis2000nonlinear}
Sam~T Roweis and Lawrence~K Saul.
\newblock Nonlinear dimensionality reduction by {L}ocally {L}inear {E}mbedding.
\newblock {\em science}, 290(5500):2323--2326, 2000.

\bibitem{amid2019trimap}
Ehsan Amid and Manfred~K Warmuth.
\newblock {TriMap}: Large-scale dimensionality reduction using triplets.
\newblock {\em arXiv preprint arXiv:1910.00204}, 2019.

\bibitem{tang2016visualizing}
Jian Tang, Jingzhou Liu, Ming Zhang, and Qiaozhu Mei.
\newblock Visualizing large-scale and high-dimensional data.
\newblock In {\em Proceedings of the 25th international conference on world
  wide web}, pages 287--297, 2016.

\bibitem{van2014accelerating}
Laurens Van Der~Maaten.
\newblock Accelerating t-{SNE} using tree-based algorithms.
\newblock {\em The Journal of Machine Learning Research}, 15(1):3221--3245,
  2014.

\bibitem{linderman2019fast}
George~C Linderman, Manas Rachh, Jeremy~G Hoskins, Stefan Steinerberger, and
  Yuval Kluger.
\newblock Fast interpolation-based t-sne for improved visualization of
  single-cell rna-seq data.
\newblock {\em Nature methods}, 16(3):243--245, 2019.

\bibitem{song2019improved}
Weijing Song, Lizhe Wang, Peng Liu, and Kim-Kwang~Raymond Choo.
\newblock Improved t-{SNE} based manifold dimensional reduction for remote
  sensing data processing.
\newblock {\em Multimedia Tools and Applications}, 78(4):4311--4326, 2019.

\bibitem{amir2013visne}
El-ad~David Amir, Kara~L Davis, Michelle~D Tadmor, Erin~F Simonds, Jacob~H
  Levine, Sean~C Bendall, Daniel~K Shenfeld, Smita Krishnaswamy, Garry~P Nolan,
  and Dana Pe'er.
\newblock {viSNE} enables visualization of high dimensional single-cell data
  and reveals phenotypic heterogeneity of leukemia.
\newblock {\em Nature biotechnology}, 31(6):545, 2013.

\bibitem{belkin2003laplacian}
Mikhail Belkin and Partha Niyogi.
\newblock {L}aplacian {E}igenmaps for dimensionality reduction and data
  representation.
\newblock {\em Neural computation}, 15(6):1373--1396, 2003.

\bibitem{gisbrecht2015parametric}
Andrej Gisbrecht, Alexander Schulz, and Barbara Hammer.
\newblock Parametric nonlinear dimensionality reduction using {K}ernel t-{SNE}.
\newblock {\em Neurocomputing}, 147:71--82, 2015.

\bibitem{ding2018interpretable}
Jiarui Ding, Anne Condon, and Sohrab~P Shah.
\newblock Interpretable dimensionality reduction of single cell transcriptome
  data with deep generative models.
\newblock {\em Nature communications}, 9(1):2002, 2018.

\bibitem{becht2019dimensionality}
Etienne Becht, Leland McInnes, John Healy, Charles-Antoine Dutertre,
  Immanuel~WH Kwok, Lai~Guan Ng, Florent Ginhoux, and Evan~W Newell.
\newblock Dimensionality reduction for visualizing single-cell data using
  {UMAP}.
\newblock {\em Nature biotechnology}, 37(1):38, 2019.

\bibitem{zhang2016understanding}
Chiyuan Zhang, Samy Bengio, Moritz Hardt, Benjamin Recht, and Oriol Vinyals.
\newblock Understanding deep learning requires rethinking generalization.
\newblock {\em arXiv preprint arXiv:1611.03530}, 2016.

\bibitem{srivastava2015training}
Rupesh~K Srivastava, Klaus Greff, and J{\"u}rgen Schmidhuber.
\newblock Training very deep networks.
\newblock In {\em Advances in neural information processing systems}, pages
  2377--2385, 2015.

\bibitem{sturm2016interpretable}
Irene Sturm, Sebastian Lapuschkin, Wojciech Samek, and Klaus-Robert M{\"u}ller.
\newblock Interpretable deep neural networks for single-trial {EEG}
  classification.
\newblock {\em Journal of neuroscience methods}, 274:141--145, 2016.

\bibitem{fritzke1994growing}
Bernd Fritzke.
\newblock {G}rowing {C}ell structures—a self-organizing network for
  unsupervised and supervised learning.
\newblock {\em Neural networks}, 7(9):1441--1460, 1994.

\bibitem{fritzke1995growing}
Bernd Fritzke.
\newblock A {G}rowing {N}eural {G}as network learns topologies.
\newblock In {\em Advances in neural information processing systems}, pages
  625--632, 1995.

\bibitem{kamada1989algorithm}
Tomihisa Kamada, Satoru Kawai, et~al.
\newblock An algorithm for drawing general undirected graphs.
\newblock {\em Information processing letters}, 31(1):7--15, 1989.

\bibitem{weinreb2018spring}
Caleb Weinreb, Samuel Wolock, and Allon~M Klein.
\newblock {SPRING}: a kinetic interface for visualizing high dimensional
  single-cell expression data.
\newblock {\em Bioinformatics}, 34(7):1246--1248, 2018.

\bibitem{estevez2005cross}
Pablo~A Est{\'e}vez, Cristi{\'a}n~J Figueroa, and Kazimo Saito.
\newblock Cross-{E}ntropy approach to data visualization based on the {N}eural
  {G}as network.
\newblock In {\em Proceedings. 2005 IEEE International Joint Conference on
  Neural Networks, 2005.}, volume~5, pages 2724--2729. IEEE, 2005.

\bibitem{alahakoon2000dynamic}
Damminda Alahakoon, Saman~K Halgamuge, and Bala Srinivasan.
\newblock Dynamic {Self-Organizing Maps} with controlled growth for knowledge
  discovery.
\newblock {\em IEEE Transactions on neural networks}, 11(3):601--614, 2000.

\bibitem{munkres2014topology}
James Munkres.
\newblock {\em Topology}.
\newblock Pearson Education, 2014.

\bibitem{fort2002advantages}
Jean-Claude Fort, Patrick Letremy, and Marie Cottrell.
\newblock Advantages and drawbacks of the {B}atch {K}ohonen algorithm.
\newblock In {\em ESANN}, volume~2, pages 223--230, 2002.

\bibitem{mikolov2013distributed}
Tomas Mikolov, Ilya Sutskever, Kai Chen, Greg~S Corrado, and Jeff Dean.
\newblock Distributed representations of words and phrases and their
  compositionality.
\newblock In {\em Advances in neural information processing systems}, pages
  3111--3119, 2013.

\bibitem{chan2009investigation}
Chon-Kit~Kenneth Chan and Saman Halgamuge.
\newblock Investigation of average mutual information for species separation
  using {GSOM}.
\newblock In {\em International Conference on Future Generation Information
  Technology}, pages 42--49. Springer, 2009.

\bibitem{wong2016high}
Michael~Thomas Wong, David Eng~Hui Ong, Frances Sheau~Huei Lim, Karen Wei~Weng
  Teng, Naomi McGovern, Sriram Narayanan, Wen~Qi Ho, Daniela Cerny, Henry
  Kun~Kiaang Tan, Rosslyn Anicete, et~al.
\newblock A high-dimensional atlas of human {T} cell diversity reveals
  tissue-specific trafficking and cytokine signatures.
\newblock {\em Immunity}, 45(2):442--456, 2016.

\bibitem{lecun2010mnist}
Yann LeCun, Corinna Cortes, and CJ~Burges.
\newblock {MNIST} handwritten digit database.
\newblock {\em AT\&T Labs [Online]. Available: http://yann. lecun.
  com/exdb/mnist}, 2:18, 2010.

\bibitem{xiao2017fashion}
Han Xiao, Kashif Rasul, and Roland Vollgraf.
\newblock {Fashion-MNIST}: a novel image dataset for benchmarking machine
  learning algorithms.
\newblock {\em arXiv preprint arXiv:1708.07747}, 2017.

\bibitem{parks2006new}
David~R Parks, Mario Roederer, and Wayne~A Moore.
\newblock A new {“Logicle”} display method avoids deceptive effects of
  logarithmic scaling for low signals and compensated data.
\newblock {\em Cytometry Part A: The Journal of the International Society for
  Analytical Cytology}, 69(6):541--551, 2006.

\bibitem{vinh2009information}
Nguyen~Xuan Vinh, Julien Epps, and James Bailey.
\newblock Information theoretic measures for clusterings comparison: is a
  correction for chance necessary?
\newblock In {\em Proceedings of the 26th annual international conference on
  machine learning}, pages 1073--1080. ACM, 2009.

\bibitem{cao2003comparison}
LJ~Cao, Kok~Seng Chua, WK~Chong, HP~Lee, and QM~Gu.
\newblock A comparison of pca, kpca and ica for dimensionality reduction in
  support vector machine.
\newblock {\em Neurocomputing}, 55(1-2):321--336, 2003.

\bibitem{nene1996columbia}
Sameer~A Nene, Shree~K Nayar, Hiroshi Murase, et~al.
\newblock Columbia {O}bject {I}mage {L}ibrary (coil-20).
\newblock 1996.

\bibitem{gandolfo2018rle}
Luke~C Gandolfo and Terence~P Speed.
\newblock Rle plots: Visualizing unwanted variation in high dimensional data.
\newblock {\em PloS one}, 13(2):e0191629, 2018.

\bibitem{regev2017science}
Aviv Regev, Sarah~A Teichmann, Eric~S Lander, Ido Amit, Christophe Benoist,
  Ewan Birney, Bernd Bodenmiller, Peter Campbell, Piero Carninci, Menna
  Clatworthy, et~al.
\newblock Science forum: the {H}uman {C}ell {A}tlas.
\newblock {\em Elife}, 6:e27041, 2017.

\end{thebibliography}
\vspace{-20mm}

\begin{IEEEbiographynophoto}{Damith Senanayake, PhD}
Damith Senanayake obtained his PhD at the University of Melbourne in 2020, and completed his BSc in Computer Science and Engineering at the University of Moratuwa in 2015. His research focuses on developing analysis methods for high dimensional data using neural networks, computational and algebraic geometry and topology, with applications in bioinformatics. 
\end{IEEEbiographynophoto}
\vspace{-20mm}
\begin{IEEEbiographynophoto}{Wang Wei, PhD}
Wei Wang received the BE degree in Mechatronic Engineering from Zhejiang University in 2013 and the Ph.D. degree at the Department of Mechanical Engineering, the University of Melbourne in 2019. He is currently a postdoctoral at the University of Melbourne and will join Alibaba in 2020. He has research interests in Bayesian statistics, deep generative models, neural architecture search, and causal inference, and would like to devote himself to pushing the frontier of deep learning research.

\end{IEEEbiographynophoto}
\vspace{-22mm}
\begin{IEEEbiographynophoto}{Shalin H. Naik, PhD}
Shalin Naik received his BSc (Hons) from the University of QLD, Department of Microbiology and Immunology, his PhD from the University of Melbourne, Department of Medical Biology and The Walter \& Eliza Hall Institute of Medical Research (WEHI), and did his postdoc at the Netherlands Cancer Institute. He is a group leader at WEHI with a research interest in clonal biology in the areas of haematopoiesis, dendritic cells, stem cells and cancer evolution.\end{IEEEbiographynophoto}
\vspace{-20mm}

\begin{IEEEbiographynophoto}{Prof. Saman Halgamuge}
 Saman Halgamuge, (F’17) received the B.Sc. Engineering degree in Electronics and Telecommunication from the University of Moratuwa, Sri Lanka, and the Dipl.-Ing and Ph.D. degrees in data engineering from the Technical University of Darmstadt, Germany. He is currently a Professor of the Department of Mechanical Engineering of the School of EMI, The University of Melbourne, Australia. He is also a Distinguished Lecturer of IEEE Computational Intelligence Society (2018-21). His research interests are in AI, machine learning including deep learning, optimization, big data analytics and their applications in energy, mechatronics, bioinformatics and neural engineering. \end{IEEEbiographynophoto}
\vspace{-20mm}

\end{document}